\documentclass{article}
\usepackage{arxiv}
\usepackage{times}
\usepackage[round]{natbib}

\input{layout-options/fonts_02_anthropic}

\usepackage{amsmath,amsfonts,bm}

\def\eqref#1{equation~\ref{#1}}

\def\1{\bm{1}}

\def\vb{{\bm{b}}}

\def\vu{{\bm{u}}}

\DeclareMathAlphabet{\mathsfit}{\encodingdefault}{\sfdefault}{m}{sl}
\SetMathAlphabet{\mathsfit}{bold}{\encodingdefault}{\sfdefault}{bx}{n}

\usepackage{mathtools,amsmath}
\usepackage{amssymb}

\renewcommand{\eqref}[1]{\textup{(\ref{#1})}}
\usepackage{graphicx}
\usepackage{array,tabularx,booktabs}
\usepackage{multirow}
\usepackage{enumitem}
\usepackage{needspace}
\setlist[itemize]{leftmargin=*,itemsep=0.2em,topsep=0.4em,parsep=0pt,partopsep=0pt}
\setlist[enumerate]{leftmargin=*,itemsep=0.2em,topsep=0.5em,parsep=0pt,partopsep=0pt}
\usepackage{hyperref}
\hypersetup{pdfborder={0 0 0}}
\usepackage{url}
\usepackage{amsthm}
\usepackage[ruled,vlined,noend]{algorithm2e}

\usepackage{xcolor}
\usepackage[most]{tcolorbox}
\newcommand{\placeholder}[1]{{\color{red}[#1]}}
\newcommand{\figfile}[2][]{\IfFileExists{fig/#2}{\includegraphics[#1]{#2}}{\fbox{\parbox{0.9\linewidth}{\centering\placeholder{figure \detokenize{#2}}}}}}
\newcommand{\ourmethod}{{\normalfont\textsc{EventFlip}}}

\newtheoremstyle{paperresult}{6pt}{6pt}{\normalfont}{}{\bfseries}{.}{0.5em}{}
\theoremstyle{paperresult}
\newtheorem{theorem}{Theorem}
\newtheorem{proposition}[theorem]{Proposition}

\theoremstyle{definition}

\definecolor{resultbackground}{RGB}{243,249,242}
\definecolor{resultborder}{RGB}{167,193,169}
\colorlet{resulttext}{black}
\tcbset{paperresult/.style={enhanced jigsaw,breakable=true,colback=resultbackground,
  colframe=resultborder,coltext=resulttext,boxrule=0.4pt,arc=1pt,
  left=8pt,right=8pt,top=5pt,bottom=5pt,before skip=9pt,after skip=9pt}}
\tcolorboxenvironment{theorem}{paperresult}
\tcolorboxenvironment{proposition}{paperresult}
\tcolorboxenvironment{lemma}{paperresult}
\tcolorboxenvironment{corollary}{paperresult}

\RequirePackage[T1]{fontenc}
\RequirePackage{microtype}
\RequirePackage{titlesec}
\RequirePackage{caption}
\RequirePackage{xstring}
\makeatletter
\definecolor{layoutink}{HTML}{1F252B}
\definecolor{layoutmuted}{HTML}{667079}
\definecolor{layoutaccent}{HTML}{315FA5}
\newcommand{\LayoutLead}{}
\newcommand{\LayoutSubtitle}{}
\newcommand{\LayoutSplitTitle}{%
  \StrBefore{\@title}{:}[\LayoutLead]%
  \StrBehind{\@title}{:}[\LayoutSubtitle]}
\newcommand{\LayoutAuthor}{%
  {\fontsize{9.5}{13}\selectfont\begin{tabular}{@{}l@{}}\@author\end{tabular}}}
\newcommand{\LayoutBodySize}{10.5}
\newcommand{\LayoutLeading}{13.8}
\renewcommand{\normalsize}{%
  \@setfontsize\normalsize{\LayoutBodySize}{\LayoutLeading}%
  \abovedisplayskip=8pt plus 2pt minus 2pt
  \belowdisplayskip=8pt plus 2pt minus 2pt
  \abovedisplayshortskip=3pt plus 1pt
  \belowdisplayshortskip=5pt plus 1pt}
\renewcommand{\small}{\@setfontsize\small{9.3}{12}}
\renewcommand{\footnotesize}{\@setfontsize\footnotesize{8.5}{10.8}}
\newcommand{\LayoutGeometry}{\newgeometry{textwidth=168mm,top=23mm,bottom=24mm,headheight=14pt,headsep=13pt,footskip=25pt}}
\newcommand{\LayoutTitle}{%
  {\fontsize{24}{29}\selectfont\bfseries\@title\par}\vspace{15pt}\LayoutAuthor}
\renewcommand{\maketitle}{%
  \thispagestyle{plain}\par\begingroup\LayoutTitle\par\endgroup\vspace{19pt}}
\renewenvironment{abstract}{%
  \par\begingroup\small\setlength{\parskip}{5pt}%
  \noindent{\sffamily\bfseries\color{layoutink}Abstract}\par\vspace{3pt}%
}{\par\endgroup\vspace{15pt}}
\titleformat{\section}{\sffamily\bfseries\fontsize{14}{17}\selectfont\color{layoutink}}{\thesection}{0.8em}{}
\titleformat{\subsection}{\sffamily\bfseries\fontsize{11.5}{14}\selectfont\color{layoutink}}{\thesubsection}{0.65em}{}
\titleformat{\subsubsection}{\sffamily\bfseries\normalsize\color{layoutink}}{\thesubsubsection}{0.65em}{}
\titleformat{\paragraph}[runin]{\bfseries\normalsize}{\theparagraph}{0.5em}{}
\titlespacing*{\section}{0pt}{18pt plus 3pt minus 2pt}{8pt}
\titlespacing*{\subsection}{0pt}{13pt plus 2pt minus 2pt}{6pt}
\titlespacing*{\subsubsection}{0pt}{10pt plus 2pt}{5pt}
\titlespacing*{\paragraph}{0pt}{8pt plus 1pt}{0.8em}
\fancypagestyle{plain}{\fancyhf{}\fancyfoot[R]{\sffamily\fontsize{8.5}{10}\selectfont\color{layoutmuted}\thepage}}
\hypersetup{colorlinks=true,linkcolor=layoutaccent,citecolor=layoutaccent,urlcolor=layoutaccent,pdftitle={Flipping the Coins: Training Physical Simulators through Hard Switches}}
\renewcommand{\ourmethod}{{\rmfamily\mdseries\scshape EventFlip}}
\tcbset{paperresult/.append style={arc=2pt,boxrule=0.35pt,left=9pt,right=9pt,top=6pt,bottom=6pt,before skip=11pt,after skip=11pt}}
\AddToHook{begindocument/end}{\LayoutGeometry\normalsize\raggedbottom\pagestyle{fancy}}
\makeatother

\makeatletter
\definecolor{layoutaccent}{HTML}{9F563E}
\definecolor{layoutdisplay}{HTML}{713C43}
\definecolor{layoutink}{HTML}{292620}
\definecolor{layoutmuted}{HTML}{86766B}
\definecolor{resultbackground}{HTML}{F1F6F2}
\definecolor{resultborder}{HTML}{A8BCAF}
\renewcommand{\LayoutBodySize}{10.5}
\renewcommand{\LayoutLeading}{14.1}
\renewcommand{\LayoutGeometry}{\newgeometry{textwidth=166mm,top=24mm,bottom=24mm,headheight=14pt,headsep=15pt,footskip=25pt}}
\renewcommand{\LayoutTitle}{%
  \LayoutSplitTitle
  {\fontsize{9}{11}\selectfont\color{layoutmuted}\ourmethod\par}
  \vspace{19pt}
  {\fontsize{35}{39}\selectfont\color{layoutdisplay}\LayoutLead\par}
  \vspace{9pt}
  {\fontsize{21}{26}\selectfont\raggedright
    \StrSubstitute{\LayoutSubtitle}{ through }{\\through }[\LayoutSubtitle]%
    \color{layoutink}\LayoutSubtitle\par}
  \vspace{19pt}\LayoutAuthor\par
  \vspace{18pt}{\color{layoutaccent}\hrule height 0.65pt}}
\renewenvironment{abstract}{%
  \par\begingroup\fontsize{9.7}{13.1}\selectfont\setlength{\parskip}{5pt}%
  \noindent{\sffamily\bfseries\fontsize{10}{13}\selectfont\color{layoutaccent}Abstract}\par\vspace{4pt}%
}{\par\endgroup\vspace{14pt}}
\titleformat{\section}{\sffamily\bfseries\fontsize{14.5}{18}\selectfont\color{layoutaccent}}{\thesection}{0.8em}{}
\titleformat{\subsection}{\sffamily\bfseries\fontsize{11.8}{14.5}\selectfont\color{layoutink}}{\thesubsection}{0.65em}{}
\titlespacing*{\section}{0pt}{19pt plus 3pt minus 2pt}{8pt}
\titlespacing*{\subsection}{0pt}{14pt plus 2pt minus 2pt}{6pt}
\DeclareCaptionFont{layoutcaption}{\fontsize{9.2}{12.2}\selectfont}
\tcbset{paperresult/.append style={
  colback=resultbackground,colframe=resultborder,boxrule=0.3pt,arc=1.5pt,
  borderline west={1.2pt}{0pt}{resultborder},left=10pt,right=10pt,
  top=7pt,bottom=7pt,before skip=12pt,after skip=12pt}}

\makeatother

\graphicspath{{fig/}}

\title{Flipping the Coins: Training Physical Simulators through Hard Switches}
\renewcommand{\shorttitle}{Flipping the Coins}

\author{
{\fontsize{11}{14}\selectfont Tianrun Gao\textsuperscript{1,*}\quad Ruiqi Feng\textsuperscript{2,*}\quad Haomin Wen\textsuperscript{3,*}\quad Tailin Wu\textsuperscript{2}} \\[2pt]
{\fontsize{11}{14}\selectfont Liangqiong Qu\textsuperscript{1}\quad Xiaomeng Huang\textsuperscript{4,\ensuremath{\dagger}}\quad Ruiqi Shu\textsuperscript{4,*,\ensuremath{\dagger}}} \\[7pt]
\textsuperscript{1}\,School of Computing and Data Science, The University of Hong Kong \\
\textsuperscript{2}\,Department of Artificial Intelligence, School of Engineering, Westlake University \\
\textsuperscript{3}\,Shanghai Innovation Institute, Shanghai, China \\
\textsuperscript{4}\,Department of Earth System Science, Tsinghua University \\[5pt]
{\fontsize{8.5}{11}\selectfont\textsuperscript{*}\,Equal contribution.\quad\textsuperscript{\ensuremath{\dagger}}\,Corresponding authors.} \\[3pt]
Ruiqi Shu: \href{mailto:srq24@mails.tsinghua.edu.cn}{\texttt{srq24@mails.tsinghua.edu.cn}}\quad Xiaomeng Huang: \href{mailto:hxm@tsinghua.edu.cn}{\texttt{hxm@tsinghua.edu.cn}}
}
\hypersetup{pdfauthor={Tianrun Gao; Ruiqi Feng; Haomin Wen; Tailin Wu; Liangqiong Qu; Xiaomeng Huang; Ruiqi Shu}}

\begin{document}

\maketitle

\begin{abstract}
Hybrid simulators embed learnable components such as neural networks in physical solvers and train them on reference trajectories. Many scientific solvers feature hard switches, such as a tripping power line, an igniting grid cell, or depolarizing brain tissue, where the loss exhibits discrete jumps across switch flips, leaving standard automatic differentiation to capture only its smooth part.
Existing training paradigms face a fundamentally challenging trade-off, as gradient-based relaxations smooth the switches and introduce bias across jumps, whereas derivative-free methods treat the solver as a black box and discard path gradients of the continuous dynamics.
Here we introduce \ourmethod{}, which flips a near-threshold switch from a stored state, replays the original solver, and measures the loss difference between the two runs, in effect a finite difference across the switch.
By perturbing every threshold with small Gaussian noise, we cast each switch as a stochastic binary decision and derive an unbiased gradient estimator that combines the continuous path gradient with a jump term constructed from these counterfactual rollouts. \textsc{\ourmethod-Rand} optimizes this noise-averaged objective, whereas \textsc{\ourmethod-Hard} operates on the deterministic hard-switch simulator used at deployment and provably targets the complete discontinuous loss. 
Under an identical forward evaluation budget, we benchmark \ourmethod{} against twelve baselines across a power-grid cascade, a detonation wave, and cortical spreading depolarization, achieving the lowest error on every task and reducing the strongest baseline's error by factors of 1.5, 1.4, and 1.4, respectively.

\end{abstract}

\begin{figure}[ht!]
\centering
\includegraphics[width=\linewidth]{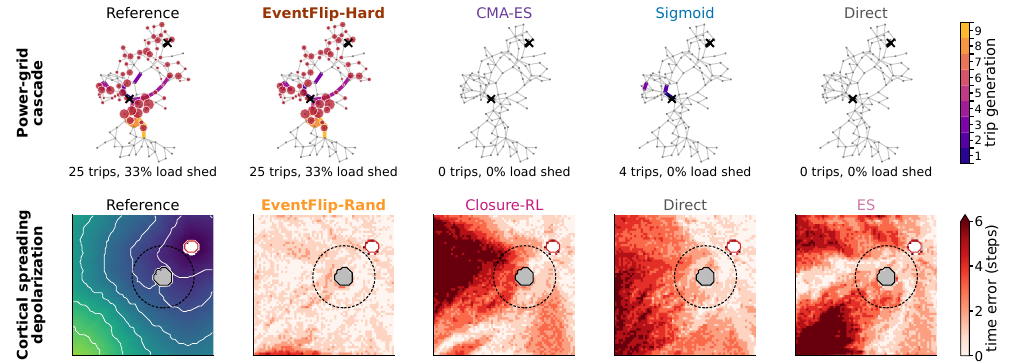}
\caption{\textbf{Representative test cases on the power-grid cascade and cortical spreading depolarization.} In each row, the leftmost panel depicts the ground-truth reference, while the remaining panels compare four hybrid simulators trained under an identical forward evaluation budget. In the top row, line colors indicate the cascade generation in which each transmission line trips, disk areas denote the load shed at each bus, and black crosses mark the two initial line outages. In the bottom row, the reference panel displays the cellular activation time with contour lines, whereas the four model panels show the corresponding activation-time error in steps.}
\label{fig:teaser}
\end{figure}

\section{Introduction}
\label{sec:intro}

Scientific machine learning increasingly embeds learnable components into numerical solvers, such as neural networks that correct coarse-grid errors or networks that supply unknown thresholds and constitutive relations, and trains the resulting hybrid simulator end-to-end on reference trajectories \citep{um2020solver,kochkov2021machine,shen2023differentiable,kochkov2024neural}. Many systems targeted by these hybrid simulators are governed by threshold crossings, where the dynamics change abruptly once a state variable crosses a critical value. In power grids, whether an overloaded line trips determines whether an initial outage stays local or grows into a blackout \citep{soltan2014cascading}. In combustion, whether a mixture ignites decides whether a detonation forms and how fast it travels, which governs the safety of fuel handling and the operation of combustion devices \citep{oran2007origins,shepherd2009detonation}. In the brain, whether cells cross a depolarization threshold determines whether spreading depolarization waves invade the ischemic penumbra to enlarge an infarct \citep{dreier2011role}.
Making these simulators with switches trainable on observed dynamics is therefore a prerequisite for reliably deploying learned models in various high-consequence physical regimes.

The fundamental obstacle is that many simulators represent threshold crossings as hard switches evaluated at individual grid cells or components. A switch is a step function of the physical state, so the simulated trajectory is a piecewise-smooth, discontinuous function of the learnable parameters. Changing these parameters influences the simulated outcome both by changing the continuous evolution and by advancing or delaying switch flips. The flips introduce discrete jumps into the subsequent trajectory that standard derivatives cannot capture. The magnitude of each jump is governed by the nonlinear evolution unfolding after the flip and cannot be deduced from local gradient information at the switch alone.

\begin{figure*}[t]
\centering
\includegraphics[width=0.88\linewidth,trim=0 3 0 34,clip]{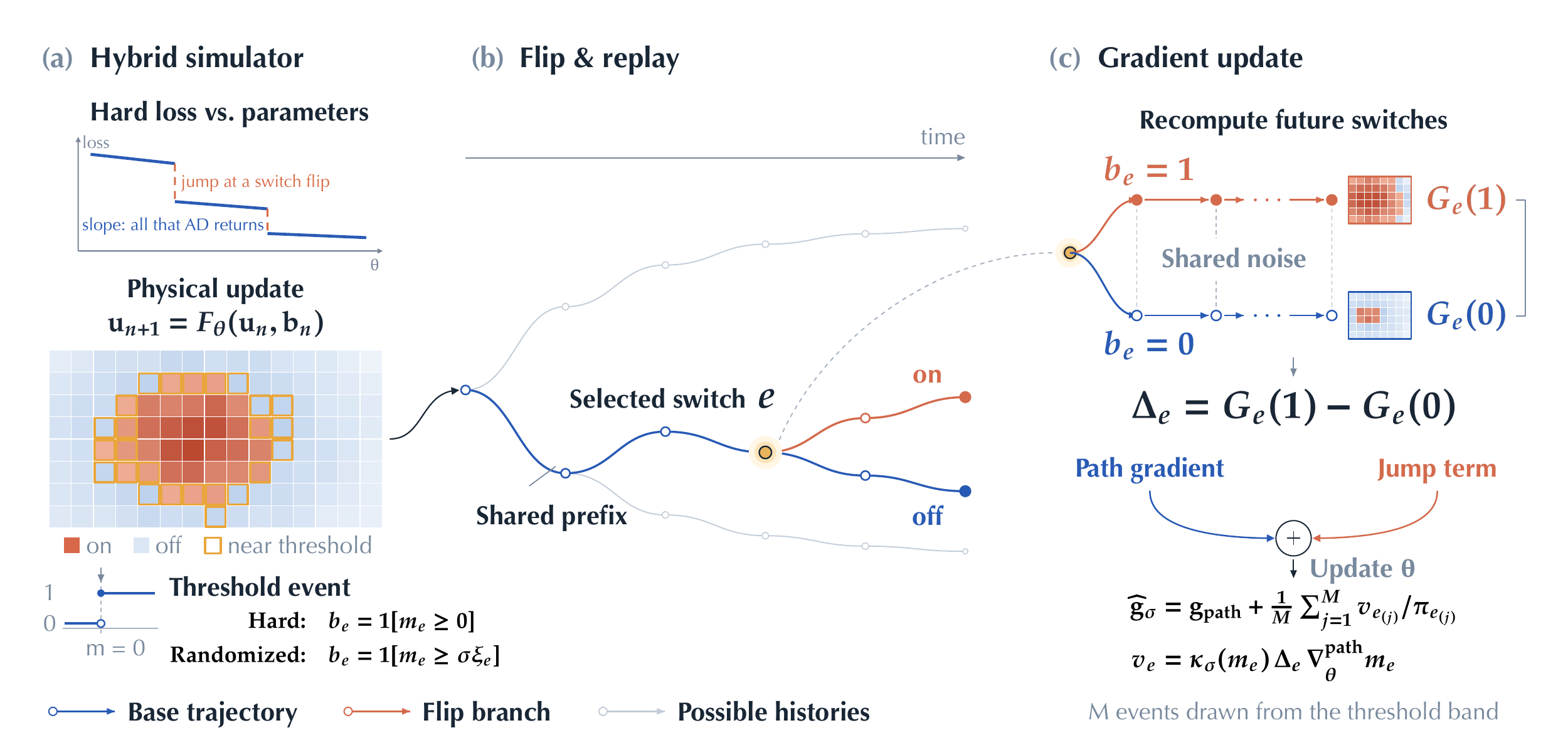}
\caption{\textbf{Overview of \ourmethod{}.} (a) The hard loss exhibits discrete jumps across switch flips, whereas standard automatic differentiation captures only the smooth part. (b) The base trajectory (blue) and the flip branch (orange) share an identical rollout prefix before diverging at a sampled near-threshold switch. (c) Both branches re-evaluate all subsequent switches and arrive at distinct final grid states. The resulting downstream loss difference weights the jump term, which combines with path gradient to form a complete update.}
\label{fig:method-overview}
\end{figure*}

Existing approaches to training hybrid simulators fall into two classes. Gradient-based methods differentiate through the solver. The most direct approach holds every switch at its current value and differentiates the rest of the solver, as in differentiable hydrological models \citep{feng2022differentiable}, and therefore misses every jump of the loss. To recover a gradient at the switch, continuous relaxations replace the threshold with a smooth function such as a sigmoid \citep{tsuyuki1996variational,seo2026tokagrad}, while straight-through estimators keep the hard switch in the forward pass and substitute a smooth derivative in the backward pass \citep{bengio2013estimating}. All of these methods make the gradient computable by simplifying the switch, so the gradient they return is not that of the hard-switch simulator. Derivative-free methods (such as ensemble Kalman inversion, evolution strategies, and black-box reinforcement learning \citep{iglesias2013ensemble,salimans2017evolution,lopez2022training,novati2021automating,von2024closure}) avoid these simplifications by perturbing parameters or sampling actions, but because they treat the entire simulation as a black box, they cannot exploit path gradients of the smooth solver dynamics. Neither class uses the simulator itself to measure the downstream consequence of an individual switch flip.

\ourmethod{}, as outlined in Fig.~\ref{fig:method-overview}, uses interventional differencing to obtain the effect of a switch flip on the loss, which differentiation cannot provide. It restarts the original simulator from a stored state, sets a switch near its threshold to its other value, reruns to the end of the training window, and takes the difference between the two losses, which contains every consequence of the flip within the window. The remaining task is to turn these scalar differences into a gradient with respect to the parameters.

Because the gradient of the hard loss contains no information about the jumps, we perturb every threshold with small Gaussian noise during training. Every switch stays binary in each run, and the hard-switch loss averaged over this noise approaches the original hard loss as the noise vanishes. We show that the gradient of this noise-averaged loss is the sum of a path gradient and a jump term. The path gradient is readily available from automatic differentiation with all switches held fixed. In the jump term, each switch contributes the expected loss difference between its on and off states. This quantity cannot be obtained by differentiation, but the loss difference from interventional differencing is a sample of it. Estimating the expectations in the jump term from these samples and adding the path gradient yields an unbiased estimator of the gradient of the noise-averaged loss. Optimizing the noise-averaged loss with this estimator gives \textsc{\ourmethod-Rand}, whose training trajectories carry threshold noise. Since the deployed simulator runs without noise, \textsc{\ourmethod-Hard} instead carries out the original run and the reruns on the noise-free simulator and adds the same jump term to the path gradient of the hard loss. We further prove that this update targets the complete hard loss, both its continuous variation and its jumps.

Under the same forward budget, we compare \ourmethod{} with twelve baselines on a power-grid cascade, a detonation wave, and cortical spreading depolarization. Each variant of \ourmethod{} achieves lower error than every baseline on all three benchmarks, and the better of the variants lowers the error of the strongest baseline on each by factors of 1.5, 1.4, and 1.4.
We summarize our contributions as follows:
\begin{itemize}[leftmargin=*]
\item \textbf{Formulation and Theory.} We model each hard switch as a random binary decision, so that every jump of the loss becomes a gradient term given by the loss difference between the two outcomes of the switch. Building on smoothed perturbation analysis, we prove that the resulting estimator is unbiased and that its noise-free counterpart still targets the complete hard loss.
\item \textbf{Method.} To our knowledge, \ourmethod{} is the first method that trains hybrid simulators through their hard switches by interventional differencing, in which the simulator is rerun with a flipped switch and the loss difference takes the place of the missing derivative. \textsc{\ourmethod-Rand} trains under the randomized model, while \textsc{\ourmethod-Hard} is a principled variant that trains on the complete hard loss with noise-free simulator that is actually deployed.
\item \textbf{Empirical Validation and Analysis.} \ourmethod{} achieves the lowest error on all three scenarios we benchmarked. Ablation studies and gradient analyses further explain why it succeeds where the baselines fail.
\end{itemize}

\section{Related Work}
\label{sec:related}

\textbf{Hybrid simulators and learned closures.} Training a closure through the solver has improved turbulence closures and atmospheric models \citep{um2020solver,belbute2020combining,list2022learned,frezat2022posteriori,kochkov2021machine,kochkov2024neural,shen2023differentiable}. Derivative-free approaches learn closures with reinforcement learning \citep{novati2021automating,bae2022scientific,kurz2023deep,von2024closure} or ensemble Kalman inversion \citep{dunbar2021calibration,lopez2022training}. None of this work learns from the switch flips of the solver itself. Existing models apply automatic differentiation with every switch held constant \citep{feng2022differentiable}, restructure the model to avoid threshold functions \citep{song2023ancient}, relax the threshold into a sigmoid \citep{seo2026tokagrad}, replace it together with its physics \citep{kochkov2024neural,varbella2024powergraph,poli2021neural,chen2025eqcollide}, or learn a trigger from observations \citep{zhang2021improving}. We keep the hard switches and train on the measured consequences of their flips.

\textbf{Gradients through switches and discontinuities.} Event-time differentiation and saltation matrices handle a single continuous-time event or a fixed event sequence \citep{chen2020learning,kong2024saltation}. Our switches are evaluated once per cell and step, so the step of a flip has no derivative and a flip changes which later switches flip. Randomized smoothing \citep{berthet2020learning} and straight-through estimation \citep{bengio2013estimating,yin2019understanding,liu2023bridging} smooth the discontinuity or replace its derivative. First-order gradients are biased at discontinuities \citep{suh2022differentiable}. Another line of work adds the term that automatic differentiation drops, as edge sampling and reparameterization do for visibility discontinuities in rendering \citep{li2018differentiable,loubet2019reparameterizing,bangaru2020unbiased}. These methods rely on scene geometry to locate each discontinuity in advance. In a solver, which switches flip is known only by running it, so we measure each jump by a rerun.

\textbf{Gradient estimation for discrete and black-box programs.} Policy-gradient methods sample actions \citep{williams1992simple,kool2019buy,schulman2017proximal}, discrete-latent estimators relax the variable or add a score function \citep{jang2017categorical,maddison2017concrete,tucker2017rebar}, and zeroth-order methods perturb the parameters \citep{salimans2017evolution,spall1992multivariate,hansen2001completely,nesterov2017random}. Smoothed perturbation analysis \citep{gong1987smoothed,glasserman1991gradient,fu2012conditional} and local expectation gradients \citep{aueb2015local,tokui2017evaluating,ciosek2020expected} make the effect of one discrete event differentiable by conditioning on part of the randomness or summing exactly over one choice. Stochastic automatic differentiation automates such estimators for discrete random variables and leaves deterministic comparisons as future work \citep{arya2022automatic}, and DiscoGrad smooths conditional branches by perturbing the program inputs \citep{kreikemeyer2023smoothing}. Both execute the program inside their own framework, so an existing solver would have to be rewritten. \ourmethod{} shares the gradient form of this family but needs no such rewriting. It randomizes only the thresholds of deterministic switches, obtains each loss difference by rerunning the existing solver, and reruns only switches in a threshold band. \textsc{\ourmethod-Hard} takes the same measurement on hard-switch trajectories without noise.

\section{Method}
\label{sec:method}

The loss can jump when a hard switch flips, and its local gradient does not capture these jumps. \ourmethod{} flips one switch and continues the simulation, comparing the losses of the two trajectories to measure the effect of this change directly. It combines the measured loss difference and gradients through the continuous dynamics to update the model parameters.

\subsection{Problem Formulation}
\label{sec:setup}

We consider a physical simulator with learnable parameters $\theta$ and hard switches $\vb$, whose state is updated by
\begin{equation}
\vu_{n+1}=F_\theta(\vu_n,\vb_n),\qquad b_{n,i}=\mathbf 1[m_i(\vu_n;\theta)\geq0].
\label{eq:sim}
\end{equation}
Here $\vu_n\in\mathbb R^{N\times C}$ is the state at step $n$, with $N$ cells or components indexed by $i=1,\ldots,N$ and $C$ features. $F_\theta$ is the state update map. 
The vector $\vb_n=(b_{n,i})_{i=1}^{N}$ contains the switch values at step $n$. For cell or component $i$, $b_{n,i}=1$ when the relevant physical quantity reaches or exceeds its threshold, and $b_{n,i}=0$ otherwise. We write this threshold condition in the general parameterized form $m_i(\vu_n;\theta)\geq0$, where $m_i$ is the switch margin. 

The neural network forms the learnable part of the state update $F_\theta$ or the switch margin $m_i(\cdot;\theta)$. When it is part of $F_\theta$, parameter changes affect the physical state and thereby influence later switch decisions. When it is part of $m_i(\cdot;\theta)$, parameter changes directly affect the threshold condition. A switch decision in turn affects subsequent states and may change later switch decisions.

We learn $\theta$ by fitting reference trajectories $\{\vu_n^*\}$. Each training window starts from a reference state and advances the simulator $K$ steps according to Eq.~\eqref{eq:sim}. Summing the prediction errors over these steps gives the window loss of the hard-switch simulator,
$J_{\rm hard}(\theta)=\sum_{n=1}^{K}\ell_n(\vu_n)$.
Here $\ell_n$ measures the error between the predicted state $\vu_n$ and the reference state $\vu_n^*$ and is differentiable in $\vu_n$. We call this window loss the hard loss, and training aims to minimize its average over training windows.

We call each evaluation of one switch at one step an \emph{event}, whether or not its value changes, and index the $N_{\rm evt}$ events of a window by $e$ in temporal order. Event $e$ occurs at step $n_e$ in cell or component $i_e$ with margin $m_e=m_{i_e}(\vu_{n_e};\theta)$ and switch value $b_e$. The values $b_e$ of a trajectory form its \emph{switch history}. The pre-event history $H_e$ consists of the state at which event $e$ is evaluated and all earlier switch values. The trajectory of Eq.~\eqref{eq:sim} is the hard-switch trajectory, and a superscript $h$ marks the quantities on it.

\subsection{Theoretical Foundation}
\label{sec:jump}

The total change in the hard loss $J_{\rm hard}$ as the parameters vary consists of two parts, the smooth change while the switches stay fixed and the jumps caused by switch flips.

When every switch decision stays fixed, the continuous states along the trajectory can still vary with the parameters, which changes the loss smoothly. To compute this change, we hold all switch values fixed and differentiate the remaining continuous computations with respect to the parameters, denoting this total derivative by $\nabla_\theta^{\rm path}$. Applied to the window loss of a simulated trajectory, it gives the \emph{path gradient} $g_{\rm path}$, which is also the gradient returned by automatic differentiation. On the hard-switch trajectory, this gradient is $g_{\rm path}^h=\nabla_\theta^{\rm path}J_{\rm hard}$.

When a switch flips as its event margin crosses zero, the subsequent trajectory can jump, and the loss can jump with it. To describe the consequences of the two switch values of an event $e$, we start from its pre-event state, force its switch to $b\in\{0,1\}$, and run to the end of the window while every later switch is decided again by Eq.~\eqref{eq:sim}. We denote the downstream loss by $G_e^h(b)=\sum_{n>n_e}\ell_n$. The \emph{flip loss difference} is $\Delta_e^h=G_e^h(1)-G_e^h(0)$.

Now consider a continuously differentiable parameter path $\theta(t)$, $t\in[0,1]$. On each interval where the switch history stays fixed, $g_{\rm path}^h\cdot\dot\theta$ is the rate of change of the loss along the path, and every switch flip along the path adds a jump. Under the crossing conditions in Appendix~\ref{app:hard-continuation},
\begin{equation}
\begin{aligned}
&J_{\rm hard}(\theta(1))-J_{\rm hard}(\theta(0)) =\int_0^1 g_{\rm path}^h\cdot\dot\theta\,dt+\sum_{k=1}^{r}\omega_k\,\Delta_{e_k}^h.
\end{aligned}
\label{eq:hard-loss-path}
\end{equation}
Here $r$ is the number of zero crossings along the path, $e_k$ is the event at the $k$-th crossing, and $\Delta_{e_k}^h$ is evaluated at that parameter value. The sign $\omega_k$ is $+1$ when the switch turns on and $-1$ when it turns off. Wherever a small change of $\theta$ flips no switch, $\nabla_\theta J_{\rm hard}=g_{\rm path}^h$, so the gradient of the hard loss contains nothing of the second part.

To let a gradient carry the jumps, we randomize the threshold of every event with small Gaussian noise and define the smoothed objective $J_\sigma$ by
\begin{equation}
\begin{gathered}
b_e=\mathbf 1[m_e\geq\sigma\xi_e],\qquad \xi_e\sim\mathcal N(0,1)\ \text{i.i.d.},\\
J_\sigma(\theta)=\mathbb E_\xi\Bigl[\,\sum_{n=1}^{K}\ell_n(\vu_n)\Bigr].
\end{gathered}
\label{eq:randomized-objective}
\end{equation}
Here $\sigma>0$ is the width, $\xi=(\xi_e)_e$ collects the noise of all events, and the sum is the window loss of the randomized simulator. When no margin of the hard-switch trajectory is zero, $J_\sigma\to J_{\rm hard}$ as $\sigma\to0$ (Proposition~\ref{prop:objective-gap}(b) in Appendix~\ref{app:hard-continuation}). Training with randomized switches minimizes $J_\sigma$, and deployment uses the hard-switch simulator. The switch values stay binary in every randomized run, and the only change is that each switch takes the value $1$ with probability $\Pr(b_e=1\mid H_e)=\Phi(m_e/\sigma)$, where $\Phi$ is the standard normal distribution function. So $J_\sigma(\theta)=\sum_\beta P_\theta(\beta)\,L_\theta(\beta)$, the loss $L_\theta(\beta)$ of each switch history $\beta$ weighted by its probability $P_\theta(\beta)$, and $\theta$ enters both factors.

We call the derivative of the probability $\Phi(m_e/\sigma)$ with respect to the margin the \emph{sensitivity}, $\kappa_\sigma(m)=\phi(m/\sigma)/\sigma$, where $\phi$ is the standard normal density. It is a Gaussian density of width $\sigma$ centered at zero, so it is larger the closer the margin is to zero. The \emph{advantage} of event $e$ is its expected flip loss difference given the history before the event,
\begin{equation}
\mathcal A_e=\mathbb E\bigl[G_e(1)-G_e(0)\mid H_e\bigr].
\label{eq:method-switch-advantage}
\end{equation}
Here $G_e$ and $\Delta_e=G_e(1)-G_e(0)$ are $G_e^h$ and $\Delta_e^h$ with the later switches decided by the randomized rule in Eq.~\eqref{eq:randomized-objective}, and the expectation is over their noise. Following smoothed perturbation analysis \citep{fu2012conditional,arya2022automatic,mohamed2020monte}, we derive and prove the gradient of $J_\sigma$ for switches that are deterministic functions of the state with randomized thresholds (Appendix~\ref{app:paired-consequences}).

\begin{proposition}[Gradient of the smoothed objective]
\label{prop:smoothed-gradient}
Fix a training window and a width $\sigma>0$. Under the regularity conditions of Appendix~\ref{app:finite-histories},
\begin{equation}
\nabla_\theta J_\sigma(\theta)=
\mathbb E_\xi\Bigl[
g_{\rm path}+\sum_{e=1}^{N_{\rm evt}}
\underbrace{\mathcal A_e}_{
  \substack{
    \textnormal{Advantage}\\
    \textnormal{(on vs.\ off)}
  }
}\,
\underbrace{
  \kappa_\sigma(m_e)\,\nabla_\theta^{\rm path}m_e
}_{
  \substack{
    \textnormal{Gradient of probability}\\
    \textnormal{that the switch is on}
  }
}
\Bigr].
\label{eq:grad}
\end{equation}
Here $g_{\rm path}$, $m_e$, and $\mathcal A_e$ are evaluated on the randomized trajectory.
\end{proposition}
The first term, the expected path gradient, is the pathwise term, and the second is the jump term. Each randomized switch is a coin, a Bernoulli policy that is on with probability $\Phi(m_e/\sigma)$. The parameters $\theta$ set the bias of this coin through the margin $m_e$. Each event contributes how far the parameters push its coin toward on, $\nabla_\theta^{\rm path}\Phi(m_e/\sigma)=\kappa_\sigma(m_e)\nabla_\theta^{\rm path}m_e$, times how much more loss the switch gives in expectation when on than when off, $\mathcal A_e$, so the jump term is an expected policy gradient over switch decisions \citep{tokui2017evaluating,ciosek2020expected}.

The path gradient, the sensitivities, and the margin derivatives $\nabla_\theta^{\rm path}m_e$ in Eq.~\eqref{eq:grad} can all be computed on the trajectory that is actually simulated, which we call the \emph{base trajectory}. The remaining quantity in Eq.~\eqref{eq:grad}, the advantage $\mathcal A_e$, is an expectation and must be estimated from samples. The \emph{flip branch} of event $e$ (Fig.~\ref{fig:method-overview}b) restarts the simulator from the state $\vu_{n_e}$ of the base trajectory, sets the switch of event $e$ to $1-b_e$, decides every later switch under the threshold noise of the base trajectory, and runs to the end of the window. This gives a sample of the flip loss difference $\Delta_e=G_e(1)-G_e(0)$, where $G_e(b_e)$ is taken from the base trajectory and $G_e(1-b_e)$ from the flip branch. The conditional expectation of this sample is the advantage, $\mathbb E[\Delta_e\mid H_e,b_e]=\mathcal A_e$. We share the noise between the two runs so that they differ only through this one flip, to lower the variance of the sample without changing its mean (Appendix~\ref{app:variance-surrogate}).

\subsection{\ourmethod{}}
\label{sec:paired-method}

\ourmethod{} has two variants, which differ in whether the training trajectories draw threshold noise. \textsc{EventFlip-Rand} draws the noise as in Eq.~\eqref{eq:randomized-objective} and rolls out the base trajectory.

The flip branch of event $e$ costs $K-n_e$ solver steps, so a branch for every event is too expensive, and each window therefore runs branches only for $M$ sampled events. The base trajectory and the $M$ flip branches together cost at most $(1+M)K$ solver steps per window, all counted in the forward budget. An event whose two values give the same state update has zero flip loss difference, so only the other events, the \emph{eligible} events $\mathcal E_{\rm elig}$, need a branch. The weight of an eligible event in the jump term is its sensitivity $\kappa_\sigma(m_e)$, which is close to zero once the margin is more than a few $\sigma$ away from the threshold. We therefore sample only from the \emph{threshold band} $\mathcal B_\tau=\{e\in\mathcal E_{\rm elig}\mid|m_e|<\tau\sigma\}$, where $\tau$ is a fixed cutoff, and an event of the band is selected with probability
\begin{equation}
\pi_e=(1-\lambda)\,\frac{\kappa_\sigma(m_e)}{\sum_{e'\in\mathcal B_\tau}\kappa_\sigma(m_{e'})}+\frac{\lambda}{|\mathcal B_\tau|},\qquad e\in\mathcal B_\tau.
\label{eq:method-band-proposal}
\end{equation}
Here $\lambda\in(0,1]$ is a fixed uniform fraction. The first part is proportional to the sensitivity because the sensitivity is the weight of the event in the jump term and is also known before any branch runs, whereas the flip loss difference is known only after its branch. The uniform part keeps every selection probability in the band away from zero, so the weights $1/\pi_e$ stay bounded.

For any selection probabilities computed from the base trajectory, the gradient estimate is
\begin{equation}
\widehat g_\sigma=g_{\rm path}+\frac{1}{M}\sum_{j=1}^{M}\frac{\Delta_{e_{(j)}}\,\kappa_\sigma(m_{e_{(j)}})\,\nabla_\theta^{\rm path}m_{e_{(j)}}}{\pi_{e_{(j)}}}.
\label{eq:method-full-estimator}
\end{equation}
Here the events $e_{(1)},\ldots,e_{(M)}$ are drawn from $\pi$ independently with replacement given $\xi$, and $v_e=\kappa_\sigma(m_e)\,\Delta_{e}\,\nabla_\theta^{\rm path}m_{e}$ is the event vector. The estimate differs from Eq.~\eqref{eq:grad} in two places, where the sample $\Delta_e$ measured by the flip branch replaces the advantage and the average over the $M$ sampled events replaces the sum over all events. Dividing by $\pi_e$ makes the expected average over the sampled events equal to the sum of $v_e$ over every event with positive selection probability.

\begin{proposition}[Unbiasedness under full-support event sampling]
\label{prop:full-support-unbiased}
Let the flip branches share the later threshold noise of the base trajectory. Let the selection probabilities in Eq.~\eqref{eq:method-full-estimator} depend only on the base trajectory and be positive for every eligible event, which we call full-support event sampling. Then, under the regularity conditions of Appendix~\ref{app:finite-histories}, the estimator is unbiased for every integer $M\geq1$, that is,
\begin{equation}
    \mathbb E[\widehat g_\sigma]
    = \nabla_\theta J_\sigma(\theta).
\end{equation}
\end{proposition}
The rule in Eq.~\eqref{eq:method-band-proposal} gives positive probability only inside the threshold band, so the eligible events outside the band are never drawn. However, their weight $\kappa_\sigma(m_e)$ is close to zero, so omitting them leaves a small bias. The expected update of \textsc{EventFlip-Rand} differs from $\nabla_\theta J_\sigma$ by a bias whose bound decays as $e^{-\tau^2/2}$, and this bound holds however many switches are near their thresholds (Appendix~\ref{app:paired-consequences} proves Proposition~\ref{prop:full-support-unbiased}, Appendix~\ref{app:sampling} proves the bound, and Appendix~\ref{app:fidelity} measures the bias).

\label{sec:hard-continuation}\textsc{EventFlip-Rand} trains with threshold noise, and deployment has none. \textsc{\ourmethod-Hard} trains on the hard-switch simulator, the one that is deployed. It draws no threshold noise, which sets $\xi\equiv0$, and still uses the estimate in Eq.~\eqref{eq:method-full-estimator}. This is possible because every factor of Eq.~\eqref{eq:method-full-estimator} can be computed on the simulator without noise. Its base trajectory is the hard-switch trajectory, on which the path gradient $g_{\rm path}^h$ and the margin derivatives $\nabla_\theta^{\rm path}m_e^h$ are computed as before. The flip branches decide the later switches by Eq.~\eqref{eq:sim}, so they measure the flip loss difference $\Delta_e^h$ of the hard-switch simulator. The sensitivity and the selection probabilities are functions of the margins and $\sigma$ only. Here $\sigma$ is no longer the size of a noise, and as a width it sets how close to the threshold a margin must be for its event to enter the threshold band and how fast the weight of an event in the band decays with this distance.

Averaging Eq.~\eqref{eq:method-full-estimator} over the event draws gives the expected update of \textsc{\ourmethod-Hard},
\begin{equation}
g_{\rm hard,band}=g_{\rm path}^{h}+\sum_{e\in\mathcal B_\tau^h}\kappa_\sigma(m_e^h)\,\Delta_e^h\,\nabla_\theta^{\rm path}m_e^h.
\label{eq:hard-update}
\end{equation}
Its two parts are its pathwise term and its jump term. The update in Eq.~\eqref{eq:hard-update} is computed entirely on the hard-switch simulator, so we compare its integral along a parameter path with the change of the hard loss in Eq.~\eqref{eq:hard-loss-path}.

\begin{proposition}[Consistency of \textsc{\ourmethod-Hard} with the hard loss]
\label{prop:hard-jumps}
Let $\theta(t)$ be a parameter path as in Eq.~\eqref{eq:hard-loss-path}, under the conditions of Appendix~\ref{app:hard-continuation}. The expected update in Eq.~\eqref{eq:hard-update} of {\normalfont\textsc{\ourmethod-Hard}} satisfies
\begin{align}
&\lim_{\sigma\to0}\int_0^1 g_{\rm hard,band}\cdot\dot\theta\,dt =\int_0^1 g_{\rm path}^h\cdot\dot\theta\,dt
+\bigl(2\Phi(\tau)-1\bigr)\sum_{k=1}^{r}\omega_k\,\Delta_{e_k}^h.
\label{eq:consistency}
\end{align}
\end{proposition}
The limit has the same two parts as Eq.~\eqref{eq:hard-loss-path}, with the extra factor $2\Phi(\tau)-1$ on the jumps. This factor is the mass of the sensitivity $\kappa_\sigma$, a density of total mass one, inside the threshold band, and it differs from $1$ by at most $e^{-\tau^2/2}$. The update of \textsc{\ourmethod-Hard} therefore targets the complete hard loss, including the jumps that its gradient misses. The same integral of $\nabla_\theta J_\sigma$ tends to the full change of the hard loss in Eq.~\eqref{eq:hard-loss-path} as $\sigma\to0$ (Appendix~\ref{app:hard-continuation}). So \textsc{EventFlip-Rand} carries the jumps in its training objective $J_\sigma$, and \textsc{\ourmethod-Hard} in its update. Algorithm~\ref{alg:eventflip} in Appendix~\ref{app:variance-surrogate} gives training updates of both variants on one window.

\section{Experiments}
\label{sec:experiments}

\subsection{\texorpdfstring{Experimental Setup}{Experimental Setup}}
\label{sec:exp-setup}

\textbf{Benchmarks.} We evaluate our method across three physical benchmarks featuring distinct switching regimes (Table~\ref{tab:properties-full}). In each benchmark, an embedded neural closure parameterizes unresolved physical processes. The \emph{power-grid cascade} simulates a cascading failure on the IEEE 118-bus transmission network, where any line whose power flow exceeds its rating (maximum allowable flow) permanently trips. The closure predicts these latent line ratings without direct supervision; because the post-trip power flow depends solely on the network topology, the path gradient vanishes identically. The \emph{detonation wave} simulates a one-dimensional combustion wave propagating through unburnt gas on a coarse grid \citep{bao2000random}, where a computational cell ignites once its temperature surpasses a critical ignition threshold. Following standard learned-closure formulations \citep{um2020solver,kochkov2021machine}, the network corrects inter-cell numerical fluxes to match a fine-grid reference, with the dominant learning signal propagating through the path gradient. \emph{Cortical spreading depolarization} models post-stroke wave propagation across brain tissue \citep{dreier2015stroke} via the reaction-diffusion of extracellular K$^+$ on a coarse $64\times64$ grid \citep{tuckwell1978mathematical,nicholson1998extracellular}. A cell irreversibly activates once its local K$^+$ concentration crosses an excitation threshold. The domain comprises an unexcitable infarct core surrounded by a vulnerable penumbra with a reduced activation threshold. The closure similarly predicts inter-cell K$^+$ flux corrections.

\textbf{Training configurations.} The twelve baselines fall into two groups (Appendix~\ref{app:baselines}). Four are gradient-based and differentiate through the solver, namely direct differentiation (Direct) \citep{um2020solver,feng2022differentiable}, sigmoid relaxation (Sigmoid) \citep{seo2026tokagrad}, straight-through estimation (STE) \citep{bengio2013estimating}, and AoBG \citep{suh2022differentiable}. The other eight are derivative-free and use only rollouts of the solver, including ES \citep{salimans2017evolution}, SPSA \citep{spall1992multivariate}, CMA-ES \citep{hansen2001completely}, EKI \citep{iglesias2013ensemble}, RLOO \citep{kool2019buy}, GRPO-clip \citep{shao2024deepseekmath}, PPO \citep{schulman2017proximal}, and Closure-RL \citep{von2024closure}, which does not apply to the power grid. All methods are trained under an identical forward evaluation budget (750k simulator steps on the power grid and 800k on the remaining two benchmarks), accounting for all flip branches, ensemble members, and perturbed rollouts (Appendix~\ref{app:protocol}). Table~\ref{tab:baseline-config} lists the configurations of all methods.

\textbf{Evaluation.} We assess performance using a primary metric per benchmark, namely the rollout loss over line flows and shed load for the power-grid cascade, the relative wave-speed error for the detonation wave, and the normalized activation-time error for cortical spreading depolarization, defined as the mean absolute error in first-activation time normalized by the observation span. At test time, all trained closures are deployed inside the deterministic hard-switch simulator (Appendix~\ref{app:protocol}). Each method is trained across three random seeds, and test performance is reported at the checkpoint with the lowest validation error on the primary metric.

\subsection{\texorpdfstring{Main Results}{Main Results}}
\label{sec:exp-accuracy}

\begin{table}[t]
\centering
\caption{\textbf{Test error on the primary metric across benchmarks}, reported as the mean across three random seeds with standard deviations shown as subscripts. Bold, underlined and italicized entries denote the lowest, second-lowest and third-lowest errors. Gray entries indicate runs whose error remains within 10\% of the untrained closure.}
\label{tab:main}
\small
\setlength{\tabcolsep}{8pt}
\begin{tabular}{@{}lccc@{}}
\toprule
Scenario & \shortstack[c]{Power-grid\\cascade} & \shortstack[c]{Detonation\\wave} & \shortstack[c]{Cortical spreading\\depolarization} \\
Metric & \shortstack[c]{Rollout\\loss} & \shortstack[c]{Wave-speed\\error (\%)} & \shortstack[c]{Activation-time\\error (\%)} \\
\midrule
\textsc{EventFlip-Hard} & $\mathbf{0.067}_{\pm 0.007}$ & $\mathbf{0.104}_{\pm 0.011}$ & $\underline{0.793}_{\pm 0.004}$ \\
\textsc{EventFlip-Rand} & $\underline{0.072}_{\pm 0.013}$ & $\underline{0.120}_{\pm 0.010}$ & $\mathbf{0.759}_{\pm 0.014}$ \\
\midrule
Direct~\citep{um2020solver,feng2022differentiable} & \textcolor{black!45}{$0.181_{\pm 0.000}$} & $0.641_{\pm 0.319}$ & $1.068_{\pm 0.097}$ \\
Sigmoid~\citep{seo2026tokagrad} & $0.144_{\pm 0.027}$ & {$\mathit{0.150}_{\pm 0.011}$} & \textcolor{black!45}{$13.319_{\pm 0.253}$} \\
STE~\citep{bengio2013estimating} & \textcolor{black!45}{$0.179_{\pm 0.001}$} & \textcolor{black!45}{$12.945_{\pm 0.230}$} & \textcolor{black!45}{$13.319_{\pm 0.181}$} \\
AoBG~\citep{suh2022differentiable} & $0.143_{\pm 0.027}$ & $0.411_{\pm 0.038}$ & $1.248_{\pm 0.076}$ \\
\midrule
ES~\citep{salimans2017evolution} & $0.116_{\pm 0.008}$ & \textcolor{black!45}{$13.188_{\pm 0.109}$} & $2.302_{\pm 0.735}$ \\
SPSA~\citep{spall1992multivariate} & \textcolor{black!45}{$0.177_{\pm 0.008}$} & \textcolor{black!45}{$13.030_{\pm 0.187}$} & $4.245_{\pm 2.162}$ \\
CMA-ES~\citep{hansen2001completely} & {$\mathit{0.101}_{\pm 0.010}$} & \textcolor{black!45}{$13.016_{\pm 0.105}$} & $9.371_{\pm 0.237}$ \\
EKI~\citep{iglesias2013ensemble} & $0.145_{\pm 0.018}$ & \textcolor{black!45}{$13.677_{\pm 0.188}$} & $1.237_{\pm 0.104}$ \\
RLOO~\citep{kool2019buy} & $0.118_{\pm 0.004}$ & \textcolor{black!45}{$12.734_{\pm 0.252}$} & $11.563_{\pm 1.336}$ \\
GRPO-clip~\citep{shao2024deepseekmath} & $0.136_{\pm 0.021}$ & \textcolor{black!45}{$13.023_{\pm 0.247}$} & $8.462_{\pm 0.364}$ \\
PPO~\citep{schulman2017proximal} & $0.143_{\pm 0.009}$ & \textcolor{black!45}{$12.915_{\pm 0.015}$} & \textcolor{black!45}{$13.390_{\pm 0.110}$} \\
Closure-RL~\citep{von2024closure} & n/a & \textcolor{black!45}{$13.360_{\pm 2.342}$} & $\mathit{1.050}_{\pm 0.133}$ \\
\midrule
Untrained closure & $0.181$ & $13.569$ & $13.465$ \\
\bottomrule
\end{tabular}
\end{table}

Both variants of \ourmethod{} consistently achieve the lowest error across all three benchmarks (Table~\ref{tab:main}).
Against the strongest baseline in each regime (namely CMA-ES on power grid, sigmoid relaxation on detonation wave, and Closure-RL on cortical spreading depolarization) our top-performing variant reduces error by factors of 1.5, 1.4, and 1.4, respectively.
Notably, existing baselines fail to generalize across different switching mechanisms, as the strongest baseline in one benchmark often fails in another, whereas \ourmethod{} ranks first across all three.

\begin{figure}[t]
\centering
\figfile[width=0.62\linewidth]{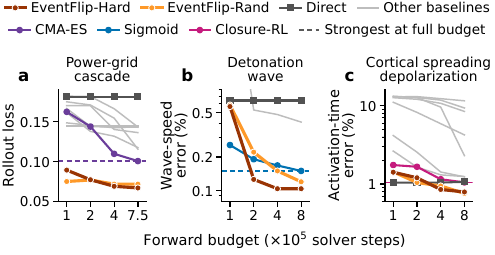}
\caption{\textbf{Test error against forward budget}, evaluated using the best validation checkpoint at each budget level (three-seed mean). Horizontal dashed lines indicate the full-budget error of the strongest baseline in each benchmark. Additional baselines are plotted in gray, with failed runs omitted for clarity.}
\label{fig:budget}
\end{figure}

On the power grid, the lower rollout loss of \ourmethod{} means more faithful predictions of cascading dynamics. Specifically, \textsc{\ourmethod-Hard} reduces the mean shed-load error to 0.63\% of the system load, compared with 1.03\% for the strongest baseline, CMA-ES (Appendix~\ref{app:full-results}), and lowers the rollout loss by 38\% on severe cascades (Appendix~\ref{app:severity}). In the representative case shown in Fig.~\ref{fig:teaser}, \textsc{\ourmethod-Hard} recovers the reference cascade trajectory, whereas CMA-ES and direct differentiation trigger no line trips.

On the detonation wave, CMA-ES fails to learn beyond initialization, yielding a wave-speed error of 13.016\% compared with 13.569\% for untrained closure, leaving sigmoid relaxation as the strongest baseline. This advantage also extends to the post-shock flow field, where \textsc{EventFlip-Rand} achieves peak-pressure errors of 0.92\%, outperforming sigmoid relaxation at 1.77\% (Appendix~\ref{app:full-results}).

On cortical spreading depolarization, sigmoid relaxation in turn stays near initialization, yielding a nearly untrained error, leaving Closure-RL and direct differentiation as the closest baselines, whereas Closure-RL itself fails to train on the detonation wave. Our performance gain also holds within the penumbra, where accurately capturing wave invasion is clinically critical, with \textsc{\ourmethod-Hard} and \textsc{EventFlip-Rand} achieving activation-time errors of 0.522\% and 0.544\%, respectively, outperforming direct differentiation at 0.651\% and Closure-RL at 1.055\% (Appendix~\ref{app:full-results}). As visualized in Fig.~\ref{fig:teaser}, \textsc{EventFlip} keeps the activation-time error within two time steps across most of the tissue domain, whereas other baselines exhibit widespread temporal drift.

While the comparisons above evaluate performance at the full forward budget, \ourmethod{} establishes a clear lead earlier in training (Fig.~\ref{fig:budget}). On power grid, both variants already achieve lower loss than the full-budget performance of every baseline within the first 100k simulator steps. Similarly, on detonation wave, our top-performing variant matches the full-budget error of sigmoid relaxation using only 1/4 of the solver budget. These sample-efficiency gains confirm that the advantage of \ourmethod{} stems from the fidelity of its gradient estimator rather than increased computation.

\subsection{\texorpdfstring{Ablation and Mechanism Analysis}{Ablation and Mechanism Analysis}}
\label{sec:exp-mechanism}

\begin{table}[t]
\centering
\caption{\textbf{Test error on ablation study.} Test error on the primary metrics (mean across three random seeds). On power grid, the jump-only variants reduce to the full methods ($=$).}
\label{tab:mechanism}
\small
\setlength{\tabcolsep}{8pt}
\resizebox{\ifdim\width>\linewidth\linewidth\else\width\fi}{!}{%
\begin{tabular}{@{}lccc@{}}
\toprule
 & \shortstack[c]{Power-grid\\cascade} & \shortstack[c]{Detonation\\wave} & \shortstack[c]{Cortical spreading\\depolarization} \\
\midrule
Direct (path gradient only) & 0.181 & 0.641 & 1.068 \\
\textsc{EventFlip-Hard}, jump term only & $=$ & 4.432 & 0.883 \\
\textsc{EventFlip-Rand}, jump term only & $=$ & 4.322 & 0.858 \\
\textsc{EventFlip-Hard} & 0.067 & 0.104 & 0.793 \\
\textsc{EventFlip-Rand} & 0.072 & 0.120 & 0.759 \\
\midrule
\textsc{EventFlip-Rand}, independent noise & 0.072 & 0.132 & 1.101 \\
\textsc{EventFlip-Rand}, no flip branch & 0.064 & 0.169 & 11.720 \\
\midrule
Near-threshold switches per window & 1.0--1.3 & 3.4--8.0 & 347--374 \\
\bottomrule
\end{tabular}}
\end{table}

\textbf{Gradient decomposition.} Proposition~\ref{prop:smoothed-gradient} decomposes the gradient of the smoothed objective into a continuous path gradient and a discrete jump term, as mirrored by \textsc{\ourmethod-Hard} update in Eq.~\eqref{eq:hard-update}. To isolate their roles, we train with the path gradient alone (direct differentiation), the jump term alone, and their combination (upper block of Table~\ref{tab:mechanism}). On power grid, where the path gradient vanishes identically, direct differentiation receives zero learning signal and stays at the untrained loss, whereas incorporating the jump term reduces the loss to 0.067 for \textsc{\ourmethod-Hard}. Consequently, direct differentiation recovers only 8.5\% of the reference line trips, compared with 71\% for \textsc{\ourmethod-Hard} (Appendix~\ref{app:timing}). Because the closure directly parameterizes line ratings whose ground-truth is known in this benchmark, we can further verify the underlying learning mechanism. Across line-case pairs loaded to at least 80\% of capacity, \textsc{\ourmethod-Hard} reduces the RMSE of the predicted logarithmic rating from untrained 0.211 to 0.026, and from 0.186 to 0.046--0.051 across all pairs (Appendix~\ref{app:ratings}), confirming that the jump term recovers the switching thresholds.

In the remaining two scenarios, where the path gradient is nonzero, we examine whether the jump term remains necessary and whether it suffices in isolation. On the detonation wave, neither component is sufficient on its own, yielding wave-speed errors of 0.641\% without the jump term and 4.3--4.4\% without the path gradient, compared with approximately 0.1\% when both are combined. Cortical spreading depolarization exhibits the same answer, with both terms together achieving the lowest error, whereas omitting the jump term degrades accuracy more severely than omitting the path gradient, increasing the error to 1.068\%.

\textbf{Flip-branch measurement.} Unlike the path gradient, the jump term cannot be computed via automatic differentiation and must instead be estimated from downstream loss difference across the two states of each switch. \ourmethod{} isolates this effect using a flip branch that diverges from the base trajectory solely at the perturbed switch (Section~\ref{sec:jump}). To quantify the importance of this counterfactual rollout, we ablate this design in two stages (lower block of Table~\ref{tab:mechanism}). Both ablations preserve the expected gradient update while inflating its variance. First, we sample independent threshold noise along the flip branch rather than sharing it with the base trajectory. The resulting variance inflation scales with the number of near-threshold switches per training window, as uncorrelated noise introduces extraneous stochasticity across all remaining near-threshold events. On power-grid cascade and detonation wave, which average only 1 to 8 near-threshold switches per window, jump-term variance and test errors remain largely unaffected (Appendix~\ref{app:common-noise}). By contrast, on cortical spreading depolarization, where each window contains about 350 near-threshold switches, independent noise inflates gradient variance by 9$\times$ at the trained closure, degrading the error from 0.759\% to 1.101\%. Thus, despite remaining unbiased, the estimator loses nearly all gains from the jump term due to variance explosion. Conversely, \textsc{\ourmethod-Hard} eliminates threshold noise altogether, ensuring that the base and flip trajectories diverge solely at the target switch and achieving a low error of 0.793\%.

Second, we remove flip branches entirely, estimating each switch advantage solely from the downstream loss of the base trajectory, centered by its mean across the remaining minibatch windows. This REINFORCE-style ablation \citep{williams1992simple} is detailed in Appendix~\ref{app:switch-reinforce}. However, raw downstream losses vary substantially across trajectories and confound the stochastic outcomes of all concurrent switches. Under the same forward budget, this credit-assignment noise inflates the jump-term variance at initialization on cortical spreading depolarization by 4{,}681$\times$ relative to the full method. Consequently, test error degrades sharply with the number of near-threshold switches. While accuracy remains comparable on power grid at 0.064 versus 0.072, the error rises from 0.120\% to 0.169\% on detonation wave and stalls at 11.720\% on cortical spreading depolarization. These results demonstrate that unbiasedness alone is insufficient without low-variance credit assignment. By perturbing one switch along an otherwise identical rollout, each flip isolates the causal consequence of that event, enabling \ourmethod{} to scale from one switch to hundreds of concurrent switches.

\textbf{Gradient fidelity.} While the preceding evaluations measure post-training accuracy, we now directly assess the fidelity of the gradient updates themselves. By Proposition~\ref{prop:full-support-unbiased}, under full-support event sampling \textsc{EventFlip-Rand} provides an unbiased gradient estimator for the smoothed objective. To check the gradient in practice, we project the gradients returned in the training configuration onto probe directions and compare them against finite-difference directional derivatives of the smoothed objective (Appendix~\ref{app:fidelity}). Across all three benchmarks and at both untrained and trained checkpoints, the directional derivatives returned by \textsc{EventFlip-Rand} remain within 0.34--1.2 times of the reference (Table~\ref{tab:fidelity-ratio}). Applying the same diagnostic to sigmoid relaxation and straight-through estimation reveals severe gradient distortion. Although sigmoid relaxation trains successfully on detonation wave where switches are re-evaluated at every step, on cortical spreading depolarization, where an activated cell raises its own threshold, both baselines overestimate directional derivatives by $10^4$--$10^7$ times at the trained closure, and neither trains there. Even on detonation wave, straight-through estimation inflates gradients by $10^2$--$10^5$ times at the trained closure, leaving our \textsc{EventFlip-Rand} as the only method whose gradients consistently stay within a factor of three of the reference across all benchmarks.

\section{\texorpdfstring{Discussion and Conclusion}{Discussion and Conclusion}}
\label{sec:discussion}
\label{sec:conclusion}

We showed that hybrid physical simulators with hard threshold switches can be trained end-to-end by querying the native solver to measure the causal consequences of switch flips. These measured jumps recover the discontinuous gradients omitted by standard automatic differentiation, and perturbing one switch at a time preserves low-variance credit assignment even when hundreds of events lie near thresholds.

Several promising directions remain for future work. First, because each flip branch currently re-executes the solver to the end of the training window, exploiting spatiotemporal locality to reuse unaffected computations from the base trajectory could substantially reduce rollout overhead. Second, establishing formal convergence guarantees for \textsc{\ourmethod-Hard}, which couples the pathwise gradient of the hard loss $J_{\rm hard}$ with deterministic flip loss differences (Appendix~\ref{app:hard-continuation}), remains an open theoretical question. Finally, we plan to scale \ourmethod{} to more complex solvers and higher-capacity neural closures.

\subsubsection*{LLM Usage Statement}
We used Codex (OpenAI) for two purposes. In the writing, it polished the language of the manuscript text. In the research, it assisted the literature search for related work, and the authors read and verified every cited paper. It was not used for scientific ideation, methodology development, or experimental data collection. The authors have reviewed all AI-assisted content and take full responsibility for the final manuscript.

\subsubsection*{Acknowledgments}
This work was supported by the Fundamental and Interdisciplinary Disciplines Breakthrough Plan of the Ministry of Education of China (JYB2025XDXIM801) and the National Natural Science Foundation of China (42125503 and 42430602).

\subsubsection*{Ethics Statement}
This work develops a training method for hybrid physical simulators and evaluates it on numerical benchmarks of a cascading failure in a power grid, a detonation wave, and cortical spreading depolarization in brain tissue. It involves no human participants or personal data. Learned simulators of combustion and power systems can inform safety assessments, and those of brain tissue can support mechanistic studies of spreading depolarization. Their use in such settings requires validation against reference solvers and measurements.

\subsubsection*{Reproducibility Statement}
Section~\ref{sec:method} and Algorithm~\ref{alg:eventflip} specify the estimator and the training update. Appendix~\ref{app:derivation} gives the complete derivations and proofs, including the conditions of Proposition~\ref{prop:smoothed-gradient} and Proposition~\ref{prop:full-support-unbiased} and the bound of Proposition~\ref{prop:truncation-bias}. Appendix~\ref{app:exp-details} documents all three simulators, the closure architectures, the configuration of \ourmethod{}, the forward budgets, the training and evaluation protocol, and the configuration of every baseline. Appendix~\ref{app:extra-results} reports the full results on the three simulators. Source code and training configurations are available at \url{https://github.com/gaotianrun1/pderl}.

\bibliography{refs}

\begin{thebibliography}{66}
\providecommand{\natexlab}[1]{#1}
\providecommand{\url}[1]{\texttt{#1}}
\expandafter\ifx\csname urlstyle\endcsname\relax
  \providecommand{\doi}[1]{doi: #1}\else
  \providecommand{\doi}{doi: \begingroup \urlstyle{rm}\Url}\fi

\bibitem[Arya et~al.(2022)Arya, Schauer, Sch{\"a}fer, and Rackauckas]{arya2022automatic}
Gaurav Arya, Moritz Schauer, Frank Sch{\"a}fer, and Christopher Rackauckas.
\newblock Automatic differentiation of programs with discrete randomness.
\newblock \emph{Advances in Neural Information Processing Systems}, 35:\penalty0 10435--10447, 2022.

\bibitem[Babaeinejadsarookolaee et~al.(2019)Babaeinejadsarookolaee, Birchfield, Christie, Coffrin, DeMarco, Diao, Ferris, Fliscounakis, Greene, Huang, et~al.]{babaeinejadsarookolaee2019power}
Sogol Babaeinejadsarookolaee, Adam Birchfield, Richard~D Christie, Carleton Coffrin, Christopher DeMarco, Ruisheng Diao, Michael Ferris, Stephane Fliscounakis, Scott Greene, Renke Huang, et~al.
\newblock The power grid library for benchmarking {AC} optimal power flow algorithms.
\newblock \emph{arXiv preprint arXiv:1908.02788}, 2019.

\bibitem[Bae and Koumoutsakos(2022)]{bae2022scientific}
H~Jane Bae and Petros Koumoutsakos.
\newblock Scientific multi-agent reinforcement learning for wall-models of turbulent flows.
\newblock \emph{Nature Communications}, 13\penalty0 (1):\penalty0 1443, 2022.

\bibitem[Bangaru et~al.(2020)Bangaru, Li, and Durand]{bangaru2020unbiased}
Sai~Praveen Bangaru, Tzu-Mao Li, and Fr{\'e}do Durand.
\newblock Unbiased warped-area sampling for differentiable rendering.
\newblock \emph{ACM Transactions on Graphics}, 39\penalty0 (6):\penalty0 1--18, 2020.
\newblock \doi{10.1145/3414685.3417833}.

\bibitem[Bao and Jin(2000)]{bao2000random}
Weizhu Bao and Shi Jin.
\newblock The random projection method for hyperbolic conservation laws with stiff reaction terms.
\newblock \emph{Journal of Computational Physics}, 163\penalty0 (1):\penalty0 216--248, 2000.

\bibitem[Belbute-Peres et~al.(2020)Belbute-Peres, Economon, and Kolter]{belbute2020combining}
Filipe De~Avila Belbute-Peres, Thomas Economon, and Zico Kolter.
\newblock Combining differentiable {PDE} solvers and graph neural networks for fluid flow prediction.
\newblock In \emph{international conference on machine learning}, pages 2402--2411. PMLR, 2020.

\bibitem[Bengio et~al.(2013)Bengio, L{\'e}onard, and Courville]{bengio2013estimating}
Yoshua Bengio, Nicholas L{\'e}onard, and Aaron Courville.
\newblock Estimating or propagating gradients through stochastic neurons for conditional computation.
\newblock \emph{arXiv preprint arXiv:1308.3432}, 2013.

\bibitem[Bernstein et~al.(2014)Bernstein, Bienstock, Hay, Uzunoglu, and Zussman]{bernstein2014power}
Andrey Bernstein, Daniel Bienstock, David Hay, Meric Uzunoglu, and Gil Zussman.
\newblock Power grid vulnerability to geographically correlated failures—analysis and control implications.
\newblock In \emph{IEEE INFOCOM 2014-IEEE Conference on Computer Communications}, pages 2634--2642. IEEE, 2014.

\bibitem[Berthet et~al.(2020)Berthet, Blondel, Teboul, Cuturi, Vert, and Bach]{berthet2020learning}
Quentin Berthet, Mathieu Blondel, Olivier Teboul, Marco Cuturi, Jean-Philippe Vert, and Francis Bach.
\newblock Learning with differentiable perturbed optimizers.
\newblock \emph{Advances in neural information processing systems}, 33:\penalty0 9508--9519, 2020.

\bibitem[Chen et~al.(2025)Chen, Gao, Jiang, and Wu]{chen2025eqcollide}
Qianyi Chen, Tianrun Gao, Chenbo Jiang, and Tailin Wu.
\newblock Eqcollide: Equivariant and collision-aware deformable objects neural simulator.
\newblock \emph{arXiv preprint arXiv:2506.05797}, 2025.

\bibitem[Chen et~al.(2020)Chen, Amos, and Nickel]{chen2020learning}
Ricky~TQ Chen, Brandon Amos, and Maximilian Nickel.
\newblock Learning neural event functions for ordinary differential equations.
\newblock \emph{arXiv preprint arXiv:2011.03902}, 2020.

\bibitem[Ciosek and Whiteson(2020)]{ciosek2020expected}
Kamil Ciosek and Shimon Whiteson.
\newblock Expected policy gradients for reinforcement learning.
\newblock \emph{Journal of Machine Learning Research}, 21\penalty0 (52):\penalty0 1--51, 2020.

\bibitem[Dreier(2011)]{dreier2011role}
Jens~P Dreier.
\newblock The role of spreading depression, spreading depolarization and spreading ischemia in neurological disease.
\newblock \emph{Nature medicine}, 17\penalty0 (4):\penalty0 439--447, 2011.

\bibitem[Dreier and Reiffurth(2015)]{dreier2015stroke}
Jens~P Dreier and Clemens Reiffurth.
\newblock The stroke-migraine depolarization continuum.
\newblock \emph{Neuron}, 86\penalty0 (4):\penalty0 902--922, 2015.

\bibitem[Dunbar et~al.(2021)Dunbar, Garbuno-Inigo, Schneider, and Stuart]{dunbar2021calibration}
Oliver~RA Dunbar, Alfredo Garbuno-Inigo, Tapio Schneider, and Andrew~M Stuart.
\newblock Calibration and uncertainty quantification of convective parameters in an idealized {GCM}.
\newblock \emph{Journal of Advances in Modeling Earth Systems}, 13\penalty0 (9):\penalty0 e2020MS002454, 2021.

\bibitem[Feng et~al.(2022)Feng, Liu, Lawson, and Shen]{feng2022differentiable}
Dapeng Feng, Jiangtao Liu, Kathryn Lawson, and Chaopeng Shen.
\newblock Differentiable, learnable, regionalized process-based models with multiphysical outputs can approach state-of-the-art hydrologic prediction accuracy.
\newblock \emph{Water Resources Research}, 58\penalty0 (10):\penalty0 e2022WR032404, 2022.

\bibitem[Frezat et~al.(2022)Frezat, Le~Sommer, Fablet, Balarac, and Lguensat]{frezat2022posteriori}
Hugo Frezat, Julien Le~Sommer, Ronan Fablet, Guillaume Balarac, and Redouane Lguensat.
\newblock A posteriori learning for quasi-geostrophic turbulence parametrization.
\newblock \emph{Journal of Advances in Modeling Earth Systems}, 14\penalty0 (11):\penalty0 e2022MS003124, 2022.

\bibitem[Fu and Hu(2012)]{fu2012conditional}
Michael~C Fu and Jian-Qiang Hu.
\newblock \emph{Conditional {M}onte {C}arlo: Gradient estimation and optimization applications}, volume 392.
\newblock Springer Science \& Business Media, 2012.

\bibitem[Glasserman(1991)]{glasserman1991gradient}
Paul Glasserman.
\newblock \emph{Gradient estimation via perturbation analysis}, volume 116.
\newblock Springer Science \& Business Media, 1991.

\bibitem[Gong and Ho(1987)]{gong1987smoothed}
Wei-Bo Gong and Yu-Chi Ho.
\newblock Smoothed (conditional) perturbation analysis of discrete event dynamical systems.
\newblock \emph{IEEE Transactions on Automatic Control}, 32\penalty0 (10):\penalty0 858--866, 1987.

\bibitem[Hansen and Ostermeier(2001)]{hansen2001completely}
Nikolaus Hansen and Andreas Ostermeier.
\newblock Completely derandomized self-adaptation in evolution strategies.
\newblock \emph{Evolutionary computation}, 9\penalty0 (2):\penalty0 159--195, 2001.

\bibitem[{IEEE}(2013)]{ieee2012std738}
{IEEE}.
\newblock {IEEE} standard for calculating the current-temperature relationship of bare overhead conductors.
\newblock \emph{IEEE Std 738-2012 (Revision of IEEE Std 738-2006 - Incorporates IEEE Std 738-2012 Cor 1-2013)}, pages 1--72, 2013.
\newblock \doi{10.1109/IEEESTD.2013.6692858}.

\bibitem[Iglesias et~al.(2013)Iglesias, Law, and Stuart]{iglesias2013ensemble}
Marco~A Iglesias, Kody~JH Law, and Andrew~M Stuart.
\newblock Ensemble {K}alman methods for inverse problems.
\newblock \emph{Inverse Problems}, 29\penalty0 (4):\penalty0 045001, 2013.

\bibitem[Jang et~al.(2016)Jang, Gu, and Poole]{jang2017categorical}
Eric Jang, Shixiang Gu, and Ben Poole.
\newblock Categorical reparameterization with gumbel-softmax.
\newblock \emph{arXiv preprint arXiv:1611.01144}, 2016.

\bibitem[Kochkov et~al.(2021)Kochkov, Smith, Alieva, Wang, Brenner, and Hoyer]{kochkov2021machine}
Dmitrii Kochkov, Jamie~A Smith, Ayya Alieva, Qing Wang, Michael~P Brenner, and Stephan Hoyer.
\newblock Machine learning--accelerated computational fluid dynamics.
\newblock \emph{Proceedings of the National Academy of Sciences}, 118\penalty0 (21):\penalty0 e2101784118, 2021.

\bibitem[Kochkov et~al.(2024)Kochkov, Yuval, Langmore, Norgaard, Smith, Mooers, Kl{\"o}wer, Lottes, Rasp, D{\"u}ben, et~al.]{kochkov2024neural}
Dmitrii Kochkov, Janni Yuval, Ian Langmore, Peter Norgaard, Jamie Smith, Griffin Mooers, Milan Kl{\"o}wer, James Lottes, Stephan Rasp, Peter D{\"u}ben, et~al.
\newblock Neural general circulation models for weather and climate.
\newblock \emph{Nature}, 632\penalty0 (8027):\penalty0 1060--1066, 2024.

\bibitem[Kong et~al.(2024)Kong, Payne, Zhu, and Johnson]{kong2024saltation}
Nathan~J Kong, J~Joe Payne, James Zhu, and Aaron~M Johnson.
\newblock Saltation matrices: The essential tool for linearizing hybrid dynamical systems.
\newblock \emph{Proceedings of the IEEE}, 112\penalty0 (6):\penalty0 585--608, 2024.

\bibitem[Kool et~al.(2019)Kool, van Hoof, and Welling]{kool2019buy}
Wouter Kool, Herke van Hoof, and Max Welling.
\newblock Buy 4 {REINFORCE} samples, get a baseline for free!
\newblock ICLR 2019 Workshop on Deep Reinforcement Learning Meets Structured Prediction, 2019.
\newblock URL \url{https://openreview.net/forum?id=r1lgTGL5DE}.

\bibitem[Kreikemeyer and Andelfinger(2023)]{kreikemeyer2023smoothing}
Justin~N Kreikemeyer and Philipp Andelfinger.
\newblock Smoothing methods for automatic differentiation across conditional branches.
\newblock \emph{IEEE Access}, 11:\penalty0 143190--143211, 2023.

\bibitem[Kurz et~al.(2023)Kurz, Offenh{\"a}user, and Beck]{kurz2023deep}
Marius Kurz, Philipp Offenh{\"a}user, and Andrea Beck.
\newblock Deep reinforcement learning for turbulence modeling in large eddy simulations.
\newblock \emph{International journal of heat and fluid flow}, 99:\penalty0 109094, 2023.

\bibitem[Li et~al.(2018)Li, Aittala, Durand, and Lehtinen]{li2018differentiable}
Tzu-Mao Li, Miika Aittala, Fr{\'e}do Durand, and Jaakko Lehtinen.
\newblock Differentiable {Monte Carlo} ray tracing through edge sampling.
\newblock \emph{ACM Transactions on Graphics}, 37\penalty0 (6):\penalty0 1--11, 2018.
\newblock \doi{10.1145/3272127.3275109}.

\bibitem[List et~al.(2022)List, Chen, and Thuerey]{list2022learned}
Bj{\"o}rn List, Li-Wei Chen, and Nils Thuerey.
\newblock Learned turbulence modelling with differentiable fluid solvers: physics-based loss functions and optimisation horizons.
\newblock \emph{Journal of Fluid Mechanics}, 949:\penalty0 A25, 2022.

\bibitem[Liu et~al.(2023)Liu, Dong, Liu, Yu, and Gao]{liu2023bridging}
Liyuan Liu, Chengyu Dong, Xiaodong Liu, Bin Yu, and Jianfeng Gao.
\newblock Bridging discrete and backpropagation: Straight-through and beyond.
\newblock \emph{Advances in Neural Information Processing Systems}, 36:\penalty0 12291--12311, 2023.

\bibitem[Lopez-Gomez et~al.(2022)Lopez-Gomez, Christopoulos, Langeland~Ervik, Dunbar, Cohen, and Schneider]{lopez2022training}
Ignacio Lopez-Gomez, Costa Christopoulos, Haakon~Ludvig Langeland~Ervik, Oliver~RA Dunbar, Yair Cohen, and Tapio Schneider.
\newblock Training physics-based machine-learning parameterizations with gradient-free ensemble {K}alman methods.
\newblock \emph{Journal of Advances in Modeling Earth Systems}, 14\penalty0 (8):\penalty0 e2022MS003105, 2022.

\bibitem[Loubet et~al.(2019)Loubet, Holzschuch, and Jakob]{loubet2019reparameterizing}
Guillaume Loubet, Nicolas Holzschuch, and Wenzel Jakob.
\newblock Reparameterizing discontinuous integrands for differentiable rendering.
\newblock \emph{ACM Transactions on Graphics}, 38\penalty0 (6):\penalty0 1--14, 2019.
\newblock \doi{10.1145/3355089.3356510}.

\bibitem[Maddison et~al.(2016)Maddison, Mnih, and Teh]{maddison2017concrete}
Chris~J Maddison, Andriy Mnih, and Yee~Whye Teh.
\newblock The concrete distribution: A continuous relaxation of discrete random variables.
\newblock \emph{arXiv preprint arXiv:1611.00712}, 2016.

\bibitem[Mohamed et~al.(2020)Mohamed, Rosca, Figurnov, and Mnih]{mohamed2020monte}
Shakir Mohamed, Mihaela Rosca, Michael Figurnov, and Andriy Mnih.
\newblock {M}onte {C}arlo gradient estimation in machine learning.
\newblock \emph{Journal of Machine Learning Research}, 21\penalty0 (132):\penalty0 1--62, 2020.

\bibitem[Nesterov and Spokoiny(2017)]{nesterov2017random}
Yurii Nesterov and Vladimir Spokoiny.
\newblock Random gradient-free minimization of convex functions.
\newblock \emph{Foundations of Computational Mathematics}, 17\penalty0 (2):\penalty0 527--566, 2017.

\bibitem[Nicholson and Sykov{\'a}(1998)]{nicholson1998extracellular}
Charles Nicholson and Eva Sykov{\'a}.
\newblock Extracellular space structure revealed by diffusion analysis.
\newblock \emph{Trends in neurosciences}, 21\penalty0 (5):\penalty0 207--215, 1998.

\bibitem[Novati et~al.(2021)Novati, De~Laroussilhe, and Koumoutsakos]{novati2021automating}
Guido Novati, Hugues~Lascombes De~Laroussilhe, and Petros Koumoutsakos.
\newblock Automating turbulence modelling by multi-agent reinforcement learning.
\newblock \emph{Nature Machine Intelligence}, 3\penalty0 (1):\penalty0 87--96, 2021.

\bibitem[Oran and Gamezo(2007)]{oran2007origins}
Elaine~S Oran and Vadim~N Gamezo.
\newblock Origins of the deflagration-to-detonation transition in gas-phase combustion.
\newblock \emph{Combustion and flame}, 148\penalty0 (1-2):\penalty0 4--47, 2007.

\bibitem[Poli et~al.(2021)Poli, Massaroli, Scimeca, Chun, Oh, Yamashita, Asama, Park, and Garg]{poli2021neural}
Michael Poli, Stefano Massaroli, Luca Scimeca, Sanghyuk Chun, Seong~Joon Oh, Atsushi Yamashita, Hajime Asama, Jinkyoo Park, and Animesh Garg.
\newblock Neural hybrid automata: Learning dynamics with multiple modes and stochastic transitions.
\newblock \emph{Advances in Neural Information Processing Systems}, 34:\penalty0 9977--9989, 2021.

\bibitem[Ros and Hansen(2008)]{ros2008simple}
Raymond Ros and Nikolaus Hansen.
\newblock A simple modification in {CMA-ES} achieving linear time and space complexity.
\newblock In \emph{International conference on parallel problem solving from nature}, pages 296--305. Springer, 2008.

\bibitem[Salimans et~al.(2017)Salimans, Ho, Chen, Sidor, and Sutskever]{salimans2017evolution}
Tim Salimans, Jonathan Ho, Xi~Chen, Szymon Sidor, and Ilya Sutskever.
\newblock Evolution strategies as a scalable alternative to reinforcement learning.
\newblock \emph{arXiv preprint arXiv:1703.03864}, 2017.

\bibitem[Schulman et~al.(2015)Schulman, Moritz, Levine, Jordan, and Abbeel]{schulman2015high}
John Schulman, Philipp Moritz, Sergey Levine, Michael Jordan, and Pieter Abbeel.
\newblock High-dimensional continuous control using generalized advantage estimation.
\newblock \emph{arXiv preprint arXiv:1506.02438}, 2015.

\bibitem[Schulman et~al.(2017)Schulman, Wolski, Dhariwal, Radford, and Klimov]{schulman2017proximal}
John Schulman, Filip Wolski, Prafulla Dhariwal, Alec Radford, and Oleg Klimov.
\newblock Proximal policy optimization algorithms.
\newblock \emph{arXiv preprint arXiv:1707.06347}, 2017.

\bibitem[Seo(2026)]{seo2026tokagrad}
Jaemin Seo.
\newblock {TokaGrad}: End-to-end differentiable tokamak simulator for {L}-to-{H} full scenario optimization.
\newblock \emph{arXiv preprint arXiv:2607.09088}, 2026.

\bibitem[Shao et~al.(2024)Shao, Wang, Zhu, Xu, Song, Bi, Zhang, Zhang, Li, Wu, et~al.]{shao2024deepseekmath}
Zhihong Shao, Peiyi Wang, Qihao Zhu, Runxin Xu, Junxiao Song, Xiao Bi, Haowei Zhang, Mingchuan Zhang, YK~Li, Yang Wu, et~al.
\newblock {DeepSeekMath}: Pushing the limits of mathematical reasoning in open language models.
\newblock \emph{arXiv preprint arXiv:2402.03300}, 2024.

\bibitem[Shen et~al.(2023)Shen, Appling, Gentine, Bandai, Gupta, Tartakovsky, Baity-Jesi, Fenicia, Kifer, Li, et~al.]{shen2023differentiable}
Chaopeng Shen, Alison~P Appling, Pierre Gentine, Toshiyuki Bandai, Hoshin Gupta, Alexandre Tartakovsky, Marco Baity-Jesi, Fabrizio Fenicia, Daniel Kifer, Li~Li, et~al.
\newblock Differentiable modelling to unify machine learning and physical models for geosciences.
\newblock \emph{Nature Reviews Earth \& Environment}, 4\penalty0 (8):\penalty0 552--567, 2023.

\bibitem[Shepherd(2009)]{shepherd2009detonation}
Joseph~E Shepherd.
\newblock Detonation in gases.
\newblock \emph{Proceedings of the Combustion Institute}, 32\penalty0 (1):\penalty0 83--98, 2009.

\bibitem[Soltan et~al.(2014)Soltan, Mazauric, and Zussman]{soltan2014cascading}
Saleh Soltan, Dorian Mazauric, and Gil Zussman.
\newblock Cascading failures in power grids: analysis and algorithms.
\newblock In \emph{Proceedings of the 5th international conference on Future energy systems}, pages 195--206, 2014.

\bibitem[Song et~al.(2023)Song, Knoben, Clark, Feng, Lawson, and Shen]{song2023ancient}
Yalan Song, Wouter~JM Knoben, Martyn~P Clark, Dapeng Feng, Kathryn~E Lawson, and Chaopeng Shen.
\newblock When ancient numerical demons meet physics-informed machine learning: adjoint-based gradients for implicit differentiable modeling.
\newblock \emph{Hydrology and Earth System Sciences Discussions}, 2023:\penalty0 1--35, 2023.

\bibitem[Spall(1992)]{spall1992multivariate}
James~C Spall.
\newblock Multivariate stochastic approximation using a simultaneous perturbation gradient approximation.
\newblock \emph{IEEE transactions on automatic control}, 37\penalty0 (3):\penalty0 332--341, 1992.

\bibitem[Suh et~al.(2022)Suh, Simchowitz, Zhang, and Tedrake]{suh2022differentiable}
Hyung~Ju Suh, Max Simchowitz, Kaiqing Zhang, and Russ Tedrake.
\newblock Do differentiable simulators give better policy gradients?
\newblock In \emph{International Conference on Machine Learning}, pages 20668--20696. PMLR, 2022.

\bibitem[Titsias and L{\'a}zaro-Gredilla(2015)]{aueb2015local}
Michalis Titsias and Miguel L{\'a}zaro-Gredilla.
\newblock Local expectation gradients for black box variational inference.
\newblock \emph{Advances in neural information processing systems}, 28, 2015.

\bibitem[Tokui and Sato(2017)]{tokui2017evaluating}
Seiya Tokui and Issei Sato.
\newblock Evaluating the variance of likelihood-ratio gradient estimators.
\newblock In \emph{International Conference on Machine Learning}, pages 3414--3423. PMLR, 2017.

\bibitem[Tsuyuki(1996)]{tsuyuki1996variational}
T~Tsuyuki.
\newblock Variational data assimilation in the tropics using precipitation data part i: Column model.
\newblock \emph{Meteorology and Atmospheric Physics}, 60\penalty0 (1):\penalty0 87--104, 1996.

\bibitem[Tucker et~al.(2017)Tucker, Mnih, Maddison, Lawson, and Sohl-Dickstein]{tucker2017rebar}
George Tucker, Andriy Mnih, Chris~J Maddison, John Lawson, and Jascha Sohl-Dickstein.
\newblock Rebar: Low-variance, unbiased gradient estimates for discrete latent variable models.
\newblock \emph{Advances in Neural Information Processing Systems}, 30, 2017.

\bibitem[Tuckwell and Miura(1978)]{tuckwell1978mathematical}
Henry~C Tuckwell and Robert~M Miura.
\newblock A mathematical model for spreading cortical depression.
\newblock \emph{Biophysical Journal}, 23\penalty0 (2):\penalty0 257--276, 1978.

\bibitem[Um et~al.(2020)Um, Brand, Fei, Holl, and Thuerey]{um2020solver}
Kiwon Um, Robert Brand, Yun~Raymond Fei, Philipp Holl, and Nils Thuerey.
\newblock Solver-in-the-loop: Learning from differentiable physics to interact with iterative {PDE}-solvers.
\newblock \emph{Advances in neural information processing systems}, 33:\penalty0 6111--6122, 2020.

\bibitem[Varbella et~al.(2024)Varbella, Amara, Gjorgiev, El-Assady, and Sansavini]{varbella2024powergraph}
Anna Varbella, Kenza Amara, Blazhe Gjorgiev, Mennatallah El-Assady, and Giovanni Sansavini.
\newblock Powergraph: A power grid benchmark dataset for graph neural networks.
\newblock \emph{Advances in Neural Information Processing Systems}, 37:\penalty0 110784--110804, 2024.

\bibitem[von Bassewitz et~al.(2024)von Bassewitz, Kaltenbach, and Koumoutsakos]{von2024closure}
Jan-Philipp von Bassewitz, Sebastian Kaltenbach, and Petros Koumoutsakos.
\newblock Closure discovery for coarse-grained partial differential equations using grid-based reinforcement learning.
\newblock \emph{arXiv preprint arXiv:2402.00972}, 2024.

\bibitem[Williams(1992)]{williams1992simple}
Ronald~J Williams.
\newblock Simple statistical gradient-following algorithms for connectionist reinforcement learning.
\newblock \emph{Machine learning}, 8\penalty0 (3):\penalty0 229--256, 1992.

\bibitem[Yin et~al.(2019)Yin, Lyu, Zhang, Osher, Qi, and Xin]{yin2019understanding}
Penghang Yin, Jiancheng Lyu, Shuai Zhang, Stanley Osher, Yingyong Qi, and Jack Xin.
\newblock Understanding straight-through estimator in training activation quantized neural nets.
\newblock \emph{arXiv preprint arXiv:1903.05662}, 2019.

\bibitem[Zhang et~al.(2021)Zhang, Lin, Vogelmann, Zhang, Xie, Qin, and Golaz]{zhang2021improving}
Tao Zhang, Wuyin Lin, Andrew~M Vogelmann, Minghua Zhang, Shaocheng Xie, Yi~Qin, and Jean-Christophe Golaz.
\newblock Improving convection trigger functions in deep convective parameterization schemes using machine learning.
\newblock \emph{Journal of Advances in Modeling Earth Systems}, 13\penalty0 (5):\penalty0 e2020MS002365, 2021.

\bibitem[Zimmerman et~al.(2010)Zimmerman, Murillo-S{\'a}nchez, and Thomas]{zimmerman2010matpower}
Ray~Daniel Zimmerman, Carlos~Edmundo Murillo-S{\'a}nchez, and Robert~John Thomas.
\newblock {MATPOWER}: Steady-state operations, planning, and analysis tools for power systems research and education.
\newblock \emph{IEEE Transactions on power systems}, 26\penalty0 (1):\penalty0 12--19, 2010.

\end{thebibliography}
\bibliographystyle{plainnat}

\newpage
\appendix
\numberwithin{equation}{section}
\numberwithin{table}{section}
\numberwithin{figure}{section}

\makeatletter
\setlength{\@fptop}{0pt}\setlength{\@fpsep}{14pt}\setlength{\@fpbot}{0pt plus 1fil}
\setlength{\@dblfptop}{0pt}\setlength{\@dblfpsep}{14pt}\setlength{\@dblfpbot}{0pt plus 1fil}
\makeatother
\raggedbottom
\setcounter{theorem}{0}
\renewcommand{\thetheorem}{A.\arabic{theorem}}
\def\theHtheorem{A.\arabic{theorem}}

\section{Theory and algorithm}
\label{app:derivation}

This appendix proves the propositions of Section~\ref{sec:method} and gives the training update. Appendix~\ref{app:finite-histories} writes the smoothed objective as a sum over finite event histories and states the regularity conditions. Appendix~\ref{app:paired-consequences} sums over the two values of each switch to express the switch contribution in terms of the flip loss difference and thereby proves Propositions~\ref{prop:smoothed-gradient} and~\ref{prop:full-support-unbiased}. Appendix~\ref{app:sampling} restricts event sampling to the threshold band and bounds the resulting bias (Proposition~\ref{prop:truncation-bias}). Appendix~\ref{app:variance-surrogate} then presents the backward implementation. Finally, Appendix~\ref{app:hard-continuation} relates both variants to the hard loss (Propositions~\ref{prop:objective-gap} and~\ref{prop:hard-jumps-full}).

\begin{table*}[!ht]
\centering
\footnotesize
\setlength{\tabcolsep}{5pt}
\renewcommand{\arraystretch}{1.04}
\caption{\textbf{Notation for the gradient derivation.} Gradients and event contributions are parameter-space vectors.}
\label{tab:notation}
\begin{tabular}{p{0.32\linewidth}p{0.62\linewidth}}
\toprule
Symbol & Meaning \\
\midrule
$\theta$, $x$ & Model parameters and fixed window context. \\
$\vu_n$, $\vu_n^*$, $\vb_n$, $N$, $C$, $F_\theta$, $\ell_n$ & State, reference state, switch values $(b_{n,i})_i$, number of cells or components, quantities per cell, state update, and per-step loss. \\
$K$, $N_{\rm evt}$, $e$, $n_e$, $i_e$ & Window length, event count, event index, and its time step and cell or component. \\
$m_e$, $b_e$, $\beta$, $\beta_{<e}$ & Pre-event margin, switch value, complete history, and earlier decisions. \\
$\bar\vu_e$, $H_e$, $H_{\rm base}$ & State before the event\textquotesingle s stage, pre-event history and state, and complete base record. \\
$\sigma$, $\xi_e$, $\zeta_e$, $\xi$ & Positive width, standard normal threshold perturbation, remaining perturbations, and complete threshold-noise field. \\
$\beta^\xi$ & Switch history generated by Eq.~\eqref{eq:randomized-objective} under the noise $\xi$. \\
$\Phi$, $\phi$, $q_e$, $\mathsf p_e$, $P_\theta$ & Standard normal distribution function and density, probability that a switch is on, specified-value probability, and complete-history probability. \\
$\kappa_\sigma(m)$, $s_e$ & Sensitivity $\phi(m/\sigma)/\sigma$ and conditional probability score. \\
$\nabla_\theta^{\rm path}$, $\nabla_\theta^{\rm path}m_e$ & \leavevmode Total derivative taken with every switch value held fixed, and the margin derivative, which acts through the continuous states. \\
$L_\theta$, $J_\sigma$, $g_{\rm path}$ & Window loss, smoothed objective, and fixed-history path gradient. \\
$L_{<e}$, $G_e(b)$, $\mathcal Q_e(b\mid H_e)$ & Pre-event loss, downstream loss with event $e$ forced to $b$ and randomized switches afterwards, and conditional mean of the downstream loss. \\
$\mathcal A_e$, $\Delta_e$ & Advantage and flip loss difference measured by the flip branch. \\
$Z_e$, $W_e$ & Score-weighted loss and its conditional expectation over the current binary switch. \\
$\mathcal E_{\rm elig}$, $\mathcal B_\tau$ & Eligible events and threshold band. \\
$\tau$, $N_{{\rm out},\tau}$, $R_{\rm tail}(\tau)$ & Dimensionless cutoff, eligible omitted-event count, and omitted mean. \\
$B$, $M$, $\pi_e$, $e_{(j)}$ & Band size, branches per window, selection probability, and event of draw $j$. \\
$\lambda$, $c_j$ & Uniform fraction $\lambda\in(0,1]$ of the band proposal and coefficient of draw $j$. \\
$v_e$, $\widehat g_{\rm jump}$ & Measured event vector $\kappa_\sigma(m_e)\Delta_e\nabla_\theta^{\rm path}m_e$ and band jump estimate. \\
$\widehat g_\sigma$, $\widehat g_{\rm band}$ & The estimate in Eq.~\eqref{eq:method-full-estimator}, and the same estimate with the band selection probabilities of Eq.~\eqref{eq:method-band-proposal}. \\
$\widetilde L_\theta$, $\operatorname{sg}$ & Differentiation surrogate and stop-gradient. \\
$J_{\rm hard}$, $\beta^h$, $g_{\rm hard,band}$ & Hard loss, hard history, and expected update of \textsc{\ourmethod-Hard}. \\
$\vu_n^h$, $m_e^h$, $b_e^h$, $H_e^h$, $g_{\rm path}^h$, $\mathcal B_\tau^h$ & State, margin, switch value, pre-event history, path gradient, and threshold band evaluated on the hard-switch trajectory, marked by the superscript $h$. \\
$G_e^h(b)$ & Downstream loss with event $e$ forced to $b$ and hard switches afterwards. \\
$p_e^{\rm flip}$, $\mathcal A_e^h$, $\Delta_e^h$ & Flip probability $\Phi(-|m_e^h|/\sigma)$, advantage, and flip loss difference of the hard-switch simulator of event $e$ on the hard-switch trajectory. \\
$\theta(t)$, $\dot\theta(t)$, $r$, $t_k$, $e_k$, $\omega_k$ & Parameter path, its velocity, number of points of the path where a hard margin changes sign (crossing times), time and event of the $k$-th of them, and sign of its direction. \\
$D$, $C_m$, $\nu$, $\overline J_\sigma$ & Uniform bounds on $|\Delta_e|$ and $\|\nabla_\theta^{\rm path}m_e\|$, context distribution, and average-window smoothed objective. \\
\bottomrule
\end{tabular}
\end{table*}

Table~\ref{tab:notation} lists the notation used in the derivations.

\subsection{Finite event histories and the path gradient}
\label{app:finite-histories}

This subsection gives the finite-history form of the smoothed objective in Section~\ref{sec:jump} and states the regularity conditions assumed by Propositions~\ref{prop:smoothed-gradient}--\ref{prop:hard-jumps} and~\ref{prop:truncation-bias}--\ref{prop:objective-gap}. We collect the window's fixed initial state, reference observations, physical inputs, and time-step schedule into a context $x$. Let $N_{\rm evt}$ denote the fixed number of potential binary events. Throughout this finite-window derivation, we hold $x$ fixed and suppress it in the expectation notation.

A complete history $\beta=(\beta_1,\ldots,\beta_{N_{\rm evt}})$ specifies every switch value, and $\beta_{<e}$ denotes the values preceding event $e$. Once the history is fixed, the simulator is deterministic. We write the resulting loss as $L_\theta(\beta;x)$ and the conditional probability that event $e$ is on as $q_e(\theta,\beta_{<e};x)=\Phi\bigl(m_e(\theta,\beta_{<e};x)/\sigma\bigr)$.

\paragraph{Regularity conditions.}
At the parameter value of interest, we assume an open neighborhood on which every complete history keeps the same outcomes for the simulator's validity checks and substep selection. Any history that fails a check receives the penalty of Appendix~\ref{app:exp-details}. On this neighborhood, $L_\theta(\beta;x)$ and $m_e(\theta,\beta_{<e};x)$ are continuously differentiable for every history and its prefixes. This condition also covers unrandomized numerical operations such as absolute values and limiters. It holds when, for each fixed history, the branch that every such operation takes does not change on the neighborhood. In the three simulators of Appendix~\ref{app:exp-details}, for example, it holds when, for each history, the minmod limiter of the detonation wave keeps its branch in every cell and step and the rollout of cortical spreading depolarization neither enters nor leaves the invalid states. The power-grid margins $|f|/c_\theta-1$ are continuously differentiable in the closure because the switch history fixes the line flows $f$. Since the margins are finite and $\sigma>0$ is fixed, we have $0<q_e<1$. Conditions for averaging over contexts are stated at the end of this subsection.

An event is eligible when its two switch values produce different state updates, and we write $\mathcal E_{\rm elig}$ for the set of eligible events of a history. Every potential event, including every ineligible event, receives an independent threshold perturbation. At an ineligible event, either switch value produces the same state update and zero consequence. This convention keeps $N_{\rm evt}$ fixed. Simultaneous switches are ordered for conditioning, while their physical state update remains joint.

The state $\vu_n$ may also store continuous quantities read by the loss, such as fuel consumed between observations, which are updated alongside the other continuous variables. When a single step evaluates switches across several stages, $F_\theta$ composes those stages into one step. In that case, the state $\bar\vu_e$ at which event $e$ is tested is the state before that event's stage, and the losses $\ell_1,\ldots,\ell_{n_e}$ still precede event $e$.

Let $\mathsf p_e$ denote the probability of the specified value $\beta_e$ conditioned on the preceding history. By the chain rule, these conditional probabilities determine the complete-history probability, so realized events can depend on earlier decisions.
\begin{align}
\mathsf p_e(\theta,\beta;x)&=q_e^{\beta_e}(1-q_e)^{1-\beta_e},\nonumber\\
P_\theta(\beta\mid x)&=\prod_{e=1}^{N_{\rm evt}}\mathsf p_e(\theta,\beta;x),
\label{eq:history-probability}\\
J_\sigma(\theta\mid x)&=\sum_{\beta\in\{0,1\}^{N_{\rm evt}}}P_\theta(\beta\mid x)L_\theta(\beta;x).
\label{eq:finite-history-objective}
\end{align}

Because the threshold noise $\xi$ determines the history $\beta$, expectations over histories are written $\mathbb E_\xi$. The fixed-history path gradient is $g_{\rm path}(\beta;x)=\nabla_\theta L_\theta(\beta;x)$. When differentiating a history-specific loss, state, margin, or probability factor, we hold $x$ and the specified binary values fixed. The superscript $\mathrm{path}$ marks this convention explicitly on state-dependent quantities, so $\nabla_\theta^{\rm path}$ is the total derivative taken with every switch value held fixed, as in Section~\ref{sec:jump}. Differentiating the finite sum term by term separates the change in each history's loss from the contribution of its changing probability.
\begin{align}
\nabla_\theta J_\sigma(\theta\mid x)
&=\sum_\beta\nabla_\theta\bigl[P_\theta(\beta\mid x)L_\theta(\beta;x)\bigr]\nonumber\\
&=\sum_\beta P_\theta(\beta\mid x)g_{\rm path}(\beta;x) +\sum_\beta L_\theta(\beta;x)\nabla_\theta P_\theta(\beta\mid x).
\label{eq:objective-product-rule}
\end{align}

The derivative of the history probability follows by differentiating each conditional factor in turn. Because every factor is positive, the result can be written as a sum of logarithmic derivatives.
\begin{align}
\nabla_\theta P_\theta(\beta\mid x)
&=\sum_{e=1}^{N_{\rm evt}}\left(\prod_{e'\ne e}\mathsf p_{e'}\right)\nabla_\theta \mathsf p_e\nonumber\\
&=P_\theta(\beta\mid x)\sum_{e=1}^{N_{\rm evt}}\frac{\nabla_\theta \mathsf p_e}{\mathsf p_e}\nonumber\\
&=P_\theta(\beta\mid x)\sum_{e=1}^{N_{\rm evt}}\nabla_\theta\log \mathsf p_e.
\label{eq:history-probability-derivative}
\end{align}

Let $s_e=\nabla_\theta\log \mathsf p_e$ be the conditional probability score. With $\beta_e$ fixed at zero or one, its Bernoulli form gives
\begin{align}
\log \mathsf p_e&=\beta_e\log q_e+(1-\beta_e)\log(1-q_e),\nonumber\\
s_e&=\beta_e\frac{\nabla_\theta^{\rm path}q_e}{q_e}
 -(1-\beta_e)\frac{\nabla_\theta^{\rm path}q_e}{1-q_e}\nonumber\\
&=\left(\frac{\beta_e}{q_e}-\frac{1-\beta_e}{1-q_e}\right)\nabla_\theta^{\rm path}q_e.
\label{eq:event-score}
\end{align}

Substituting the probability derivative, written with these scores, yields a pathwise term and a score-weighted loss term, both averaged over the complete-history distribution.
\begin{align}
\nabla_\theta J_\sigma(\theta\mid x)
&=\sum_\beta P_\theta(\beta\mid x)g_{\rm path}(\beta;x) +\sum_\beta P_\theta(\beta\mid x)L_\theta(\beta;x)\sum_e s_e\nonumber\\
&=\mathbb E_\xi[g_{\rm path}]
 +\mathbb E_\xi\left[L_\theta\sum_{e=1}^{N_{\rm evt}}s_e\right].
\label{eq:path-score-decomposition}
\end{align}

Even with earlier switches held fixed, the continuous state can still depend on the parameters. For example, we write $m_e=m_e(\bar\vu_e(\theta,\beta_{<e};x);\theta)$, where $\bar\vu_e$ is the state at which the threshold of event $e$ is tested (that is, $\vu_{n_e}$ in Eq.~\eqref{eq:sim} when a step evaluates its switches in a single stage). The margin derivative accounts for both direct parameter dependence and the dependence propagated through the preceding continuous states.
\begin{equation}
\nabla_\theta^{\rm path}m_e
=\left.\nabla_\theta m_e(\vu;\theta)\right|_{\vu=\bar\vu_e}
 +\left(\frac{\partial^{\rm path}\bar\vu_e}{\partial\theta}\right)^{\!\mathsf T}
   \nabla_{\vu}m_e(\bar\vu_e;\theta).
\label{eq:margin-total-path-derivative}
\end{equation}

State derivatives act on the continuous coordinates while the discrete memory remains fixed by the specified history. The decomposition in Eq.~\eqref{eq:path-score-decomposition} forms the product-rule basis of pathwise and likelihood-ratio estimators \citep{mohamed2020monte,arya2022automatic}.

\paragraph{Averaging over window contexts.}
\label{app:window-context}
The training windows in Section~\ref{sec:setup} start from reference states that do not depend on $\theta$. Given a parameter-independent distribution $\nu$ of window contexts $x$ for which the losses and derivatives are integrable and differentiation can be interchanged with the expectation, the averaged objective and its gradient are
\begin{align}
\overline J_\sigma(\theta)
&=\mathbb E_{x\sim\nu}[J_\sigma(\theta\mid x)],
\label{eq:average-window-objective}\\
\nabla_\theta\overline J_\sigma
&=\mathbb E_{x\sim\nu}[\nabla_\theta J_\sigma(\theta\mid x)].
\label{eq:average-window-gradient}
\end{align}

\subsection{Conditional summation and paired consequences}
\label{app:paired-consequences}

This subsection proves Proposition~\ref{prop:smoothed-gradient} in Section~\ref{sec:jump} and Proposition~\ref{prop:full-support-unbiased} in Section~\ref{sec:paired-method}. It also shows that each event vector is a Rao--Blackwellization of the corresponding REINFORCE term.

\paragraph{From one switch to many.}
Consider first a simulator with a single switch of margin $m$. Let $L(b)$ be the window loss with the switch set to $b\in\{0,1\}$, so that $\Delta=L(1)-L(0)$ is the flip loss difference. The switch is on with probability $\Phi(m/\sigma)$, so $J_\sigma=L(0)+\Phi(m/\sigma)\,\Delta$ and $\nabla_\theta J_\sigma=\bigl(1-\Phi(m/\sigma)\bigr)\nabla_\theta L(0)+\Phi(m/\sigma)\,\nabla_\theta L(1)+\kappa_\sigma(m)\,\Delta\,\nabla_\theta m$. The first two terms are the path gradient averaged over the two switch values, which is the expected path gradient. The last term comes from differentiating the probability, and in it the flip loss difference is weighted by the sensitivity. Two things change when a window contains many switches. Event $e$ is on with probability $\Phi(m_e/\sigma)$ given the earlier switch values, and the probability of a history is the product of these conditional factors, so each factor is differentiated with the earlier switch values fixed and the derivative of $m$ becomes $\nabla_\theta^{\rm path}m_e$, which still contains the dependence of $m_e$ on $\theta$ through the earlier states. The loss after event $e$ depends on the later randomized switches, so the flip loss difference is replaced by its conditional expectation given $H_e$, the advantage $\mathcal A_e$ of Eq.~\eqref{eq:method-switch-advantage}. The proof below carries out these two steps.

\begin{proof}[Proof of Proposition~\ref{prop:smoothed-gradient}]
To evaluate the score contribution of event $e$, we condition on its preceding history and continuous state, collected in $H_e$, which fixes both $q_e$ and its path derivative. Let $L_{<e}$ be the loss accumulated before the event's state update, and let $G_e(b)$ be the downstream loss when this event is forced to $b\in\{0,1\}$ and every later switch is decided by the randomized rule in Eq.~\eqref{eq:randomized-objective}, as in Section~\ref{sec:jump}. The conditional mean of this downstream loss is $\mathcal Q_e(b\mid H_e)=\mathbb E[G_e(b)\mid H_e]$. Because the perturbation $\xi_e$ is independent of the preceding history and all later threshold perturbations, conditioning on the resulting switch value $\beta_e=b$ leaves the distribution of later perturbations unchanged and gives
\begin{equation}
\mathbb E[L_\theta\mid H_e,\beta_e=b]=L_{<e}+\mathcal Q_e(b\mid H_e).
\label{eq:prefix-suffix-loss}
\end{equation}

Conditional on the preceding history, the score averages to zero.
\begin{equation}
\mathbb E[s_e\mid H_e]
=q_e\frac{\nabla_\theta^{\rm path}q_e}{q_e}
 -(1-q_e)\frac{\nabla_\theta^{\rm path}q_e}{1-q_e}=0.
\label{eq:conditional-score-zero}
\end{equation}

Because $L_{<e}$ is fixed by $H_e$, this zero-mean property removes the accumulated loss from the score contribution. Summing over the two current decisions then leaves the difference between the conditional means of their downstream losses.
\begin{align}
\mathbb E[L_\theta s_e\mid H_e]
&=\mathbb E[(L_\theta-L_{<e})s_e\mid H_e]\nonumber\\
&=q_e\mathcal Q_e(1\mid H_e)\frac{\nabla_\theta^{\rm path}q_e}{q_e} -(1-q_e)\mathcal Q_e(0\mid H_e)\frac{\nabla_\theta^{\rm path}q_e}{1-q_e}\nonumber\\
&=\nabla_\theta^{\rm path}q_e\bigl[\mathcal Q_e(1\mid H_e)-\mathcal Q_e(0\mid H_e)\bigr].
\label{eq:conditional-event-sum}
\end{align}

Since the conditional advantage is $\mathcal A_e=\mathcal Q_e(1\mid H_e)-\mathcal Q_e(0\mid H_e)$, the right-hand side of Eq.~\eqref{eq:conditional-event-sum} equals $\nabla_\theta^{\rm path}q_e\,\mathcal A_e$. Equation~\eqref{eq:conditional-event-sum} is therefore the expected policy-gradient form for the binary switch obtained by summing over its two decisions \citep{tokui2017evaluating,ciosek2020expected}. Its left-hand side is the conditional mean of a REINFORCE term \citep{williams1992simple}.

Let $\zeta_e=\{\xi_{e'}:e'>e\}$ collect the threshold perturbations for events later in the fixed conditioning order, including those in the same time layer. As in Section~\ref{sec:jump}, the two continuations of event $e$ start from the pre-event state in $H_e$ with the switch forced to $1$ and to $0$. Both continuations share one realization of $\zeta_e$, recompute every later margin and switch from their respective states, and give the downstream losses $G_e(1)$ and $G_e(0)$. The other switches evaluated at the same stage as event $e$ keep their values in both continuations because their states and perturbations do not depend on the forced value, whereas switches at later stages of the same step are recomputed. Because the perturbations in $\zeta_e$ are independent of $H_e$ and $\xi_e$, for either forced value the continuation follows the same perturbation distribution as the original simulator conditioned on $H_e$ and that switch value. For $b\in\{0,1\}$, the continuation therefore has expected downstream loss $\mathcal Q_e(b\mid H_e)$, even when both continuations share the perturbations. Consequently, their flip loss difference $\Delta_e$ has conditional mean $\mathcal A_e$, as follows.
\begin{align}
\Delta_e&=G_e(1;\zeta_e,H_e)-G_e(0;\zeta_e,H_e),\nonumber\\
\mathbb E_{\zeta_e}[\Delta_e\mid H_e]
&=\mathbb E_{\zeta_e}[G_e(1)\mid H_e]-\mathbb E_{\zeta_e}[G_e(0)\mid H_e]\nonumber\\
&=\mathcal Q_e(1\mid H_e)-\mathcal Q_e(0\mid H_e).
\label{eq:paired-consequence-expectation}
\end{align}

The realized suffix supplies one side of each pair because the threshold perturbation $\xi_e$ affects the simulator only through $\beta_e$ and is independent of $\zeta_e$. For the same reason, conditioning also on the realized switch value leaves the mean unchanged, $\mathbb E[\Delta_e\mid H_e,\beta_e]=\mathcal A_e$. When $\beta_e=0$, the flipped-minus-base difference equals $G_e(1)-G_e(0)$, whereas when $\beta_e=1$ its sign is reversed. The factor $1-2\beta_e$, written $1-2b_e$ in Algorithm~\ref{alg:eventflip}, therefore always gives the flip loss difference $\Delta_e$.

The derivative of the probability $q_e$ follows from the Gaussian chain rule, since, for any real argument $z$, $\Phi'(z)=\phi(z)$ and the width $\sigma$ is fixed.
\begin{equation}
\begin{split}
\nabla_\theta^{\rm path}q_e
&=\nabla_\theta^{\rm path}\Phi(m_e/\sigma)
=\frac{\phi(m_e/\sigma)}{\sigma}\nabla_\theta^{\rm path}m_e\\
&=\kappa_\sigma(m_e)\nabla_\theta^{\rm path}m_e.
\end{split}
\label{eq:probability-margin-chain-rule}
\end{equation}

Because $H_e$ fixes the probability derivative, averaging the flip loss difference gives the score contribution of the event. Applying total expectation and the Gaussian chain rule yields
\begin{align}
\mathbb E[L_\theta s_e]
&=\mathbb E_{H_e}\bigl[\mathbb E[L_\theta s_e\mid H_e]\bigr]\nonumber\\
&=\mathbb E_{H_e}\bigl[\nabla_\theta^{\rm path}q_e\,
                  \mathbb E_{\zeta_e}[\Delta_e\mid H_e]\bigr]\nonumber\\
&=\mathbb E_{H_e,\zeta_e}\bigl[\nabla_\theta^{\rm path}q_e\,\Delta_e\bigr]\nonumber\\
&=\mathbb E_{H_e,\zeta_e}\bigl[\kappa_\sigma(m_e)\Delta_e\nabla_\theta^{\rm path}m_e\bigr].
\label{eq:score-to-paired-event}
\end{align}
By Eq.~\eqref{eq:paired-consequence-expectation} and Eq.~\eqref{eq:probability-margin-chain-rule}, the second line equals $\mathbb E_{H_e}[\kappa_\sigma(m_e)\,\mathcal A_e\,\nabla_\theta^{\rm path}m_e]$. Substituting this expression into Eq.~\eqref{eq:path-score-decomposition} for every event gives Eq.~\eqref{eq:grad}.
\end{proof}

\paragraph{Rao--Blackwellization.}
Once $H_e$ and $\zeta_e$ are fixed, the loss and score depend on $\xi_e$ only through the binary switch $\beta_e$, which equals $1$ with probability $q_e$ and $0$ with probability $1-q_e$. We define the score-weighted loss $Z_e=L_\theta s_e$ and $W_e=\nabla_\theta^{\rm path}q_e\,\Delta_e$, which equals the event vector $v_e$ defined after Eq.~\eqref{eq:method-full-estimator}. Summing over these two switch values gives
\begin{align}
\mathbb E[Z_e\mid H_e,\zeta_e]
&=q_e[L_{<e}+G_e(1;\zeta_e,H_e)]\frac{\nabla_\theta^{\rm path}q_e}{q_e} -(1-q_e)[L_{<e}+G_e(0;\zeta_e,H_e)]
                    \frac{\nabla_\theta^{\rm path}q_e}{1-q_e}\nonumber\\
&=\nabla_\theta^{\rm path}q_e\,[G_e(1;\zeta_e,H_e)-G_e(0;\zeta_e,H_e)]\nonumber\\
&=W_e.
\label{eq:event-rao-blackwell}
\end{align}

This identity shows that each event vector is a Rao--Blackwellization of its REINFORCE contribution. Because the window has finitely many histories, $Z_e$ is bounded. Since $\mathbb E[Z_e-W_e\mid H_e,\zeta_e]=0$, the cross term $\mathbb E[(W_e-\mathbb E W_e)^{\mathsf T}(Z_e-W_e)]$ vanishes. Expanding the centered squared norm therefore gives
\begin{align}
\mathbb E\|Z_e-\mathbb E Z_e\|^2
&=\mathbb E\|W_e-\mathbb E W_e\|^2+\mathbb E\|Z_e-W_e\|^2\nonumber\\
&\geq\mathbb E\|W_e-\mathbb E W_e\|^2.
\label{eq:event-rao-blackwell-variance}
\end{align}

This bound applies to each switch contribution individually, whereas the variance of their sum also includes covariances between switches and with the pathwise term.

\paragraph{Event sampling.}
Event selection introduces an additional layer of randomness. Let $\xi=(\xi_1,\ldots,\xi_{N_{\rm evt}})$ denote the complete threshold-noise field, which determines both the base record $H_{\rm base}$ and every potential flip branch prior to event selection. The selection probabilities $\pi_e(\theta,H_{\rm base})$ depend on the base record alone, satisfy $\sum_e\pi_e=1$ with $\pi_e>0$ for every $e\in\mathcal E_{\rm elig}$, and are written as $\pi_e(\xi)$ when conditioning on $\xi$. Given $\xi$, we draw $e_{(1)},\ldots,e_{(M)}$ independently with replacement for $M\geq1$ and form the estimator $\widehat g_\sigma$ of Eq.~\eqref{eq:method-full-estimator} with the weights $1/(M\pi_{e_{(j)}}(\xi))$.

Because event eligibility is determined by the base trajectory, an empty eligible set gives a zero jump term without sampling. At fixed $x$, the finite number of histories and the assumptions of Appendix~\ref{app:finite-histories} ensure finite first moments for the pathwise and jump terms. When averaging over $x\sim\nu$, we assume $\mathbb E\|g_{\rm path}\|<\infty$ and $\mathbb E\sum_e\|v_e\|<\infty$, together with the differentiation-expectation interchange in Appendix~\ref{app:window-context}.

\begin{proof}[Proof of Proposition~\ref{prop:full-support-unbiased}]
Conditioning on $\xi$ fixes the base path gradient, event vectors, and selection probabilities. Importance weighting then recovers the complete jump term.
\begin{align*}
\mathbb E\left[\frac{v_{e_{(j)}}}{\pi_{e_{(j)}}(\xi)}\,\middle|\,\xi\right]
&=\sum_{e:\pi_e(\xi)>0}\mathbb P(e_{(j)}=e\mid\xi)
        \frac{v_e}{\pi_e(\xi)}\\
&=\sum_{e:\pi_e(\xi)>0}v_e=\sum_{e=1}^{N_{\rm evt}}v_e,\\
\mathbb E[\widehat g_\sigma\mid\xi]
&=g_{\rm path}+\frac1M\sum_{j=1}^M\sum_e v_e
=g_{\rm path}+\sum_e v_e.
\end{align*}
Because every zero-probability event is ineligible, it has $\Delta_e=0$ and $v_e=0$, so the supported sum equals the complete sum. The same calculation gives $\mathbb E[\|v_{e_{(j)}}/\pi_{e_{(j)}}\|\mid\xi]=\sum_e\|v_e\|$ and ensures integrability. Although the proposal may depend on $\theta$ and the base record, it cancels in the conditional mean at the current parameter value.

It remains to average over $\xi$. Since $\mathbb E_\xi[v_e]=\mathbb E_\xi[L_\theta s_e]$ for every event by Eq.~\eqref{eq:score-to-paired-event}, total expectation and Eq.~\eqref{eq:path-score-decomposition} give
\begin{align*}
\mathbb E[\widehat g_\sigma]
&=\mathbb E_\xi[\mathbb E[\widehat g_\sigma\mid\xi]]
=\mathbb E_\xi[g_{\rm path}]+\sum_e\mathbb E_\xi[v_e]\\
&=\mathbb E_\xi[g_{\rm path}]+\sum_e\mathbb E_\xi[L_\theta s_e]
=\nabla_\theta J_\sigma(\theta\mid x).
\end{align*}
Because each draw has the same conditional mean, the result holds for every finite $M\geq1$. Under the context-average conditions in Appendix~\ref{app:window-context}, averaging over $x\sim\nu$ gives $\mathbb E[\widehat g_\sigma]=\nabla_\theta\overline J_\sigma(\theta)$.
\end{proof}

\subsection{Band sampling and truncation bias}
\label{app:sampling}

This subsection analyzes the selection probabilities in Eq.~\eqref{eq:method-band-proposal} on the threshold band and proves Proposition~\ref{prop:truncation-bias}. For a dimensionless cutoff $\tau>0$, the threshold band is $\mathcal B_\tau=\{e\in\mathcal E_{\rm elig}:|m_e|<\tau\sigma\}$ with size $B=|\mathcal B_\tau|$. The band estimator $\widehat g_{\rm band}=g_{\rm path}+\widehat g_{\rm jump}$ draws $e_{(1)},\ldots,e_{(M)}$ only from $\mathcal B_\tau$ and uses the jump estimate $\widehat g_{\rm jump}=M^{-1}\sum_{j=1}^Mv_{e_{(j)}}/\pi_{e_{(j)}}$, which is zero when $B=0$, so that an empty band reduces the estimate to the path gradient $g_{\rm path}$. Conditioning on the same realization $\xi$ leaves only the event-index selection random and fixes the band $\mathcal B_\tau$, its size $B$, and every event vector $v_e$. All norms below are Euclidean parameter-space norms.

\paragraph{Band mean.}
The selection probabilities in Eq.~\eqref{eq:method-band-proposal} of \ourmethod{} combine a component proportional to the sensitivity with a uniform fraction $\lambda\in(0,1]$ that keeps the importance weights bounded, since $\pi_e\geq\lambda/B$ gives $1/\pi_e\leq B/\lambda$. They are positive on the band and zero outside it. Conditional on $\xi$, the importance-weighting calculation in the proof of Proposition~\ref{prop:full-support-unbiased}, restricted to the support $\mathcal B_\tau$, gives $\mathbb E[v_{e_{(j)}}/\pi_{e_{(j)}}\mid\xi]=\sum_{e\in\mathcal B_\tau}v_e$ for every draw. Consequently, the jump estimate $\widehat g_{\rm jump}$ of $\widehat g_{\rm band}$ has the band sum as its conditional mean.

\paragraph{Truncation bias.}
To compare the band mean with the complete gradient, we define the omitted mean $R_{\rm tail}(\tau)$ by
\begin{equation}
R_{\rm tail}(\tau)=\mathbb E_\xi\left[\sum_{e\notin\mathcal B_\tau}v_e\right].
\label{eq:omitted-event-sum}
\end{equation}

\begin{proposition}[Truncation bias]
\label{prop:truncation-bias}
Under the conditions of Appendix~\ref{app:finite-histories} and with flip branches that share the later threshold noise, let the selection probabilities of $\widehat g_{\rm band}$ depend only on the base trajectory and be positive exactly on $\mathcal B_\tau$, as in Eq.~\eqref{eq:method-band-proposal}. Suppose $|\Delta_e|\leq D$ and $\|\nabla_\theta^{\rm path}m_e\|\leq C_m$ almost surely for every event. Then
\begin{equation*}
\bigl\|\mathbb E[\widehat g_{\rm band}]-\nabla_\theta J_\sigma(\theta)\bigr\|
\leq\frac{N_{\rm evt}\,D\,C_m}{\sigma\sqrt{2\pi}}\,e^{-\tau^2/2}.
\end{equation*}
\end{proposition}

\begin{proof}[Proof of Proposition~\ref{prop:truncation-bias}]
In conditional expectation, the band estimator recovers the band sum. Subtracting the complete gradient from the mean of the band estimator cancels the full pathwise term and leaves the omitted events.
\begin{align}
\mathbb E[\widehat g_{\rm band}]
&=\mathbb E_\xi\left[g_{\rm path}+\sum_{e\in\mathcal B_\tau}v_e\right],\nonumber\\
\nabla_\theta J_\sigma(\theta\mid x)
&=\mathbb E_\xi\left[g_{\rm path}+\sum_{e\in\mathcal B_\tau}v_e
                         +\sum_{e\notin\mathcal B_\tau}v_e\right],\nonumber\\
\mathbb E[\widehat g_{\rm band}]-\nabla_\theta J_\sigma(\theta\mid x)
&=-\mathbb E_\xi\left[\sum_{e\notin\mathcal B_\tau}v_e\right]
=-R_{\rm tail}(\tau).
\label{eq:truncation-bias}
\end{align}
Because the partition is made for each trajectory, this identity also holds for a trajectory-dependent band. Every eligible omitted event satisfies $|m_e|\geq \tau\sigma$, which bounds its sensitivity
\begin{equation}
\kappa_\sigma(m_e)
=\frac1{\sigma\sqrt{2\pi}}\exp\left(-\frac{m_e^2}{2\sigma^2}\right)
\leq\frac{\exp(-\tau^2/2)}{\sigma\sqrt{2\pi}}.
\label{eq:outside-band-kernel}
\end{equation}
Ineligible events contribute zero. Let $N_{{\rm out},\tau}=|\mathcal E_{\rm elig}\setminus\mathcal B_\tau|$ denote the eligible omitted-event count. Applying Jensen's inequality to the norm and the triangle inequality to the finite sum then gives
\begin{align}
\|R_{\rm tail}(\tau)\|
&\leq\mathbb E_\xi\left[\left\|\sum_{e\notin\mathcal B_\tau}v_e\right\|\right]\nonumber\\
&\leq\mathbb E_\xi\left[\sum_{e\in\mathcal E_{\rm elig}\setminus\mathcal B_\tau}
         \|\kappa_\sigma(m_e)\Delta_e\nabla_\theta^{\rm path}m_e\|\right]\nonumber\\
&=\mathbb E_\xi\left[\sum_{e\in\mathcal E_{\rm elig}\setminus\mathcal B_\tau}
         \kappa_\sigma(m_e)|\Delta_e|\,\|\nabla_\theta^{\rm path}m_e\|\right]\nonumber\\
&\leq\frac{D\,C_m\exp(-\tau^2/2)}{\sigma\sqrt{2\pi}}
         \mathbb E_\xi[N_{{\rm out},\tau}]\nonumber\\
&\leq\frac{N_{\rm evt}D\,C_m}{\sigma\sqrt{2\pi}}\exp(-\tau^2/2).
\label{eq:tail-bound}
\end{align}
At $\tau=4$, the Gaussian factor is $\exp(-8)$.
\end{proof}

\subsection{Backward implementation and variance constructions}
\label{app:variance-surrogate}

This subsection gives the backward implementation of the training update of Section~\ref{sec:paired-method}, written out in Algorithm~\ref{alg:eventflip}, and the variance identity behind the shared noise in Section~\ref{sec:jump}.

\paragraph{Stop-gradient.}
Let $\operatorname{sg}$ denote the stop-gradient operator and $c_j=\kappa_\sigma(m_{e_{(j)}})\Delta_{e_{(j)}}/(M\pi_{e_{(j)}})$ the coefficient of draw $j$. With the switch history held fixed, the gradient of the surrogate loss
\begin{equation}
\widetilde L_\theta=\sum_{n=1}^{K}\ell_n(\vu_n)+\sum_{j=1}^{M}\operatorname{sg}(c_j)\,m_{e_{(j)}}
\label{eq:method-surrogate}
\end{equation}
equals the estimator in Eq.~\eqref{eq:method-full-estimator}, since each added term differentiates to $c_j\nabla_\theta^{\rm path}m_{e_{(j)}}$. An event drawn more than once contributes one term for each draw.

Automatic differentiation holds the binary switches fixed and returns this estimate. The margins retain their full continuous histories. Algorithm~\ref{alg:eventflip} implements Eq.~\eqref{eq:method-surrogate} with the default kernel-weighted proposal in Eq.~\eqref{eq:method-band-proposal}. Its line~\ref{alg:line-xi} selects the variant, drawing the threshold noise for \textsc{EventFlip-Rand} and setting it to zero for \textsc{\ourmethod-Hard}. The flip branches keep no graph, and a window costs at most $(1+M)K$ solver steps. Batch averaging is described below. Reported loss and validation selection use $L_\theta$. Recomputation for the backward pass retains the same binary history and continuous state connections within the window.

\begin{algorithm*}[t]
\footnotesize

\DontPrintSemicolon
\LinesNumbered
\SetKwComment{tcp}{$\triangleright$~}{}
\SetCommentSty{textnormal}
\SetInd{0.2em}{0.6em}
\SetKwInput{Input}{Input}
\SetKwInput{Known}{Given}
\SetKwInput{Cost}{Cost}
\caption{\textbf{One training update of \ourmethod{} on one window.} Line~\ref{alg:line-xi} selects the variant, and $\kappa_\sigma(m)=\phi(m/\sigma)/\sigma$.}
\label{alg:eventflip}
\Input{Start state $\vu_0$, reference states $\{\vu_n^*\}_{n=1}^{K}$, parameters $\theta$, width $\sigma$, cutoff $\tau$, {number of branches $M$, and uniform fraction $\lambda$.}}
\leIf{\textup{\textsc{\ourmethod-Hard}}\nllabel{alg:line-xi}}{$\xi_e\leftarrow0$ for all events $e$}{$\xi_e\sim\mathcal N(0,1)$ i.i.d.}
\For(\tcp*[f]{base trajectory, with graph}){$n=0,\ldots,K-1$}{
  {$m_e\leftarrow m_{i_e}(\vu_n;\theta)$ and $b_e\leftarrow\mathbf 1[m_e\geq\sigma\xi_e]$ for each event $e$ at step $n$ (rule in Eq.~\eqref{eq:randomized-objective})};\quad $\vu_{n+1}\leftarrow F_\theta(\vu_n,\vb_n)$;\quad store $\vu_n$, $\vb_n$\nllabel{alg:line-switch}\;
}
$L\leftarrow\sum_{n=1}^{K}\ell_n(\vu_n)$;\quad $\mathcal B_\tau\leftarrow\{e\ \text{eligible}:|m_e|<\tau\sigma\}$ \tcp*{if empty, $\widetilde L_\theta\leftarrow L$, go to line~\ref{alg:line-update}}
{Compute the selection probabilities $\pi_e$ of Eq.~\eqref{eq:method-band-proposal} for $e\in\mathcal B_\tau$}\;
Draw $e_{(1)},\ldots,e_{(M)}\sim\pi$ i.i.d.\ (with replacement) {given the base trajectory}\;
\For(\tcp*[f]{flip branch, no graph}){$j=1,\ldots,M$, \textup{with} $e=e_{(j)}$}{
  $\vu^{\rm flip}_{n_e+1}\leftarrow F_\theta(\vu_{n_e},\vb_{n_e}\text{ with }b_e\to1-b_e)$ \tcp*{restart from stored $\vu_{n_e}$}
  \lFor{$n=n_e{+}1,\ldots,K{-}1$}{{$\vb^{\rm flip}_n\leftarrow$ rule in Eq.~\eqref{eq:randomized-objective} at $\vu^{\rm flip}_n$ with the same $\xi$;\quad $\vu^{\rm flip}_{n+1}\leftarrow F_\theta(\vu^{\rm flip}_n,\vb^{\rm flip}_n)$;\quad the stored $\vu_n$, $\vb_n$ stay unchanged}}
  {$G_e(1-b_e)\leftarrow\sum_{n>n_e}\ell_n(\vu^{\rm flip}_n)$;\quad $G_e(b_e)\leftarrow\sum_{n>n_e}\ell_n(\vu_n)$;\quad $\Delta_e\leftarrow(1-2b_e)\bigl[G_e(1-b_e)-G_e(b_e)\bigr]$} \tcp*{{loss difference $\Delta_e$}}
  $c_j\leftarrow\kappa_\sigma(m_e)\,\Delta_e/(M\pi_e)$\;
}
$\widetilde L_\theta\leftarrow L+\sum_{j=1}^{M}\operatorname{sg}(c_j)\,m_{e_{(j)}}(\theta)$ \tcp*{$m_e$ on the base trajectory}\nllabel{alg:line-surrogate}
$\theta\leftarrow\operatorname{OptimizerStep}(\theta,\nabla_\theta\widetilde L_\theta)$ \tcp*{{switches held constant, so $\nabla_\theta\widetilde L_\theta$ is the estimate in Eq.~\eqref{eq:method-full-estimator}}}\nllabel{alg:line-update}
\end{algorithm*}

\paragraph{Batch averaging.}
For a batch of $N_{\rm batch}\geq1$ windows indexed by $k$, each surrogate $\widetilde L_\theta^{(k)}$ is computed with its own context, history, and fixed coefficients, after which the update uses the mean $N_{\rm batch}^{-1}\sum_{k=1}^{N_{\rm batch}}\widetilde L_\theta^{(k)}$. Linearity of differentiation gives the mean of the window gradient estimates. Averaging preserves their expectation under the context conditions of Appendix~\ref{app:window-context}. Independence between windows is not needed for this mean identity.

\paragraph{Shared noise.}
Sharing the threshold noise changes the covariance between the two downstream losses. With finite conditional second moments,
\begin{equation}
\begin{aligned}
\operatorname{Var}(\Delta_e\mid H_e)
&=\operatorname{Var}(G_e(1)\mid H_e)+\operatorname{Var}(G_e(0)\mid H_e) -2\operatorname{Cov}(G_e(1),G_e(0)\mid H_e).
\end{aligned}
\label{eq:paired-noise-variance}
\end{equation}

Independent threshold noise gives zero covariance with the same marginal variances. Shared noise reduces the consequence variance when its induced covariance is positive (Appendix~\ref{app:common-noise}). With shared noise, the flip branch and the base trajectory use the same threshold noise after event $e$, so the two runs differ only through the flipped switch. \textsc{\ourmethod-Hard} draws no threshold noise, so the same holds for its flip branches.

\subsection{The update defined by hard continuation}
\label{app:hard-continuation}

This subsection supports Sections~\ref{sec:jump} and~\ref{sec:paired-method} and proves Propositions~\ref{prop:objective-gap} and~\ref{prop:hard-jumps-full}, which relate the smoothed objective and the expected updates of both variants to the hard loss. The window context $x$ is fixed and omitted from the notation. The losses and margins of a fixed history do not depend on $\sigma$. Let $\beta^h(\theta)$ be the hard history, namely the switch history of the deterministic simulator, so that $J_{\rm hard}(\theta)=L_\theta(\beta^h)$. A superscript $h$ marks a quantity evaluated on the hard-switch trajectory. Thus $b_e^h=\beta_e^h$, $m_e^h$, and $H_e^h$ are the switch value, the margin, and the pre-event history of event $e$, $g_{\rm path}^h$ is the path gradient, and $\mathcal B_\tau^h$ is the threshold band of the hard-switch trajectory. We write $q_e^h=\Phi(m_e^h/\sigma)$ for the probability that the randomized switch of event $e$ is on given $H_e^h$. When the earlier events take their hard values, the randomized switch of event $e$ differs from $b_e^h$ with the flip probability $p_e^{\rm flip}=\Phi(-|m_e^h|/\sigma)$. The factor $1-2b_e^h$ turns a flip loss difference into the difference between the flipped value and the hard value. The hard continuations of event $e$ start from $H_e^h$ with the switch forced on and forced off and use hard switches after event $e$, with $\Delta_e^h=G_e^h(1)-G_e^h(0)$ the flip loss difference of the hard-switch simulator (Section~\ref{sec:jump}). We write $\mathcal A_e^h=\mathcal A_e(H_e^h)$ for the advantage in Eq.~\eqref{eq:method-switch-advantage} evaluated at $H_e^h$. Its two continuations use randomized switches after event $e$, whereas the two continuations of $\Delta_e^h$ use hard switches. Sums over $e$ without a range run over all $N_{\rm evt}$ events.

Wherever no margin of the hard-switch trajectory is zero, the proof of Proposition~\ref{prop:objective-gap}(c) below shows that $\nabla_\theta J_{\rm hard}=g_{\rm path}^h$. The expected update of \textsc{\ourmethod-Hard}, which is Eq.~\eqref{eq:hard-update} in the main text, is
\begin{equation}
g_{\rm hard,band}=g_{\rm path}^{h}+\sum_{e\in\mathcal B_\tau^h}\kappa_\sigma(m_e^h)\Delta_e^h\nabla_\theta^{\rm path}m_e^h.
\label{eq:method-hard-direction}
\end{equation}

\textsc{\ourmethod-Hard} adds the measured flip loss differences $\Delta_e^h$ weighted by $\kappa_\sigma$ to this path gradient. Event sampling recovers the specified jump term in expectation and evaluates both the continuous sensitivity and the switch consequences on the hard-switch trajectory and its hard continuations. Because $\xi\equiv0$, the width $\sigma$ enters only through the threshold band and $\kappa_\sigma$.

Each summand of Eq.~\eqref{eq:method-hard-direction} is the jump term of Proposition~\ref{prop:smoothed-gradient} for the objective in which only the threshold of event $e$ is randomized and every other switch keeps the hard rule, because the advantage of event $e$ under that objective is $\Delta_e^h$. That objective is differentiable and the argument of Proposition~\ref{prop:smoothed-gradient} applies to event $e$ when every event before $e$ on the hard-switch trajectory has a nonzero margin and every other margin on the two hard continuations of event $e$ is nonzero. Under these conditions, the other switches keep their hard values near $\theta$ on the hard-switch trajectory before event $e$ and on both continuations.

The two propositions below relate the smoothed objective and the two expected updates to $J_{\rm hard}$.

\begin{proposition}[The two objectives and the update of \textsc{EventFlip-Hard}]
\label{prop:objective-gap}
Fix a training window and let the conditions of Appendix~\ref{app:finite-histories} hold at $\theta$.
\begin{enumerate}[label=\textup{(\alph*)},leftmargin=*]
\item For every $\sigma>0$,
\begin{equation}
\begin{split}
&J_\sigma(\theta)-J_{\rm hard}(\theta) =\sum_{e}p_e^{\rm flip}\,(1-2b_e^h)\,\mathcal A_e^h.
\end{split}
\label{eq:objective-gap}
\end{equation}
\item With $C_L=\max_\beta|L_\theta(\beta)-L_\theta(\beta^h)|$,
\begin{equation}
\begin{split}
|J_\sigma(\theta)-J_{\rm hard}(\theta)|
&\leq C_L\bigl[1-P_\theta(\beta^h)\bigr]\\
&\leq C_L\sum_{e}p_e^{\rm flip},
\end{split}
\label{eq:hard-objective-small-noise-limit}
\end{equation}
and $p_e^{\rm flip}\leq\tfrac12\exp\bigl(-(m_e^h)^2/(2\sigma^2)\bigr)$. If no margin of the hard-switch trajectory is zero, then $J_\sigma(\theta)\to J_{\rm hard}(\theta)$ as $\sigma\to0$.
\item Suppose that no margin of the hard-switch trajectory at $\theta$ is zero. Then the hard history is constant on a neighborhood of $\theta$, every $\mathcal A_e^h$ is continuously differentiable on that neighborhood, and
\begin{align}
\nabla_\theta J_\sigma
&=g_{\rm path}^h+\sum_{e}\kappa_\sigma(m_e^h)\,\mathcal A_e^h\,\nabla_\theta^{\rm path}m_e^h +\sum_{e}p_e^{\rm flip}\,(1-2b_e^h)\,\nabla_\theta\mathcal A_e^h.
\label{eq:smoothed-gradient-on-hard}
\end{align}
Consequently, the expected update in Eq.~\eqref{eq:method-hard-direction} satisfies
\begin{align}
&g_{\rm hard,band}-\nabla_\theta J_\sigma =-\sum_{e\notin\mathcal B_\tau^h}\kappa_\sigma(m_e^h)\,\Delta_e^h\,\nabla_\theta^{\rm path}m_e^h\nonumber\\
&\qquad+\sum_{e}\kappa_\sigma(m_e^h)\,(\Delta_e^h-\mathcal A_e^h)\,\nabla_\theta^{\rm path}m_e^h\nonumber\\
&\qquad-\sum_{e}p_e^{\rm flip}\,(1-2b_e^h)\,\nabla_\theta\mathcal A_e^h.
\label{eq:hard-update-gap}
\end{align}
\end{enumerate}
\end{proposition}

By the Gaussian bound in (b), the terms of Eq.~\eqref{eq:objective-gap} that are not exponentially small belong to the events whose hard margins lie within a few $\sigma$ of zero. The three sums in Eq.~\eqref{eq:hard-update-gap} have distinct sources. The first comes from band truncation, which omits the events outside $\mathcal B_\tau^h$. The second arises because each $\Delta_e^h$ is measured with hard continuations, whereas $\mathcal A_e^h$ averages over randomized continuations. The third enters because the update treats each measured flip loss difference as a constant, whereas the advantages in Eq.~\eqref{eq:objective-gap} vary with $\theta$.

\begin{proof}[Proof of Proposition~\ref{prop:objective-gap}]
\emph{Part (a).} For $e=1,\ldots,N_{\rm evt}+1$, let $V_e=\mathbb E[L_\theta\mid\beta_{<e}=\beta^h_{<e}]$ be the expected loss when the events before $e$ take their hard values and the events from $e$ on are randomized. The conditioning event has positive probability because $0<q_{e'}<1$ for every event $e'$. No event is fixed in $V_1$ and every event is fixed in $V_{N_{\rm evt}+1}$, so $V_1=J_\sigma(\theta)$ and $V_{N_{\rm evt}+1}=L_\theta(\beta^h)=J_{\rm hard}(\theta)$. For a fixed history the simulator is deterministic, so $\beta_{<e}=\beta^h_{<e}$ fixes the pre-event history at $H_e^h$. Given $H_e^h$, event $e$ is on with probability $q_e^h$. With $L_{<e}$ denoting the loss accumulated on the hard-switch trajectory before the state update of event $e$, Eq.~\eqref{eq:prefix-suffix-loss} gives
\begin{align*}
V_e&=L_{<e}+q_e^h\,\mathcal Q_e(1\mid H_e^h) +(1-q_e^h)\,\mathcal Q_e(0\mid H_e^h),\\
V_{e+1}&=L_{<e}+\mathcal Q_e(b_e^h\mid H_e^h).
\end{align*}
Subtracting the two lines, separately for $b_e^h=1$ and $b_e^h=0$, gives
\begin{align*}
V_e-V_{e+1}
&=(q_e^h-b_e^h) \times\bigl[\mathcal Q_e(1\mid H_e^h)-\mathcal Q_e(0\mid H_e^h)\bigr]\\
&=(q_e^h-b_e^h)\,\mathcal A_e^h.
\end{align*}
If $b_e^h=1$, then $m_e^h\geq0$ and $q_e^h-1=-\Phi(-m_e^h/\sigma)=-p_e^{\rm flip}$. If $b_e^h=0$, then $m_e^h<0$ and $q_e^h=\Phi(-|m_e^h|/\sigma)=p_e^{\rm flip}$. In both cases $q_e^h-b_e^h=p_e^{\rm flip}(1-2b_e^h)$. Summing $V_e-V_{e+1}$ over $e=1,\ldots,N_{\rm evt}$ telescopes to $V_1-V_{N_{\rm evt}+1}$, which is Eq.~\eqref{eq:objective-gap}.

\emph{Part (b).} Since the history probabilities sum to one, Eq.~\eqref{eq:finite-history-objective} gives $J_\sigma(\theta)-J_{\rm hard}(\theta)=\sum_\beta P_\theta(\beta)[L_\theta(\beta)-L_\theta(\beta^h)]$. The term $\beta=\beta^h$ is zero and every other bracket is at most $C_L$ in absolute value, which gives the first inequality. Given the hard values of the earlier events, event $e$ takes its hard value with probability $1-p_e^{\rm flip}$, so Eq.~\eqref{eq:history-probability} gives $P_\theta(\beta^h)=\prod_e(1-p_e^{\rm flip})$. Induction on the number of factors then yields $1-\prod_e(1-p_e^{\rm flip})\leq\sum_e p_e^{\rm flip}$. The bound on $p_e^{\rm flip}$ is the Gaussian tail bound $\Phi(-z)\leq\tfrac12e^{-z^2/2}$ for $z\geq0$. The constant $C_L$ and the margins $m_e^h$ do not depend on $\sigma$, so nonzero margins give $p_e^{\rm flip}\to0$ and the limit follows. By Eq.~\eqref{eq:prefix-suffix-loss}, $L_{<e}+\mathcal Q_e(b\mid H_e^h)$ is a conditional mean of $L_\theta$ for $b\in\{0,1\}$ and therefore differs from $L_\theta(\beta^h)$ by at most $C_L$. Because $L_{<e}$ cancels in $\mathcal A_e^h=\mathcal Q_e(1\mid H_e^h)-\mathcal Q_e(0\mid H_e^h)$, we obtain $|\mathcal A_e^h|\leq2C_L$. This bound does not depend on $\sigma$, so part (a) gives the same limit.

\emph{Part (c).} The margin of the first event depends on no switch value, is continuous in the parameters, and is nonzero at $\theta$, so its sign and hence $\beta_1^h$ are constant near $\theta$. If $\beta^h_{<e}$ is constant near $\theta$, the same argument applied to the margin $m_e$ of the fixed history $\beta^h_{<e}$ shows that $\beta_e^h$ is constant near $\theta$. After finitely many steps, the hard history is constant on a neighborhood $\mathcal O$ of $\theta$. On $\mathcal O$, $J_{\rm hard}$ is the loss of the fixed history $\beta^h$, which gives $\nabla_\theta J_{\rm hard}=g_{\rm path}^h$ on that neighborhood. By the proof of part (a), $p_e^{\rm flip}(1-2b_e^h)=\Phi(m_e^h/\sigma)-b_e^h$ with $b_e^h$ constant on $\mathcal O$, so its gradient is $\kappa_\sigma(m_e^h)\nabla_\theta^{\rm path}m_e^h$ by Eq.~\eqref{eq:probability-margin-chain-rule}. By Eq.~\eqref{eq:prefix-suffix-loss}, the advantage is the finite sum
\begin{equation*}
\mathcal A_e^h=\sum_{\beta}(2\beta_e-1)\Bigl[\prod_{e'>e}\mathsf p_{e'}(\theta,\beta)\Bigr]L_\theta(\beta),
\end{equation*}
taken over the complete histories $\beta$ with $\beta_{<e}=\beta^h_{<e}$. Every factor is continuously differentiable on $\mathcal O$, so $\mathcal A_e^h$ is also continuously differentiable, with gradient denoted by $\nabla_\theta\mathcal A_e^h$. Differentiating Eq.~\eqref{eq:objective-gap} with the product rule gives Eq.~\eqref{eq:smoothed-gradient-on-hard}. Subtracting Eq.~\eqref{eq:smoothed-gradient-on-hard} from Eq.~\eqref{eq:method-hard-direction} cancels $g_{\rm path}^h$. Substituting $\mathcal A_e^h=\Delta_e^h-(\Delta_e^h-\mathcal A_e^h)$ into the first sum of Eq.~\eqref{eq:smoothed-gradient-on-hard} cancels the terms with $\Delta_e^h$ for the events in $\mathcal B_\tau^h$ and leaves them with a minus sign for the events outside the band, which gives Eq.~\eqref{eq:hard-update-gap}.
\end{proof}

\begingroup
\renewcommand{\thetheorem}{\ref*{prop:hard-jumps}}%
\def\theHtheorem{hard-jumps-full}%
\addtocounter{theorem}{-1}%
\begin{proposition}[The two updates and the hard loss, full statement]
\label{prop:hard-jumps-full}
Let $\theta(t)$, $t\in[0,1]$, be a continuously differentiable path in parameter space with velocity $\dot\theta(t)$. Let the conditions of Appendix~\ref{app:finite-histories} hold on an open set containing the path, and suppose that no margin of the hard-switch trajectory is zero at $\theta(0)$ or at $\theta(1)$.
\begin{enumerate}[label=\textup{(\alph*)},leftmargin=*]
\item The expected update of \textsc{EventFlip-Rand} with full-support event sampling is $\nabla_\theta J_\sigma$ by Proposition~\ref{prop:full-support-unbiased}, and
\begin{align}
&\lim_{\sigma\to0}\int_0^1\nabla_\theta J_\sigma(\theta(t))\cdot\dot\theta(t)\,dt =J_{\rm hard}(\theta(1))-J_{\rm hard}(\theta(0)).
\label{eq:smoothed-line-integral}
\end{align}
\item Call $t$ a crossing time if some margin of the hard-switch trajectory at $\theta(t)$ is zero (under the conditions below, where the margin is strictly monotone along the path near such a time, the crossing times are the points of the path at which the margin of an event on the hard-switch trajectory changes sign, as in the text after Eq.~\eqref{eq:hard-loss-path}), and suppose that every crossing time satisfies three crossing conditions. First, exactly one event $e$ has zero margin on the hard-switch trajectory. Second, with the hard values of the earlier events held fixed, the margin of event $e$ has a nonzero derivative along the path, namely $\nabla_\theta^{\rm path}m_e^h\cdot\dot\theta(t)\neq0$. Third, on each of the two hard continuations that start at event $e$ with its switch forced on and forced off, every event other than $e$ has a nonzero margin. These conditions exclude events whose margins are zero at the same time, whether on the hard-switch trajectory or on either hard continuation. Then there are finitely many crossing times $t_1<\cdots<t_r$. Let $e_k$ be the event corresponding to $t_k$, let $\omega_k\in\{-1,1\}$ be the sign of $\nabla_\theta^{\rm path}m_{e_k}^h\cdot\dot\theta(t_k)$, and evaluate $\Delta_{e_k}^h$ at $\theta(t_k)$. The hard loss along the path is continuously differentiable between crossing times with derivative $g_{\rm path}^h\cdot\dot\theta$, it jumps by $\omega_k\Delta_{e_k}^h$ at $t_k$, and, as in Eq.~\eqref{eq:hard-loss-path} of the main text,
\begin{align}
&J_{\rm hard}(\theta(1))-J_{\rm hard}(\theta(0)) =\int_0^1 g_{\rm path}^h\cdot\dot\theta\,dt+\sum_{k=1}^r\omega_k\,\Delta_{e_k}^h.
\label{eq:hard-loss-along-path}
\end{align}
The expected update in Eq.~\eqref{eq:method-hard-direction} of {\normalfont\textsc{\ourmethod-Hard}} satisfies
\begin{align}
&\lim_{\sigma\to0}\int_0^1 g_{\rm hard,band}(\theta(t))\cdot\dot\theta(t)\,dt =\int_0^1 g_{\rm path}^h\cdot\dot\theta\,dt +\bigl(2\Phi(\tau)-1\bigr)\sum_{k=1}^r\omega_k\,\Delta_{e_k}^h.
\label{eq:hard-line-integral}
\end{align}
Each jump $\omega_k\Delta_{e_k}^h$ includes every later switch that changes its value as a consequence of the flip of event $e_k$.
\end{enumerate}
\end{proposition}
\endgroup

The limit in (a) needs no condition at the crossing times, and under the conditions of (b) its right side equals the right side of Eq.~\eqref{eq:hard-loss-along-path}. The conditions of (b) are sufficient conditions. The factor $2\Phi(\tau)-1$ in Eq.~\eqref{eq:hard-line-integral} differs from $1$ by $2\Phi(-\tau)\leq e^{-\tau^2/2}$, by the Gaussian tail bound used in the proof of Proposition~\ref{prop:objective-gap}(b), and is $0.99994$ at $\tau=4$.

\begin{proof}[Proof of Proposition~\ref{prop:hard-jumps-full}]
\emph{Part (a).} For fixed $\sigma>0$, Eq.~\eqref{eq:finite-history-objective} writes $J_\sigma$ as a finite sum of products of functions that are continuously differentiable on the open set, so $t\mapsto J_\sigma(\theta(t))$ is continuously differentiable with derivative $\nabla_\theta J_\sigma(\theta(t))\cdot\dot\theta(t)$. The fundamental theorem of calculus gives $\int_0^1\nabla_\theta J_\sigma\cdot\dot\theta\,dt=J_\sigma(\theta(1))-J_\sigma(\theta(0))$ for every $\sigma$, and applying Proposition~\ref{prop:objective-gap}(b) at the two endpoints yields Eq.~\eqref{eq:smoothed-line-integral}.

\emph{Part (b), crossing times.} Let $t^\ast$ be a time that is not a crossing time. The induction in the proof of Proposition~\ref{prop:objective-gap}(c), applied to the fixed-history margins as continuous functions of $t$, shows that the hard history is constant and every hard margin is nonzero for $t$ near $t^\ast$. The times that are not crossing times therefore form an open subset of $[0,1]$, and the hard history is constant on each of its intervals. Now let $t^\ast$ be a crossing time with event $e$, and let $\beta^{(1)}$ and $\beta^{(0)}$ be the complete histories of the two hard continuations at $\theta(t^\ast)$ with event $e$ forced on and forced off. The hard-switch trajectory at $\theta(t^\ast)$ is $\beta^{(1)}$, because the hard rule $\mathbf 1[m\geq0]$ turns a switch with zero margin on. The two histories share the switch values before event $e$. For these values, the margin of event $e$ is $\psi(t)$, which is continuously differentiable in $t$ with $\psi(t^\ast)=0$ and $\dot\psi(t^\ast)=\nabla_\theta^{\rm path}m_e^h\cdot\dot\theta(t^\ast)\neq0$. By the first and third conditions, every margin of $\beta^{(1)}$ and of $\beta^{(0)}$ other than that of event $e$ is nonzero at $t^\ast$. By continuity there is $\delta>0$ such that, for $|t-t^\ast|\leq\delta$, these margins keep their signs and $|\dot\psi(t)|\geq|\dot\psi(t^\ast)|/2$. For such $t$, the hard rule reproduces the switch values before event $e$, sets event $e$ to $\mathbf 1[\psi(t)\geq0]$, and then reproduces the later values of the corresponding continuation, so
\begin{equation}
\beta^h(\theta(t))=
\begin{cases}
\beta^{(1)},&\psi(t)\geq0,\\
\beta^{(0)},&\psi(t)<0.
\end{cases}
\label{eq:hard-history-near-crossing}
\end{equation}
Since $\psi$ is strictly monotone on this interval, $t^\ast$ is the only crossing time in the interval. The crossing times thus form a closed subset of $[0,1]$ whose points are all isolated, so this set is finite.

\emph{Part (b), the hard loss.} Between consecutive crossing times the hard history is a constant $\beta$, so $J_{\rm hard}(\theta(t))=L_{\theta(t)}(\beta)$ is continuously differentiable with derivative $g_{\rm path}^h\cdot\dot\theta$ and has limits at both ends of the interval. Near $t_k$, write $\psi_k$, $\beta^{(1)}$, and $\beta^{(0)}$ for the objects of the previous step and let $\Lambda_k(t)=L_{\theta(t)}(\beta^{(1)})-L_{\theta(t)}(\beta^{(0)})$, which is continuous in $t$. By the previous step, for $|t-t_k|\leq\delta$ the hard continuations at $\theta(t)$ that start at event $e_k$ with its switch forced on and forced off reproduce $\beta^{(1)}$ and $\beta^{(0)}$. The two histories share the loss accumulated before the state update of event $e_k$, so $\Lambda_k(t)$ is the difference of their later losses, which equals $\Delta_{e_k}^h$ at $\theta(t)$. By Eq.~\eqref{eq:hard-history-near-crossing}, $J_{\rm hard}(\theta(t))$ equals $L_{\theta(t)}(\beta^{(1)})$ where $\psi_k(t)\geq0$ and $L_{\theta(t)}(\beta^{(0)})$ where $\psi_k(t)<0$. The sign of $\psi_k$ changes from $-\omega_k$ to $\omega_k$ at $t_k$. The limit from the right minus the limit from the left is therefore $\omega_k\Lambda_k(t_k)=\omega_k\Delta_{e_k}^h$. Applying the fundamental theorem of calculus on each interval between crossing times and adding the jumps gives Eq.~\eqref{eq:hard-loss-along-path}.

\emph{Part (b), the band integral.} The integrand $g_{\rm hard,band}\cdot\dot\theta$ is bounded and measurable in $t$, because it is built from finitely many fixed-history losses, margins, and their gradients, all continuous along the path, and the histories are selected by the signs of finitely many continuous margins. Shrink $\delta$ so that the intervals $U_k=[t_k-\delta,t_k+\delta]$ are disjoint, lie in $(0,1)$, and satisfy the two previous steps. The rest of $[0,1]$ is covered by finitely many closed intervals without crossing times, on each of which the hard history is constant and the hard margins are continuous and nonzero, so their absolute values have a positive minimum. On $U_k$, the margins of $\beta^{(1)}$ and $\beta^{(0)}$ other than that of $e_k$ also have a positive minimum absolute value, and $\psi_k(t_k\pm\delta)\neq0$. Take $\sigma$ so small that $\tau\sigma$ lies below all of these values. The band is then empty outside the intervals $U_k$, while on $U_k$ it can contain only $e_k$, which requires $|\psi_k(t)|<\tau\sigma$. If $e_k$ is not eligible at $\theta(t)$, its two continuations coincide and $\Lambda_k(t)=0$. Because the switch values before $e_k$ are constant on $U_k$, we have $\nabla_\theta^{\rm path}m_{e_k}^h\cdot\dot\theta=\dot\psi_k$ on this interval. In every case
\begin{align*}
&\int_0^1\bigl(g_{\rm hard,band}-g_{\rm path}^h\bigr)\cdot\dot\theta\,dt =\sum_{k=1}^r\int_{U_k}\mathbf 1[|\psi_k|<\tau\sigma]\,\kappa_\sigma(\psi_k)\,\Lambda_k\,\dot\psi_k\,dt.
\end{align*}
On $U_k$, $\psi_k$ is strictly monotone and its range contains $[-\tau\sigma,\tau\sigma]$. The substitution $z=\psi_k(t)/\sigma$, with $\kappa_\sigma(\sigma z)=\phi(z)/\sigma$ and $dz=\dot\psi_k\,dt/\sigma$, turns the $k$-th integral into
\begin{equation*}
\omega_k\int_{-\tau}^{\tau}\phi(z)\,\Lambda_k\bigl(\psi_k^{-1}(\sigma z)\bigr)\,dz,
\end{equation*}
where the factor $\omega_k$ accounts for the direction in which $z$ runs. As $\sigma\to0$, $\psi_k^{-1}(\sigma z)\to t_k$ uniformly for $|z|\leq\tau$ and $\Lambda_k$ is continuous, so this expression tends to $\omega_k\Lambda_k(t_k)\int_{-\tau}^{\tau}\phi(z)\,dz=(2\Phi(\tau)-1)\,\omega_k\Delta_{e_k}^h$. Since the integral of $g_{\rm path}^h\cdot\dot\theta$ does not depend on $\sigma$, summing over $k$ gives Eq.~\eqref{eq:hard-line-integral}.
\end{proof}

\section{Experimental details}
\label{app:exp-details}

This appendix describes the three simulators, the configurations of \ourmethod{} and of each baseline, and the training and evaluation protocol behind every result in Section~\ref{sec:experiments}.

\begin{table*}[!ht]
\centering
\caption{\textbf{Switch properties of the three simulators.} The first two rows follow from the model equations. The third row gives the number of switches in the threshold band $|m|<4\sigma$ per training window at $\sigma=0.02$ (in the units of Appendix~\ref{app:eventflip-config}), reported as a range over the untrained and the trained closure.}
\label{tab:properties-full}
\footnotesize
\begin{tabular}{@{}>{\raggedright\arraybackslash}p{0.22\linewidth}>{\raggedright\arraybackslash}p{0.23\linewidth}>{\raggedright\arraybackslash}p{0.23\linewidth}>{\raggedright\arraybackslash}p{0.23\linewidth}@{}}
\toprule
 & Power-grid cascade & Detonation wave & Cortical spreading depolarization \\
\midrule
Closure changes the outcome with every switch held fixed & {No, since the topology fixes flows and shed load} & {Yes, through pressure, temperature, and density} & {Yes, since K$^+$ rises continuously toward the threshold} \\
\addlinespace
Physical intermediate state of a switch & None, since a line is either in service or open & Partial burning & None, since a cell is either resting or activated and activation is never undone \\
\addlinespace
Switches near their thresholds per training window & 1.0--1.3 & 3.4--8.0 & 347--374 \\
\bottomrule
\end{tabular}
\end{table*}

Table~\ref{tab:properties-full} lists the switch properties behind the choice of scenarios. On the power grid, the closure cannot change the loss while the switches are held fixed, so the path gradient is zero. On the detonation wave, the closure acts mostly through this fixed-switch route, and training with the jump term alone reaches only 4.3 to 4.4\% (Table~\ref{tab:mechanism}). On cortical spreading depolarization, the switch has no physical intermediate state, and hundreds of switches lie near their thresholds in one training window.

\subsection{Power-grid cascade}
\label{app:power-grid}

\paragraph{Network and cascade model.} The simulator runs DC power flow on the IEEE 118-bus system from MATPOWER \citep{zimmerman2010matpower}, which has 118 buses and 186 branches (including 11 transformers). The nominal rating $c_{\rm nom}$ of each line is the rateA value of the PGLib-OPF case \texttt{pglib\_opf\_case118\_ieee} \citep{babaeinejadsarookolaee2019power}. In each generation, the simulator splits the in-service lines into islands and balances each island, shedding load proportionally when generation is short and reducing generation proportionally when it exceeds the load. It then solves the DC power flow and opens every in-service line whose loading reaches its rating. Opened lines stay open, and the cascade stops at the first generation without a trip \citep{soltan2014cascading,bernstein2014power}. Each line carries the switch $b=\mathbf 1[m\ge0]$ with margin $m=|f|/c_\theta-1$, where $f$ is the line flow and $c_\theta$ is the rating predicted by the closure. Once the topology is fixed, flows and shed load do not depend on the closure, so the path gradient with switches held fixed is zero. Reference cascades run for at most 10 generations.

\paragraph{Cases.} Bus loads are multiplied by independent lognormal factors with log-standard deviation 0.1 and rescaled to a total of $s$ times the base load, with $s$ uniform in $[0.65,0.75]$. The simulator uses a preventive security-constrained dispatch that keeps every line flow within $|f|\le u\,c_{\rm nom}$ both in the intact network and after every single-line outage that does not split the network, with $u$ uniform in $[0.95,1.0]$. Among such dispatches, it selects the one closest in $L^1$ distance to proportional dispatch. The operator knows only the nominal ratings, and cases without a secure dispatch are redrawn. The initial outage removes two lines that do not split the network, drawn with probability proportional to the square of their nominal pre-outage loading. Weather conditions are divided into three regions. The ambient temperature is a case-wide value uniform in $[0,40]^\circ$C plus a regional term with standard deviation $2^\circ$C. The wind speed is $\exp(\ln3.5+0.5\eta)\exp(0.2\eta_r)$, clipped to $[0.3,12]$\,m/s, and the wind direction is uniform on $[0,2\pi)$ plus $0.35\eta_r$, where $\eta$ and $\eta_r$ are case-wide and regional standard normal variables, respectively.

\paragraph{Reference ratings and data.} The reference rating of each line is $c^\star=c_{\rm nom}\exp(g^\star+\varepsilon)$, where $\varepsilon\sim\mathcal N(0,0.03^2)$ is drawn independently for every line and case and cannot be represented by the closure. The factor $g^\star$ is modeled after the IEEE Std~738 steady-state heat balance \citep{ieee2012std738}, with current proportional to $\sqrt{h(v_{\rm eff})(T_{\max}-T_a)}$ and $h=1+1.5\sqrt{v_{\rm eff}}$ for the effective wind speed $v_{\rm eff}$ at the line. The temperature limit $T_{\max}$ is 100, 75, or 65$^\circ$C depending on the conductor class, and $g^\star$ is normalized so that $g^\star=0$ for a standard conductor at 20$^\circ$C under a perpendicular wind of 2\,m/s. The 345\,kV lines are further capped by the St.~Clair limit, an empirical bound on how heavily a long line may be loaded, expressed as a multiple of its surge impedance loading and computed with a surge impedance (the characteristic impedance of the line) of 285\,$\Omega$. Aged conductors in the third of the three weather regions (buses 68--112, 116, and 118) are derated by 7.7\% (0.08 in log rating), and transformers follow limits of 120, 110, or 98$^\circ$C on the hot-spot temperature (the temperature of the hottest point of the winding) without wind cooling. The training, validation, and test sets contain 4{,}000, 600, and 600 cases.

\paragraph{Closure.} The closure predicts every rating as $c_\theta=c_{\rm nom}\exp(\mathrm{MLP}_\theta(z))$ with one MLP shared by all lines and cases. The 13 input features $z$ of a line are a one-hot voltage level (3), its length divided by 100\,km, a one-hot conductor class (3), a one-hot region (3), $(T_a-20)/10$, $(v-3.5)/2.5$, and $(v|\sin\varphi|-2)/2$, with ambient temperature $T_a$ in $^\circ$C, wind speed $v$ in m/s, and $\varphi$ the angle between wind and line. The MLP has layers $13\to32\to32\to1$ with GELU activations and 1{,}537 parameters in total. Its output layer is initialized to zero, so the untrained closure returns the nominal ratings.

\paragraph{Window loss and metrics.} A training window starts from the reference topology at generation $k_0$, advances three generations, and compares its step $j=0,1,2$ with reference generation $k_0+j+1$. The step loss is the mean over lines of $((f-f_{\rm ref})/c_{\rm nom})^2$ plus the mean over buses of the squared shed-load error normalized by the mean bus load of the case (its total load divided by 118). A window whose loss is not finite is assigned a loss of $10^4$. The main metric is the rollout loss, defined as the sum of the step losses over all generations of the cascade simulated from the initial outage and averaged over the 600 test cases, 96 of which cascade. The three physical quantities are averaged over cases for each seed and then over seeds.
\begin{itemize}
\item \emph{Shed load.} The absolute difference between the predicted and reference final shed load, divided by the total system load and expressed in percent.
\item \emph{Tripped lines.} The absolute difference in the number of lines that are in service after the initial outage and trip by the end of the cascade.
\item \emph{Severe-case AUC.} A case is severe when its reference final shed load is at least 5\% of the total load, which holds for 67 of the 600 test cases (11.2\%). The AUC scores cases by their predicted shed load.
\end{itemize}

\subsection{Detonation wave}
\label{app:detonation}

\paragraph{Physical model and discretization.} The simulator solves the one-dimensional reactive Euler equations for the density $\rho$, momentum $\rho u$, total energy $E$, and unburnt-fuel mass fraction $Z$ of an ideal gas with $\gamma=1.4$, using Heaviside ignition kinetics \citep{bao2000random}. With temperature $T=p/\rho$, every cell carries the switch $b=\mathbf 1[T\ge T_{\rm ign}]$ with margin $m=T-T_{\rm ign}$, and the reaction step integrates $\rho Z\leftarrow\rho Z\exp(-k_r\Delta t\,b)$ exactly over the time step $\Delta t$. The convective part uses HLLC fluxes with $Z$ advected as a passive scalar, MUSCL reconstruction of the primitive variables with the minmod limiter, and second-order strong-stability-preserving Runge--Kutta time stepping. In CGS units, the base rate is $k_r=0.5825\times10^{10}$, the base heat release is $q_0=0.5196\times10^{10}$, and the base ignition temperature is $T_{\rm ign}=0.1155\times10^{10}$. The unburnt state has density $1.201\times10^{-3}$ and pressure $8.321\times10^{5}$, the tube has length $0.05$, and the initial interface lies at $0.005$.

\paragraph{Grids and reference.} The coarse solver uses 200 cells. The data store one state per interval in which the faster of the burnt-state and von Neumann-state wave speeds, enlarged by a factor of 1.2, crosses half a coarse cell. The von Neumann state is the shocked but still unburnt gas immediately behind the leading shock. Each coarse step advances two stored intervals, so the coarse time step is twice the storage interval of the reference. The reference solution uses the same scheme on a grid refined 128 times (25{,}600 cells) and is coarsened to the 200-cell grid. A case ends when the coarse-grid front comes within 5 cells of the right boundary, so each case spans 526 to 582 stored steps (263 to 291 coarse steps).

\paragraph{Case distribution and data.} Each case independently and uniformly draws an overdrive factor $f\in[1.0,1.6]$ (how far the initial detonation is driven beyond the Chapman--Jouguet state), a heat-release factor $q_0/q_0^{\rm base}\in[0.8,1.2]$, a rate factor $k_r/k_r^{\rm base}=10^{u}$ with $u\in[0,\log_{10}2.5]$, and an ignition-temperature factor $T_{\rm ign}/T_{\rm ign}^{\rm base}\in[0.70,0.85]$. It also draws, independently and uniformly, a random perturbation of the initial fuel field with amplitude in $[0,0.3]$ and correlation length in $[10,20]$ coarse cells, represented by 64 Fourier modes. The training, validation, and test sets contain 256, 32, and 256 cases drawn from this distribution with separate random seeds.

\paragraph{Closure.} The closure adds a conservative correction to the numerical flux at every interface. At each interface, its input consists of five features on the six surrounding cells (three on each side), namely $\rho/\rho_0$, $u/D_{\rm CJ}$, $p/(\rho_0D_{\rm CJ}^2)$, $Z$, and $T/T_{\rm ign}$, where $\rho_0$ is the unburnt density and $D_{\rm CJ}$ is the Chapman--Jouguet speed of the case. Its four outputs correct the fluxes of $\rho$, $\rho u$, $E$, and $\rho Z$ and are scaled by $0.01\times(\rho_0D_{\rm CJ},\rho_0D_{\rm CJ}^2,\rho_0D_{\rm CJ}^3,\rho_0D_{\rm CJ})$. The network consists of a convolution with kernel size 6 from 5 to 64 channels, two convolutions with kernel size 1 and 64 channels, and an output convolution with kernel size 1 to 4 channels, with GELU activations. Its $10{,}564$ parameters are all trained. The output layer is initialized to zero, so the untrained closure reproduces the coarse solver. The closure affects the ignition switch only through the temperature.

\paragraph{Window loss.} The step loss sums the mean squared errors of $\rho$, $\rho u$, and $E$, each normalized by the standard deviation of that variable along the reference trajectory. The window loss sums the step losses over the window. A failed rollout receives a penalty of ten times the largest single-step loss of the uncorrected solver on the validation windows.

\paragraph{Metrics.} Evaluation rolls each case out with hard switches from $t=0$ to its stored length. The main metric is the relative error of the wave speed (\%), fitted from the front trajectory within 20\%--90\% of the tube length. Each seed reports the median over the 256 test cases, with a failed case counted as 100\%.

At each stored time within the 20\%--90\% window used for the wave-speed fit, the maximum pressure over the five cells behind the leading shock is compared with the reference, and the relative error is averaged over time. Because the reference coarsened to 200 cells does not resolve the von Neumann spike (the thin pressure peak right behind the leading shock), the quantity compared is the post-shock peak pressure on the coarse grid. For this quantity, each seed takes the median over the 256 cases (with a failed case counted as 100\%), and the medians are averaged across seeds.

\subsection{Cortical spreading depolarization}
\label{app:cortical-sd}

\paragraph{Physical model and discretization.} The simulator is a minimal threshold-release model in the reaction--diffusion tradition of spreading depolarization, in which extracellular K$^+$ diffuses through the tissue and a cell whose K$^+$ concentration exceeds a threshold activates and then releases its own store \citep{tuckwell1978mathematical}. Each cell carries the K$^+$ concentration $c$, a releasable store $h$, a release activity $a$, and an excitability $r$, in nondimensional units where the coarse cell size and the coarse time step are one. Over a time step $\Delta t$, extracellular K$^+$ first advances by explicit diffusion with harmonic-mean face diffusivities, plus the divergence of the closure's flux correction. Each cell then evaluates the threshold switch $b=\mathbf 1[m\ge0]$ with normalized margin $m=(c-c_{\rm th}(1+\chi(1-r)))/c_{\rm th}$, where $c_{\rm th}$ is the local release threshold. The switch latches the release activity via $a\leftarrow a+(1-a)b$ and sets the excitability to $r\leftarrow r(1-b)$. Activated cells release their store gradually according to $h\leftarrow h\exp(-ka\Delta t)$, which depletes it exponentially. The released K$^+$ is added to $c$, which then relaxes toward the resting concentration $c_{\rm rest}$ with time constant $T_{\rm rest}$. We use $k=0.6$, $c_{\rm rest}=0.05$, $T_{\rm rest}=40$, and $\chi=20$. The activity does not decay and neither $r$ nor $h$ recovers, so a cell activates at most once and, after activation, releases its store at an exponentially decaying rate. Cells in the infarct core have no switch and no store.

\paragraph{Grids and reference.} The coarse solver uses a $64\times64$ grid with $\Delta t=1$ and runs 160 steps per case. The reference solves the same model with hard switches on a $256\times256$ grid with 16 substeps per coarse step and is block-averaged to the coarse grid. The activity and excitability of each block are averaged only over its releasing cells. The coarse model uses the block mean of the fine diffusivity, the mean threshold over the releasing cells of the block, and a store equal to the fraction of releasing cells. A coarse cell carries a switch when its block contains a releasing cell. Every rollout starts from the block-averaged reference state at $t=0$.

\paragraph{Case distribution and data.} Lengths are given in coarse cells. The fine-grid diffusivity is $D_{\rm diff}=0.12\exp(0.6\,\eta_1+0.3\,\eta_2)$, clipped to $[0.02,0.24]$, where $\eta_1$ and $\eta_2$ are unit-variance Gaussian random fields with Gaussian correlation functions of correlation length $0.375$ and $8$ coarse cells, respectively, so that $\eta_1$ varies below the coarse grid scale \citep{nicholson1998extracellular}. The threshold is $c_{\rm th}=\bar c_{\rm th}(1+0.1\,\eta_3)$, with $\bar c_{\rm th}$ uniform in $[0.22,0.30]$ and $\eta_3$ a field of the same kind with correlation length $4$. The lesion center is uniform in $[16,48]^2$. Inside the penumbra (a disc with radius uniform in $[10,18]$), the threshold is multiplied by 0.7. The infarct core is a concentric disc with radius uniform in $[3,7]$. The stimulus is a disc of radius 3 centered in a uniformly random direction from the lesion center at the penumbra radius plus a distance uniform in $[4,12]$, and its center is clamped to lie at least 4 cells from the boundary. Releasing cells inside the stimulus start activated with $c=1$, $a=1$, and $r=0$, whereas all other cells start at rest with $c=c_{\rm rest}$, $a=0$, and $r=1$, with $h=1$ outside the core. The training, validation, and test sets contain 128, 32, and 64 cases drawn successively from this distribution.

\paragraph{Closure.} The closure adds a conservative correction to the K$^+$ flux across every coarse face. Its input on every coarse cell is the state $(c,h,a,r)$ and five static fields, namely the block-mean diffusivity divided by 0.2, the threshold divided by its mean over the case, the initial store multiplied by 4, the fraction of releasing cells in the block, and the coefficient of variation of the fine diffusivity within the block. The network consists of three convolutions with kernel size 3, 24 channels, and GELU activations, followed by an output convolution with kernel size 3 and two channels, which give the corrections on the $x$ and $y$ faces after scaling by 0.05. Its $12{,}818$ parameters are all trained. The output layer is initialized to zero, so the untrained closure reproduces the uncorrected coarse model, which gives the activation-time error of the untrained closure reported in Table~\ref{tab:cortical-sd}. The flux correction compensates for variations of the diffusivity finer than the coarse grid.

\paragraph{Window loss.} The step loss is the mean over coarse cells of $((c-c_{\rm ref})/s)^2$, where $s$ is the standard deviation of the reference K$^+$ concentration over the training set. The window loss sums the step losses over the window. A state is invalid when it is not finite or when $c$ leaves $(-1,20)$. A failed rollout is frozen and receives, for every remaining step, a penalty of four times the largest single-step loss of the uncorrected solver on 256 validation windows.

\paragraph{Metrics.} Evaluation rolls each case out with hard switches from $t=0$ for 160 steps. The activation time of a coarse cell is the first step at which its activity rises through 0.5, which for the reference corresponds to the step at which half of the releasing cells of the block have fired. A cell that never activates is assigned step 161. The activation-time error of a case is the mean of $|t_{\rm model}-t_{\rm ref}|$ over the cells with a switch that activate in the model or in the reference, divided by 160 and expressed in percent, with a failed case counted as 100\%. Each seed reports the median over the 64 test cases. The four physical quantities are averaged over seeds.
\begin{itemize}
\item \emph{Activation mismatch.} The percentage of cells with a switch that activate in only one of the model and the reference, with a failed case counted as 100\%, averaged over the test cases.
\item \emph{Relative error of the K$^+$ field.} The norm of the difference between the predicted and reference K$^+$ concentrations over the whole rollout, divided by the norm of the change of the reference from its initial state, with a failed case counted as 1. Each seed takes the median over the test cases.
\item \emph{Penumbra arrival error.} The absolute difference between the steps at which the first penumbra cell activates in the model and in the reference, divided by 160 and expressed in percent. Each seed takes the median over the test cases. The penumbra of a test case covers 190 to 957 coarse cells (with a median of 446), and the reference wave reaches the penumbra in all 64 test cases.
\item \emph{Penumbra activation-time error.} The activation-time error restricted to the cells of the penumbra, computed in the same way as the main metric.
\end{itemize}

\subsection{Configuration of \ourmethod{}}
\label{app:eventflip-config}

Both variants use the threshold band $|m_e|<\tau\sigma$ with $\tau=4$, and Table~\ref{tab:baseline-config} lists the width $\sigma$ and the number $M$ of flip branches per window in each scenario. The width is in units of line loading on the power grid, of $T_{\rm ign}$ on the detonation wave, and of the local threshold on cortical spreading depolarization (Appendix~\ref{app:cortical-sd}). The flipped events are drawn with replacement from the proposal $\pi_e$ of Eq.~\eqref{eq:method-band-proposal} with $\lambda=0.05$, and each event vector is weighted by $1/(M\pi_e)$. On the detonation wave, $\sigma$ is constant within a case, so the kernel component is proportional to $\phi(m_e/\sigma)$. Appendix~\ref{app:branch-sampling} compares this proposal with uniform sampling on the band, which raises the error on the detonation wave and on cortical spreading depolarization.

A flip branch restarts the simulator from the state before the sampled event, sets that switch to the opposite state, and runs to the end of the window. \textsc{EventFlip-Rand} uses randomized switches in both the base trajectory and the branch, and the branch reuses the threshold noise of the base trajectory after the flipped event. \textsc{EventFlip-Hard} uses hard switches in both the base trajectory and the branch, and $\sigma$ sets only the band and the kernel weights.

\subsection{Training and evaluation protocol}
\label{app:protocol}

Every method receives 750{,}000 generation solves on the power grid (where a generation solve is one DC power-flow solve of one case) and 800{,}000 coarse solver steps on the detonation wave and on cortical spreading depolarization. A case consists of one initial condition with its reference trajectory, and every executed case-step counts toward the budget. Flip branches, ensemble members, and perturbed rollouts all count, with each flip branch charged $K-n_e$ steps from its restart step $n_e$, where $K$ is the window length. Validation rollouts are not counted on any simulator.

\paragraph{Training windows.} On the power grid, each window spans three generations as described in Appendix~\ref{app:power-grid}. On the detonation wave, each window spans 32 coarse steps from a random start in a random training case. All candidates within an update share its windows. On cortical spreading depolarization, a window likewise covers 32 coarse steps and starts at a random step of a random training case. Table~\ref{tab:baseline-config} lists the number of windows per update. Base trajectories and flip branches apply the same invalid-state checks.

\paragraph{Optimization.} \textsc{EventFlip-Rand}, \textsc{EventFlip-Hard}, Direct, Sigmoid, STE, AoBG, ES, RLOO, GRPO-clip, PPO, and Closure-RL update the closure with Adam at learning rate $10^{-3}$, whereas CMA-ES, SPSA, and EKI use their own update rules. On cortical spreading depolarization, every Adam method uses learning rate $3\times10^{-3}$. Gradient-norm clipping at 1.0 is applied to every Adam method on the power grid and, in the other two scenarios, to the methods whose update contains a likelihood-ratio estimate (AoBG, RLOO, GRPO-clip, PPO, Closure-RL, and the ablation without flip branches in Appendix~\ref{app:switch-reinforce}). All other methods run without clipping.

\paragraph{Checkpoints and evaluation.} On the power grid, each run scores the untrained closure and 40 checkpoints at evenly spaced points of the budget by the mean rollout loss on the 600 validation cases. On the detonation wave and on cortical spreading depolarization, each run stores the untrained closure and a checkpoint whenever the forward step count passes a multiple of 25{,}000. Each candidate is scored by the median error of the main metric on the 32 validation cases (the wave-speed error on the detonation wave and the activation-time error on cortical spreading depolarization). Each candidate is rolled out with hard switches from $t=0$, with a failed case counted as 100\%. Configurations are selected on validation data. Each run keeps the candidate with the lowest validation score, freezes it, and evaluates it once on the test set. Every metric reported in this appendix comes from these frozen checkpoints on the same test set.

\paragraph{Hardware.} Each run uses one NVIDIA A100-SXM4-80GB GPU. Solvers and closures run in double precision on the power grid and the detonation wave and in single precision on cortical spreading depolarization.

\subsection{Baselines}
\label{app:baselines}

The twelve baselines form two groups. The four gradient-based baselines are direct differentiation (Direct), sigmoid relaxation (Sigmoid), straight-through estimation (STE), and the $\alpha$-order batched gradient (AoBG) \citep{suh2022differentiable}. The eight derivative-free baselines are evolution strategies (ES), simultaneous perturbation stochastic approximation (SPSA), the covariance matrix adaptation evolution strategy (CMA-ES), ensemble Kalman inversion (EKI), REINFORCE with a leave-one-out baseline (RLOO), clipping in the style of group relative policy optimization (GRPO-clip), proximal policy optimization (PPO), and multi-agent closure reinforcement learning (Closure-RL) \citep{von2024closure}. Table~\ref{tab:baseline-config} lists the configuration of each baseline in the three scenarios. Perturbation scales marked as relative are multiples of the initial standard deviation of each weight tensor, and a tensor initialized to zero (such as the output layer) uses the same multiple as an absolute scale. ES, SPSA, CMA-ES, and EKI perturb the parameters of the closure and update them from the rollouts of the perturbed parameters.

The remaining settings are the same in all three scenarios. Direct holds the hard switches constant in backpropagation. Sigmoid replaces the switch by $\mathrm{sigmoid}(m/w)$ in the forward and backward pass, and STE keeps the hard forward pass and uses the derivative of this sigmoid in the backward pass. ES uses antithetic Gaussian perturbations of the hard-switch window loss. SPSA uses the gain exponents $\alpha=0.602$ and $\gamma=0.101$ and calibrates its step size on the calibration pairs so that the root-mean-square parameter step equals the relative perturbation. CMA-ES is the separable variant \citep{ros2008simple} with a maximum step size of 1.0, and on the detonation wave and cortical spreading depolarization its population is $4+\lfloor3\ln d\rfloor$ for $d$ parameters. EKI updates the full network with 32 members, a relative initial spread of 0.3, and inflation 1.05, and resamples the ensemble every 10 updates. GRPO-clip clips the per-dimension probability ratio at $\epsilon=0.2$ and runs 4 epochs of 4 minibatches with a target KL of 0.02. PPO uses the same clipping with GAE at $\lambda_{\rm GAE}=0.95$ \citep{schulman2015high} and a critic MLP with hidden layers $[32,32]$ trained by Adam at $10^{-3}$. Closure-RL uses 4 rollouts per window, discount 0.95, GAE at $\lambda_{\rm GAE}=0.95$, clipping $\epsilon=0.2$, entropy coefficient 0.05, 4 epochs of 4 minibatches, and a target KL of 0.02, and trains its policy and critic by Adam. The output noise of AoBG, RLOO, GRPO-clip, and PPO is added to the log-rating offset of every line on the power grid and to the closure outputs before scaling in the other two scenarios.

Among the gradient-based baselines, AoBG \citep{suh2022differentiable} adds Gaussian noise to the closure output at every interface and every step and combines two estimates of the gradient of the expected window loss, each from its own group of noisy hard-switch rollouts. The first-order estimate differentiates each rollout with every switch held constant, as direct differentiation does. The zeroth-order estimate multiplies the gradient of the log density of the noise by the window loss minus the loss of the noise-free rollout. The update is $\alpha$ times the first-order estimate plus $1-\alpha$ times the zeroth-order estimate. The weight $\alpha$ starts from the inverse-variance weight of the two sample means and is lowered, down to zero, so that $\alpha$ times the distance between the two estimates plus a confidence radius of the zeroth-order estimate stays below a threshold $\gamma$. The threshold $\gamma$ is selected on validation data in each scenario and is 1.35 on the power grid and $\infty$ on the other two, where $\alpha$ keeps the inverse-variance weight. In our runs, $\alpha$ stays near one on the detonation wave and on cortical spreading depolarization and near zero on the power grid, where the first-order estimate is zero. With $S$ samples per group, each update costs $2S+1$ rollouts of each window, $S$ for each estimate and one noise-free rollout, all charged to the forward budget.

Among the derivative-free baselines, RLOO, GRPO-clip, and PPO add Gaussian noise to the closure output at every interface (every line on the power grid and every face on cortical spreading depolarization) and every step, and roll out the hard-switch simulator with the noisy outputs. Every action of one step is credited with the same return, the loss of the rollout from that step to the end of the window.

Closure-RL \citep{von2024closure}, also derivative-free, treats each interface of the detonation wave and each face of cortical spreading depolarization as an agent. All agents share one fully convolutional Gaussian policy, whose mean is the closure output and whose standard deviation comes from a convolutional head on the closure features that is used only in training. A separate fully convolutional critic gives one value per agent. After every corrected step, the simulator also takes an uncorrected step from the same state, and the reward of a cell is the squared error of the uncorrected step minus that of the corrected step, both against the reference. Each agent is trained by PPO with advantages computed by GAE from its own rewards. We adapt two details to our simulators. An agent sits between cells and receives the mean reward of its adjacent cells, and a failed rollout gives every cell a fixed negative reward set by the failure penalty of Appendix~\ref{app:exp-details}. The uncorrected steps are charged to the forward budget, so each rollout costs twice its number of steps. Closure-RL does not apply to the power grid. There a rating changes the state only through a trip, and a trip changes the flows on the whole network, so no local error measures the effect of one rating.

\begin{table*}[t]
\centering
\caption{\textbf{Configurations of all methods in the three scenarios.} The widths $\sigma$ and $w$ are in units of line loading on the power grid, of $T_{\rm ign}$ on the detonation wave, and of the local threshold on cortical spreading depolarization. SD denotes a standard deviation. The first four baselines are gradient-based and the others are derivative-free.}
\label{tab:baseline-config}
\footnotesize
\setlength{\tabcolsep}{6pt}
\begin{tabular}{@{}llccc@{}}
\toprule
Method & Setting & \shortstack[c]{Power-grid\\cascade} & \shortstack[c]{Detonation\\wave} & \shortstack[c]{Cortical spreading\\depolarization} \\
\midrule
\textsc{EventFlip-Rand} & Width $\sigma$ & 0.02 & 0.02 & 0.02 \\
 & Flip branches $M$ & 4 & 1 & 8 \\
 & Windows per update & 32 & 8 & 16 \\
\addlinespace[2pt]
\textsc{EventFlip-Hard} & Width $\sigma$ & 0.02 & 0.02 & 0.04 \\
 & Flip branches $M$ & 4 & 1 & 8 \\
 & Windows per update & 32 & 8 & 16 \\
\midrule
Direct \citep{um2020solver,feng2022differentiable} & Windows per update & 32 & 8 & 16 \\
\addlinespace[2pt]
Sigmoid \citep{seo2026tokagrad} & Sigmoid width $w$ & 0.003 & 0.008 & 0.01 \\
 & Windows per update & 32 & 8 & 16 \\
\addlinespace[2pt]
STE \citep{bengio2013estimating} & Sigmoid width $w$ & 0.1 & 0.02 & 0.01 \\
 & Windows per update & 32 & 8 & 16 \\
\addlinespace[2pt]
AoBG \citep{suh2022differentiable} & Threshold $\gamma$ & 1.35 & $\infty$ & $\infty$ \\
 & Noise SD & 0.1 & 0.3 & 0.3 \\
 & Samples per group & 8 & 4 & 4 \\
 & Windows per update & 8 & 8 & 16 \\
\midrule
ES \citep{salimans2017evolution} & Perturbation pairs & 4 & 16 & 8 \\
 & Perturbation SD (absolute) & 0.03 & 0.01 & 0.01 \\
 & Windows per update & 8 & 8 & 16 \\
\addlinespace[2pt]
SPSA \citep{spall1992multivariate} & Relative perturbation & $3\times10^{-3}$ & $3\times10^{-3}$ & $10^{-2}$ \\
 & Calibration pairs & 16 & 4 & 4 \\
 & Windows per update & 8 & 8 & 16 \\
\addlinespace[2pt]
CMA-ES \citep{hansen2001completely} & Initial step size (relative) & $3\times10^{-2}$ & $10^{-2}$ & $3\times10^{-2}$ \\
 & Population & 26 & 31 & 32 \\
 & Windows per update & 8 & 8 & 16 \\
\addlinespace[2pt]
EKI \citep{iglesias2013ensemble} & Observation-noise SD & 0.03 & 0.3 & 0.3 \\
 & Windows per update & 8 & 2 & 2 \\
\addlinespace[2pt]
RLOO \citep{kool2019buy} & Noise SD & 0.1 & 0.3 & 0.3 \\
 & Group size & 8 & 4 & 4 \\
 & Windows per update & 8 & 8 & 16 \\
\addlinespace[2pt]
GRPO-clip \citep{shao2024deepseekmath} & Noise SD & 0.1 & 0.1 & 0.1 \\
 & Group size & 4 & 4 & 4 \\
 & Windows per update & 8 & 8 & 16 \\
\addlinespace[2pt]
PPO \citep{schulman2017proximal} & Noise SD & 0.1 & 0.03 & 0.1 \\
 & Group size & 4 & 4 & 4 \\
 & Windows per update & 8 & 8 & 16 \\
\addlinespace[2pt]
Closure-RL \citep{von2024closure} & Initial policy SD & n/a & 0.03 & 0.1 \\
 & Learning rate & n/a & $10^{-3}$ & $3\times10^{-3}$ \\
 & Windows per update & n/a & 8 & 16 \\
\bottomrule
\end{tabular}
\end{table*}

\section{Additional results}
\label{app:extra-results}

\subsection{\texorpdfstring{Full results on the three simulators}{Full results on the three simulators}}
\label{app:full-results}

\begin{figure*}[t]
\centering
\begin{minipage}[t]{0.49\linewidth}
\textbf{A}\par\vspace{1pt}
\figfile[width=\linewidth]{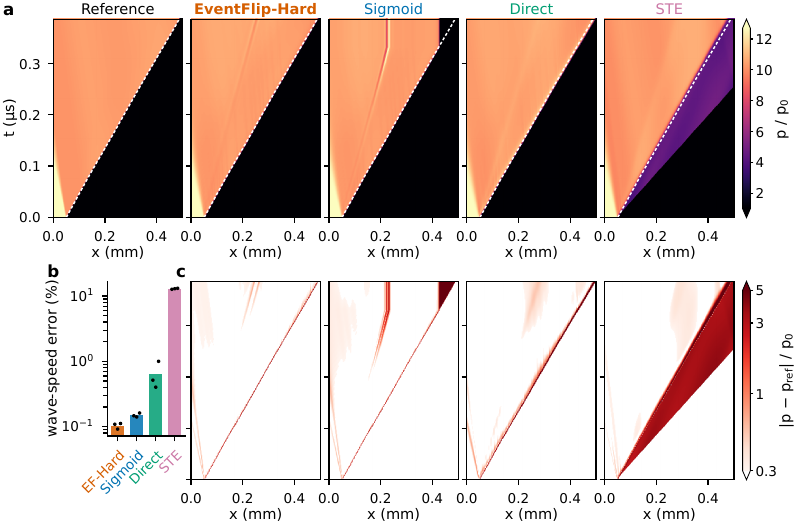}
\end{minipage}\hfill
\begin{minipage}[t]{0.49\linewidth}
\textbf{B}\par\vspace{1pt}
\figfile[width=\linewidth]{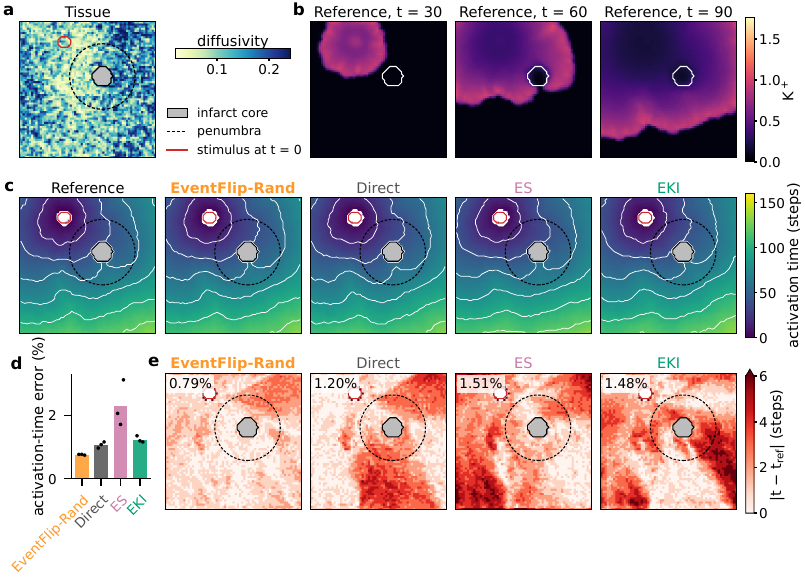}
\end{minipage}
\caption{\textbf{One test case on the detonation wave and on cortical spreading depolarization.} \textbf{A} Detonation wave. (a) Pressure in space and time, with the reference front dashed. (b) Test wave-speed error of each method. (c) Absolute pressure error against the reference, on a logarithmic color scale with errors below $0.3\,p_0$ in white. \textbf{B} Cortical spreading depolarization. (a) Tissue with the coarse-grid diffusivity, the penumbra, the infarct core, and the stimulus. (b) Reference K$^+$ concentration at three steps. (c) Activation time of each cell, with contour lines. (d) Test activation-time error of each method. (e) Absolute activation-time difference from the reference, with the color scale truncated at 6 steps. Panels within A(a), A(c), B(c), and B(e) share one color scale. In A(b) and B(d), bars are means over seeds 1--3 and dots are individual seeds.}
\label{fig:field-cases}
\end{figure*}

Fig.~\ref{fig:field-cases} shows one test case on each of the two field simulators, evaluated with the seed-1 closure of each method. In the detonation case, the front predicted by direct differentiation runs ahead of the reference front, and the closure trained by sigmoid relaxation produces a spurious wave behind the front. In the cortical case, the activation-time errors of direct differentiation accumulate in patches after the wave has passed the penumbra and the infarct core, while those of \textsc{EventFlip-Rand} remain mostly within two time steps.

Table~\ref{tab:grid}, Table~\ref{tab:detonation}, and Table~\ref{tab:cortical-sd} report the main metric of Table~\ref{tab:main} and the physical quantities for every method as means $\pm$ standard deviations over seeds 1--3.

\begin{table*}[t]
\centering
\caption{\textbf{Comparison of all methods on the power-grid cascade.} Means $\pm$ standard deviations over seeds 1--3. The untrained closure has a rollout loss of 0.181. Higher AUC is better. Rollout loss is the sum of the step losses over the cascade, and shed load and trip error are the absolute errors of the final shed load (in percent of the total load) and of the number of tripped lines. AUC is the severe-case AUC (Appendix~\ref{app:power-grid}). Bold and underlined values mark the best and second best. The first four baselines are gradient-based and the others are derivative-free.}
\label{tab:grid}
\small
\begin{tabular}{lcccc}
\toprule
Method & Rollout loss & Shed load (\%) & Trip error (lines) & AUC \\
\midrule
\textsc{EventFlip-Hard} & $\mathbf{0.067 \pm 0.007}$ & $\mathbf{0.629 \pm 0.087}$ & $\mathbf{0.647 \pm 0.059}$ & $\mathbf{0.924 \pm 0.012}$ \\
\textsc{EventFlip-Rand} & $\underline{0.072 \pm 0.013}$ & $\underline{0.682 \pm 0.197}$ & $\underline{0.679 \pm 0.162}$ & $\underline{0.907 \pm 0.040}$ \\
\midrule
Direct & $0.181 \pm 0.000$ & $2.085 \pm 0.000$ & $1.750 \pm 0.000$ & $0.612 \pm 0.000$ \\
Sigmoid & $0.144 \pm 0.027$ & $1.621 \pm 0.350$ & $1.348 \pm 0.305$ & $0.718 \pm 0.102$ \\
STE & $0.179 \pm 0.001$ & $2.064 \pm 0.016$ & $1.729 \pm 0.005$ & $0.599 \pm 0.003$ \\
AoBG & $0.143 \pm 0.027$ & $1.578 \pm 0.336$ & $1.342 \pm 0.279$ & $0.749 \pm 0.091$ \\
\midrule
ES & $0.116 \pm 0.008$ & $1.162 \pm 0.125$ & $1.064 \pm 0.180$ & $0.834 \pm 0.040$ \\
SPSA & $0.177 \pm 0.008$ & $2.003 \pm 0.142$ & $1.716 \pm 0.060$ & $0.639 \pm 0.046$ \\
CMA-ES & $0.101 \pm 0.010$ & $1.027 \pm 0.051$ & $0.935 \pm 0.053$ & $0.857 \pm 0.027$ \\
EKI & $0.145 \pm 0.018$ & $1.527 \pm 0.292$ & $1.323 \pm 0.172$ & $0.773 \pm 0.103$ \\
RLOO & $0.118 \pm 0.004$ & $1.277 \pm 0.038$ & $1.138 \pm 0.053$ & $0.800 \pm 0.014$ \\
GRPO-clip & $0.136 \pm 0.021$ & $1.478 \pm 0.234$ & $1.245 \pm 0.205$ & $0.785 \pm 0.027$ \\
PPO & $0.143 \pm 0.009$ & $1.493 \pm 0.105$ & $1.336 \pm 0.062$ & $0.764 \pm 0.036$ \\
Closure-RL & n/a & n/a & n/a & n/a \\
\bottomrule
\end{tabular}
\end{table*}

\begin{table}[t]
\centering
\caption{\textbf{Comparison of all methods on the detonation wave.} Means $\pm$ standard deviations over seeds 1--3. The untrained closure has a wave-speed error of 13.569\%. Wave-speed error is the relative error of the fitted wave speed, and peak pressure is the relative error of the post-shock peak pressure (Appendix~\ref{app:detonation}). Bold and underlined values mark the lowest and second-lowest errors. For the nine methods whose wave-speed error stays within 10\% of that of the untrained closure, the peak pressure is not reported (--). The first four baselines are gradient-based and the others are derivative-free.}
\label{tab:detonation}
\small
\resizebox{\ifdim\width>\linewidth\linewidth\else\width\fi}{!}{%
\begin{tabular}{lcc}
\toprule
Method & Wave-speed error (\%) & Peak pressure (\%) \\
\midrule
\textsc{EventFlip-Hard} & $\mathbf{0.104 \pm 0.011}$ & $\underline{1.24 \pm 0.15}$ \\
\textsc{EventFlip-Rand} & $\underline{0.120 \pm 0.010}$ & $\mathbf{0.92 \pm 0.06}$ \\
\midrule
Direct & $0.641 \pm 0.319$ & $11.3 \pm 1.2$ \\
Sigmoid & $0.150 \pm 0.011$ & $1.77 \pm 0.37$ \\
STE & $12.945 \pm 0.230$ & -- \\
AoBG & $0.411 \pm 0.038$ & $7.37 \pm 4.29$ \\
\midrule
ES & $13.188 \pm 0.109$ & -- \\
SPSA & $13.030 \pm 0.187$ & -- \\
CMA-ES & $13.016 \pm 0.105$ & -- \\
EKI & $13.677 \pm 0.188$ & -- \\
RLOO & $12.734 \pm 0.252$ & -- \\
GRPO-clip & $13.023 \pm 0.247$ & -- \\
PPO & $12.915 \pm 0.015$ & -- \\
Closure-RL & $13.360 \pm 2.342$ & -- \\
\bottomrule
\end{tabular}%
}
\end{table}

\begin{table*}[t]
\centering
\caption{\textbf{Comparison of all methods on cortical spreading depolarization.} Means $\pm$ standard deviations over seeds 1--3. The untrained closure has an activation-time error of 13.465\%. Act.-time error is the activation-time error, mismatch is the percentage of cells that activate in only one of the model and the reference, and K$^+$ error is the relative error of the K$^+$ field. Penumbra arrival is the error of the step at which the first penumbra cell activates, and penumbra act. time is the activation-time error restricted to the penumbra (Appendix~\ref{app:cortical-sd}). Bold and underlined values mark the lowest and second-lowest errors. The first four baselines are gradient-based and the others are derivative-free.}
\label{tab:cortical-sd}
\footnotesize
\setlength{\tabcolsep}{4pt}
\begin{tabular}{lccccc}
\toprule
Method & Act.-time error (\%) & Mismatch (\%) & K$^+$ error (ratio) & Penumbra arrival (\%) & Penumbra act.\ time (\%) \\
\midrule
\textsc{EventFlip-Hard} & $\underline{0.793 \pm 0.004}$ & $\underline{0.137 \pm 0.007}$ & $\underline{0.156 \pm 0.004}$ & $\mathbf{0.000 \pm 0.000}$ & $\mathbf{0.522 \pm 0.010}$ \\
\textsc{EventFlip-Rand} & $\mathbf{0.759 \pm 0.014}$ & $\mathbf{0.134 \pm 0.005}$ & $\mathbf{0.146 \pm 0.005}$ & $\mathbf{0.000 \pm 0.000}$ & $\underline{0.544 \pm 0.019}$ \\
\midrule
Direct & $1.068 \pm 0.097$ & $0.149 \pm 0.011$ & $0.174 \pm 0.006$ & $\underline{0.312 \pm 0.312}$ & $0.651 \pm 0.079$ \\
Sigmoid & $13.319 \pm 0.253$ & $3.591 \pm 0.173$ & $0.807 \pm 0.000$ & $2.500 \pm 0.000$ & $7.318 \pm 0.096$ \\
STE & $13.319 \pm 0.181$ & $4.451 \pm 1.288$ & $0.813 \pm 0.008$ & $2.500 \pm 0.000$ & $7.484 \pm 0.160$ \\
AoBG & $1.248 \pm 0.076$ & $0.144 \pm 0.006$ & $0.204 \pm 0.021$ & $0.417 \pm 0.361$ & $0.839 \pm 0.021$ \\
\midrule
ES & $2.302 \pm 0.735$ & $0.158 \pm 0.016$ & $0.309 \pm 0.059$ & $1.250 \pm 0.000$ & $1.457 \pm 0.378$ \\
SPSA & $4.245 \pm 2.162$ & $0.427 \pm 0.429$ & $0.447 \pm 0.129$ & $1.458 \pm 0.361$ & $2.424 \pm 1.055$ \\
CMA-ES & $9.371 \pm 0.237$ & $1.596 \pm 0.099$ & $0.702 \pm 0.004$ & $1.875 \pm 0.000$ & $5.014 \pm 0.158$ \\
EKI & $1.237 \pm 0.104$ & $0.146 \pm 0.011$ & $0.219 \pm 0.020$ & $0.625 \pm 0.000$ & $0.931 \pm 0.116$ \\
RLOO & $11.563 \pm 1.336$ & $2.360 \pm 0.725$ & $0.760 \pm 0.042$ & $1.979 \pm 0.180$ & $6.782 \pm 0.445$ \\
GRPO-clip & $8.462 \pm 0.364$ & $1.270 \pm 0.148$ & $0.665 \pm 0.015$ & $1.771 \pm 0.180$ & $4.508 \pm 0.226$ \\
PPO & $13.390 \pm 0.110$ & $3.644 \pm 0.017$ & $0.805 \pm 0.002$ & $2.188 \pm 0.000$ & $7.353 \pm 0.095$ \\
Closure-RL & $1.050 \pm 0.133$ & $0.152 \pm 0.027$ & $0.160 \pm 0.012$ & $0.625 \pm 0.000$ & $1.055 \pm 0.226$ \\
\bottomrule
\end{tabular}
\end{table*}

\subsection{\texorpdfstring{Budget curves}{Budget curves}}
\label{app:budget-curves}

Fig.~\ref{fig:budget} shows the test error of the plotted methods at each level of the forward budget. Direct differentiation stalls early. Its error does not decrease after the first budget level on any of the three simulators, and on the detonation wave all three seeds reach their validation-best checkpoint within the first 100k steps. The better of the two variants of \ourmethod{} reaches the full-budget error of the strongest baseline at 100k of the 750k generation solves on the power grid and at 200k of the 800k coarse solver steps on the detonation wave. On cortical spreading depolarization, PPO remains almost untrained. Sigmoid relaxation and straight-through estimation do not follow the gradient of the simulator there (Appendix~\ref{app:fidelity}) and remain at 13.3\% at every budget level.

\subsection{\texorpdfstring{Power grid by case severity}{Power grid by case severity}}
\label{app:severity}

Table~\ref{tab:severity} splits the 600 test cases of the power grid by their reference outcome into 504 cases without a cascade, 29 cascades that shed less than 5\% of the load, and 67 severe cases that shed at least 5\%, as in the AUC of Table~\ref{tab:grid}. When no line trips, all methods incur a loss below 0.02. On the 29 cascades that shed less than 5\% of the load, the losses of all methods lie between 0.13 and 0.25. In the 67 severe cases, early line trips redistribute power flows across the network and trigger multi-generation islanding and load shedding. On these severe cases, \textsc{EventFlip-Hard} lowers the rollout loss by 38\% relative to CMA-ES and by more than half relative to sigmoid relaxation and direct differentiation.

The top row of Fig.~\ref{fig:teaser} illustrates one of these severe cases, simulated with the trained seed-1 closure of each method shown.

\begin{table}[t]
\centering
\caption{\textbf{Evaluation of the rollout loss by case severity on the power-grid cascade.} Means over seeds 1--3 within each group of test cases. Bold and underlined values mark the lowest and second-lowest losses. The first four baselines are gradient-based and the others are derivative-free.}
\label{tab:severity}
\footnotesize\setlength{\tabcolsep}{3pt}
\resizebox{\ifdim\width>\linewidth\linewidth\else\width\fi}{!}{%
\begin{tabular}{lccc}
\toprule
Method & No cascade (504) & Cascade, not severe (29) & Severe (67) \\
\midrule
\textsc{EventFlip-Hard} & 0.005 & 0.167 & \textbf{0.487} \\
\textsc{EventFlip-Rand} & \underline{0.003} & 0.154 & \underline{0.551} \\
\midrule
Direct & 0.017 & 0.204 & 1.403 \\
Sigmoid & \textbf{0.002} & \textbf{0.133} & 1.211 \\
STE & 0.007 & 0.221 & 1.452 \\
AoBG & 0.009 & 0.173 & 1.138 \\
\midrule
ES & \textbf{0.002} & 0.162 & 0.953 \\
SPSA & 0.015 & 0.193 & 1.389 \\
CMA-ES & 0.008 & 0.148 & 0.780 \\
EKI & 0.010 & 0.244 & 1.115 \\
RLOO & 0.004 & \underline{0.145} & 0.963 \\
GRPO-clip & 0.007 & \textbf{0.133} & 1.111 \\
PPO & 0.013 & 0.158 & 1.118 \\
\bottomrule
\end{tabular}%
}
\end{table}

\subsection{\texorpdfstring{Trip history}{Trip history}}
\label{app:timing}

This section compares the predicted trip history with the reference history on the power grid (Table~\ref{tab:timing-grid}).

On the power grid, the path gradient is zero (Appendix~\ref{app:power-grid}), so direct differentiation keeps the untrained ratings and almost all of its error consists of missed trips (Table~\ref{tab:timing-grid}). The reference cascades trip 1.76 lines per case, and 96 of the 600 test cases cascade. Direct differentiation misses 1.612 trips per case, compared with 0.505 for \textsc{\ourmethod-Hard}. A reference trip counts as recovered when the line trips in the prediction (in the same or another generation). By this count, \textsc{\ourmethod-Hard} recovers 71\% of the reference trips and direct differentiation recovers 8.5\%. For direct differentiation, 89\% of its loss comes from cases with a missed trip. Cases whose predicted trip history matches the reference exactly incur zero loss, so the loss comes entirely from wrong trips. Per cascading case, direct differentiation misses 10.07 trips and \textsc{\ourmethod-Hard} 3.16.

\begin{table}[t]
\centering
\caption{\textbf{Evaluation of the predicted trip history on the power-grid cascade.} The loss is the rollout loss, the three trip columns count line trips per test case, and the last column gives the share of cases with an exact trip history, all as means over seeds 1--3. Wrong-generation trips occur in a different cascade generation from the reference. Missed counts reference trips of lines that never trip in the prediction. The missed, wrong-generation, and correct trips sum to the 1.76 reference trips per case.}
\label{tab:timing-grid}
\footnotesize\setlength{\tabcolsep}{3pt}
\resizebox{\ifdim\width>\linewidth\linewidth\else\width\fi}{!}{%
\begin{tabular}{lccccc}
\toprule
Method & Loss & Missed & Wrong generation & Correct & Exact history \\
\midrule
Direct & 0.181 & 1.612 & 0.042 & 0.107 & 83.5\% \\
\textsc{EventFlip-Hard} & 0.067 & 0.505 & 0.254 & 1.001 & 90.0\% \\
\textsc{EventFlip-Rand} & 0.072 & 0.579 & 0.268 & 0.913 & 89.8\% \\
\bottomrule
\end{tabular}%
}
\end{table}

\subsection{\texorpdfstring{Learned line ratings}{Learned line ratings}}
\label{app:ratings}

Table~\ref{tab:ratings} compares the ratings learned on the power grid with the reference ratings of Appendix~\ref{app:power-grid}, which come from the IEEE 738 heat balance. Because the closure is trained only on cascade trajectories, it never sees these reference ratings. The error of a line in one test case is $e=\log(c_\theta/c_{\rm nom})-g^\star$, which leaves out the scatter $\varepsilon$ that the closure cannot represent. Each of the 600 test cases has 184 branches after its two initial outages, for a total of 110{,}400 pairs of a line and a test case. On the 1{,}901 pairs whose peak loading reaches at least 0.8, \textsc{EventFlip-Hard} achieves a root-mean-square error of 0.022 to 0.028 in the logarithmic rating with a mean bias near zero, and the untrained closure has an error of 0.211 with a bias of $+0.131$. Averaged over seeds 1--3 on these pairs, the root-mean-square error is 0.026 for \textsc{EventFlip-Hard} and 0.077 for CMA-ES.

The closure maps the features of a line, including its conductor class, the wind speed, the ambient temperature, and the wind direction, to its rating. The loss depends on the ratings only through which lines trip and when, so the cascades constrain the rating of a line only when its loading comes close to that rating. The error is therefore lower on the heavily loaded pairs than over all pairs (Table~\ref{tab:ratings}). Over all 110{,}400 pairs, the error still falls from 0.186 for the untrained closure to 0.046 to 0.051 for \textsc{EventFlip-Hard}. The lightly loaded lines include the transformers, of which only 1 of the 6{,}564 pairs reaches a peak loading of 0.8, and the 345\,kV lines capped by the St.~Clair limit, which are never heavily loaded.

\begin{table}[t]
\centering
\caption{\textbf{Evaluation of the learned line ratings on the power-grid cascade.} Root-mean-square error of the logarithmic rating against the reference rating for seeds 1, 2, and 3.{ A pair is one line in one test case, and the second column keeps the 1{,}901 pairs with a peak loading of at least 0.8.}}
\label{tab:ratings}
\footnotesize
\begin{tabular}{lcc}
\toprule
Method & All pairs & Peak loading $\ge 0.8$ \\
\midrule
\textsc{EventFlip-Hard} & 0.046 / 0.051 / 0.048 & 0.022 / 0.028 / 0.028 \\
\textsc{EventFlip-Rand} & 0.046 / 0.051 / 0.043 & 0.033 / 0.031 / 0.035 \\
Sigmoid & 0.106 / 0.073 / 0.167 & 0.128 / 0.091 / 0.221 \\
CMA-ES & 0.088 / 0.093 / 0.122 & 0.080 / 0.063 / 0.087 \\
Untrained closure & 0.186 & 0.211 \\
\bottomrule
\end{tabular}
\end{table}

\subsection{Shared and independent threshold noise}
\label{app:common-noise}

In \textsc{EventFlip-Rand}, each flip branch reuses the threshold noise of the base trajectory after the flipped switch, whereas redrawing the threshold noise independently sets $\operatorname{Cov}(G_e(1),G_e(0)\mid H_e)=0$ in Eq.~\eqref{eq:paired-noise-variance} while preserving the conditional expectation and forward cost. Table~\ref{tab:common-noise} compares the two schemes at the untrained and trained \textsc{EventFlip-Rand} closures. The table also reports training from scratch with each scheme for seeds 1--3.

On cortical spreading depolarization, the sampled 32-step windows contain hundreds of band events on average. Sharing the threshold noise reduces the minibatch jump-term variance by 9--22 times and lowers the trained activation-time error from 1.101\% to 0.759\%. On the power-grid cascade and the detonation wave, only a few switches per window enter the threshold band, and the two schemes give similar minibatch jump-term variance and trained test error.

Sharing the threshold noise therefore pays off where many switches lie in the band. Independent threshold noise changes the decisions of other near-threshold switches along the wave front between the two rollouts, whereas under shared noise the switches unaffected by the flip take the same values in both.

\begin{table}[t]
\centering
\caption{\textbf{Ablation of shared and independent threshold noise in \textsc{EventFlip-Rand}.} The variance ratio divides the variance of the single-update minibatch jump-term estimate under independent noise by that under shared noise. Test errors in the metrics and units of Table~\ref{tab:main} are means over seeds 1--3.}
\label{tab:common-noise}
\small
\setlength{\tabcolsep}{6pt}
\renewcommand{\arraystretch}{1.12}
\begin{tabular}{@{}llccc@{}}
\toprule
Simulator & Closure & \shortstack[c]{Jump-term\\var.\ ratio} & {\shortstack[c]{Shared\\noise}} & \shortstack[c]{Independent\\noise} \\
\midrule
\multirow{2}{*}{Power grid} & Untrained & 1.00 & \multirow{2}{*}{0.072} & \multirow{2}{*}{0.072} \\
 & Trained & 0.94 & & \\
\midrule
\multirow{2}{*}{Detonation wave (\%)} & Untrained & 1.11 & \multirow{2}{*}{0.120} & \multirow{2}{*}{0.132} \\
 & Trained & 1.06 & & \\
\midrule
\multirow{2}{*}{\shortstack[l]{Cortical spreading\\depolarization (\%)}} & Untrained & 22.15 & \multirow{2}{*}{0.759} & \multirow{2}{*}{1.101} \\
 & Trained & 9.06 & & \\
\bottomrule
\end{tabular}
\end{table}

\subsection{No flip branch}
\label{app:switch-reinforce}

This ablation keeps the randomized simulator, $\sigma$, the threshold band, and the pathwise term of \textsc{EventFlip-Rand} and runs no flip branch. In addition to the pathwise term, the update follows the gradient of
\[
\frac{1}{N_{\rm batch}}\sum_{e\in\mathcal B_\tau}\operatorname{sg}\bigl(G_e-\bar G_e\bigr)\log p(b_e),
\]
where $\operatorname{sg}(\cdot)$ stops the gradient, $N_{\rm batch}$ is the number of windows in the update, and the sum runs over the events in the threshold band of these windows. Here $G_e$ is the downstream loss of the base trajectory after event $e$ (the quantity $G_e(b_e)$ of Section~\ref{sec:jump}), and $\bar G_e$ is the mean of the same quantity at the same step of the window over the other windows of the minibatch. Finally, $p(b_e)$ is the probability of the sampled switch decision, equal to $\Phi(m_e/\sigma)$ if the switch is on and $1-\Phi(m_e/\sigma)$ if it is off, differentiated through the margin along the base trajectory. The number of windows per update and the learning rate are those of \textsc{EventFlip-Rand} (Appendix~\ref{app:protocol}). Without flip branches, an update costs 96 generation solves on the power grid and 256 and 512 coarse solver steps on the detonation wave and cortical spreading depolarization, and each run uses the forward budget of Table~\ref{tab:main}. Table~\ref{tab:switch-reinforce-variance} compares the variance of the two estimates of the jump term.

\begin{table}[t]
\centering
\caption{\textbf{Variance of the jump term without a flip branch.} Ratio of the variance of the minibatch jump-term gradient without a flip branch to that of \textsc{EventFlip-Rand} on a fixed batch of training windows with only the threshold noise redrawn. Each variance is multiplied by the forward steps of one update. The trained closure is the validation-best checkpoint of \textsc{EventFlip-Rand} for seed 1.}
\label{tab:switch-reinforce-variance}
\footnotesize
\setlength{\tabcolsep}{4pt}
\begin{tabular}{@{}lcc@{}}
\toprule
Simulator & Untrained closure & Trained closure \\
\midrule
Power grid & 14.1 & 68.4 \\
Detonation wave & 48.0 & 70.7 \\
Cortical spreading depolarization & 4{,}681 & 30.5 \\
\bottomrule
\end{tabular}
\end{table}

Without a flip branch, the expectation of the jump-term estimate is the same as with a flip branch, but the estimate now carries the differences in downstream loss between windows as well as the decisions of every switch near its threshold. At five of the six checkpoints, the variance without a flip branch is 14 to 71 times that of \textsc{EventFlip-Rand} at equal forward cost, and at the untrained closure on cortical spreading depolarization it reaches 4{,}681 times. On the power grid, the variance without a flip branch at equal forward cost is 14.1 and 68.4 times that of \textsc{EventFlip-Rand} at the untrained and trained closures, while its test error of 0.064 is comparable to the 0.072 of \textsc{EventFlip-Rand}.

\begin{table}[t]
\centering
\caption{\textbf{How far each gradient is from the measured change of the loss.} Each cell is the directional derivative that a method returns, divided by the derivative measured with finite differences on the simulator, given as the smallest to the largest value over the tested directions. A value of 1 means that the gradient predicts the measured change of the loss exactly. The two columns evaluate all methods on the same closure (either the untrained closure or the seed-1 closure of \textsc{EventFlip-Rand}), and the \textsc{EventFlip-Rand} rows use the band estimator.}
\label{tab:fidelity-ratio}
\footnotesize\setlength{\tabcolsep}{3pt}

\resizebox{\ifdim\width>\linewidth\linewidth\else\width\fi}{!}{%
\begin{tabular}{llcc}
\toprule
 & & \multicolumn{2}{c}{Returned derivative / measured derivative} \\
Simulator & Gradient of & {Untrained closure} & {Trained closure} \\
\midrule
Power grid & \textsc{EventFlip-Rand} & 1.1--1.2 & 0.34--1.1 \\
 & Direct & 0 & 0 \\
 & Sigmoid & 0.54--2.2 & $2\times10^{-4}$--$2\times10^{-3}$ \\
 & STE & 12--34 & 0.009--0.28 \\
\midrule
Detonation wave & \textsc{EventFlip-Rand} & 0.94--1.0 & 0.87--1.0 \\
 & Direct & 0.012--1.0 & 0.18--0.82 \\
 & Sigmoid & 0.36--1.1 & 0.028--0.92 \\
 & STE & 1.7--$2.8\times10^5$ & 280--$1.3\times10^5$ \\
\midrule
\shortstack[l]{Cortical spreading\\depolarization} & \textsc{EventFlip-Rand} & 0.61--0.92 & 0.90--1.2 \\
 & Direct & 0.003--0.43 & 0.17--1.3 \\
 & Sigmoid & $2.1\times10^5$--$1.1\times10^7$ & $6.5\times10^4$--$1.5\times10^6$ \\
 & STE & 22--$5.1\times10^3$ & $1.0\times10^6$--$1.1\times10^7$ \\
\bottomrule
\end{tabular}%
}
\end{table}

\begin{table*}[t]
\centering
\caption{\textbf{Ablation of the proposal for the flip branches.} Test error of each seed in the metrics and units of Table~\ref{tab:main}.{ Kernel weight, 5\% uniform is the default proposal of Appendix~\ref{app:eventflip-config}.}}
\label{tab:sampling}
\footnotesize
\begin{tabular}{lllcccc}
\toprule
Simulator & Method & Proposal & Seed 1 & Seed 2 & Seed 3 & Mean \\
\midrule
Power grid & \textsc{EventFlip-Hard} & {Kernel weight, 5\% uniform} & 0.062 & 0.064 & 0.075 & 0.067 \\
 & & Uniform & 0.067 & 0.062 & 0.063 & 0.064 \\
 & \textsc{EventFlip-Rand} & {Kernel weight, 5\% uniform} & 0.071 & 0.059 & 0.085 & 0.072 \\
 & & Uniform & 0.068 & 0.070 & 0.063 & 0.067 \\
\midrule
Detonation wave (\%) & \textsc{EventFlip-Hard} & {Kernel weight, 5\% uniform} & 0.109 & 0.092 & 0.112 & 0.104 \\
 & & Uniform & 0.123 & 0.136 & 0.117 & 0.126 \\
 & \textsc{EventFlip-Rand} & {Kernel weight, 5\% uniform} & 0.115 & 0.113 & 0.131 & 0.120 \\
 & & Uniform & 0.171 & 0.126 & 0.362 & 0.220 \\
\midrule
\shortstack[l]{Cortical spreading\\depolarization (\%)} & \textsc{EventFlip-Rand} & {Kernel weight, 5\% uniform} & 0.765 & 0.768 & 0.743 & 0.759 \\
 & & Uniform & 0.943 & 0.856 & 0.874 & 0.891 \\
 & & Kernel only & 0.786 & 0.851 & 0.773 & 0.803 \\
 & \textsc{EventFlip-Hard} & {Kernel weight, 5\% uniform} & 0.795 & 0.795 & 0.788 & 0.793 \\
 & & Uniform & 0.866 & 0.856 & 0.928 & 0.883 \\
 & & Kernel only & 0.863 & 0.788 & 1.027 & 0.893 \\
\bottomrule
\end{tabular}
\end{table*}

\subsection{Gradient fidelity}
\label{app:fidelity}

This section checks numerically that the estimator of \textsc{EventFlip-Rand} is an unbiased estimate of the gradient of the smoothed objective, as Proposition~\ref{prop:full-support-unbiased} states, and that the band truncation of Proposition~\ref{prop:truncation-bias} removes a negligible part of it.

On each simulator, we evaluate the untrained closure and the seed-1 closure of \textsc{EventFlip-Rand} on the same 16 training windows without training, and take as the reference the directional derivative of $J_\sigma$ computed by central differences with common random numbers. We test eight unit directions, four random and four along the gradients of the band estimator of \textsc{EventFlip-Rand}, direct differentiation, sigmoid relaxation, and straight-through estimation, and keep those on which the central differences at two step sizes agree. The full-support estimator covers every eligible event with $M=64$, and the band estimator uses the configuration of the main tables.

Along the tested directions, the approximate 95\% upper bound on the relative bias of the full-support and band estimators of \textsc{EventFlip-Rand} has a median between 0.08\% and 5\% of $\|\nabla_\theta J_\sigma\|$ at every checkpoint, consistent with Proposition~\ref{prop:full-support-unbiased}. Table~\ref{tab:fidelity-ratio} reports the ratio of the returned to the measured directional derivative for every method.

Among the methods in Table~\ref{tab:fidelity-ratio}, the gradient of \textsc{EventFlip-Rand} is the only one whose ratio stays between 0.34 and 1.2 on all three simulators and both closures. Direct differentiation returns zero on the power grid and mostly less than the measured derivative on the other two simulators, because it omits the jump term. On the detonation wave, where sigmoid relaxation is the strongest baseline, the gradient of sigmoid relaxation lies between 0.36 and 1.1 of the measured derivative at the untrained closure and between 0.028 and 0.92 at the trained closure. However, it falls to $2\times10^{-4}$ to $2\times10^{-3}$ of the measured derivative at the trained closure on the power grid and exceeds it by factors of $10^4$ to $10^7$ on cortical spreading depolarization. The gradient of straight-through estimation is within a factor of about 100 of the measured derivative only on the power grid and exceeds it by up to $10^5$ on the detonation wave and $10^7$ on cortical spreading depolarization. These factors are consistent with sigmoid relaxation and straight-through estimation staying at the error of the untrained closure on cortical spreading depolarization (Table~\ref{tab:main}). For sigmoid relaxation, whether a fractional switch is a usable stand-in depends on the switch itself. On the detonation wave, the switch is evaluated anew at every step, and a fractional value represents a cell burning at a reduced rate. On cortical spreading depolarization the switch latches, and an activated cell raises its own threshold so that it cannot activate again. Sigmoid relaxation differentiates this threshold feedback through a fractional switch at every step, whereas in the hard-switch simulator the feedback has no further effect once the cell activates. Straight-through estimation differentiates a sigmoid that its forward pass never applies.

The truncation bias is measured by evaluating the sum over the events outside the band for each draw of $\xi$, which gives $\|R_{\rm tail}(\tau)\|/\|\nabla_\theta J_\sigma\|$. At the default $\tau=4$, this ratio lies between $2.3\times10^{-6}$ and $9.7\times10^{-4}$ across the checkpoints. At $\tau=1$, it lies between 0.13 and 1.07. At the default band, the measured truncation bias is therefore negligible.

\subsection{Kernel-weighted and uniform branch sampling}
\label{app:branch-sampling}

The default proposal draws 95\% of the branches by kernel weight and 5\% uniformly on the band, with importance weights $1/(M\pi_e)$. Table~\ref{tab:sampling} compares it with uniform sampling on the band and, on cortical spreading depolarization, with sampling by kernel weight alone. All three proposals give the same expected update and have the same forward cost. On the power grid, a training window holds only 1.0 to 1.3 switches in the threshold band on average (Table~\ref{tab:properties-full}), so the two proposals mostly draw the same switch and the difference between them is smaller than the inter-seed spread. Allocating branches uniformly across the band raises the error by 21\%--83\% on the detonation wave and by 11\%--17\% on cortical spreading depolarization. Uniform sampling places as many branches near the edges of the threshold band, where the sensitivity $\kappa_\sigma(m_e)$ is smallest, as near $m_e=0$, where it is largest. Retaining the 5\% uniform component prevents large importance weights near the band edges and reduces the error by 5\%--11\% relative to pure kernel sampling on cortical spreading depolarization.

\end{document}